\documentclass{article}

\usepackage{arxiv}

\usepackage[utf8]{inputenc} 
\usepackage[T1]{fontenc}    
\usepackage{hyperref}       
\usepackage{url}            
\usepackage{booktabs}       
\usepackage{amsfonts}       
\usepackage{nicefrac}       
\usepackage{microtype}      
\usepackage{lipsum}		
\usepackage{graphicx}

\usepackage{lineno,hyperref}
\usepackage{bm}
\usepackage{amsmath}
\usepackage{subcaption}
\usepackage{amssymb}
\newcommand{\rulesep}{\unskip\ \vrule\ }
\newcommand{\argmax}{\operatornamewithlimits{argmax}}
\newcommand{\argmin}{\operatornamewithlimits{argmin}}

\title{Shallow2Deep: Indoor Scene Modeling by Single Image Understanding}

\author{
	{\hspace{1mm}Yinyu Nie}\\
	Bournemouth University\\
	Poole BH12 5BB, U.K.\\
	\texttt{ynie@bournemouth.ac.uk} \\
	\And
	{\hspace{1mm}Shihui Guo}\thanks{Corresponding authors}\\
	Xiamen University\\
	Xiamen 361005, China.\\
	\texttt{guoshihui@xmu.edu.cn}\\
	\And
	{\hspace{1mm}Jian Chang$^{\ast}$}\\
	Bournemouth University\\
	Poole BH12 5BB, U.K.\\\
	\texttt{jchang@bournemouth.ac.uk}\\
	\And
	{\hspace{1mm}Xiaoguang Han}\\
	The Chinese University of Hong Kong (Shenzhen)\\
	Shenzhen 518172, China.\\\
	\texttt{hanxiaoguang@cuhk.edu.cn}\\
	\And
	{\hspace{1mm}Jiahui Huang}\\
	Tsinghua University\\
	Beijing 100084, China\\\
	\texttt{huang-jh18@mails.tsinghua.edu.cn}\\
	\And
	{\hspace{1mm}Shi-Min Hu}\\
	Tsinghua University\\
	Beijing 100084, China\\\
	\texttt{shimin@tsinghua.edu.cn}\\
	\And
	{\hspace{1mm}Jian Jun Zhang}\\
	Bournemouth University\\
	Poole BH12 5BB, U.K.\\\
	\texttt{jzhang@bournemouth.ac.uk}\\
	}

\date{}

\renewcommand{\headeright}{}
\renewcommand{\undertitle}{}

\begin{document}
\maketitle

\begin{abstract}
Dense indoor scene modeling from 2D images has been bottlenecked due to the absence of depth information and cluttered occlusions.
We present an automatic indoor scene modeling approach using deep features from neural networks.
Given a single RGB image, our method simultaneously recovers semantic contents, 3D geometry and object relationship by reasoning indoor environment context.
Particularly, we design a shallow-to-deep architecture on the basis of convolutional networks for semantic scene understanding and modeling. It involves multi-level convolutional networks to parse indoor semantics/geometry into non-relational and relational knowledge. 
Non-relational knowledge extracted from shallow-end networks (e.g. room layout, object geometry) is fed forward into deeper levels to parse relational semantics (e.g. support relationship).
A Relation Network is proposed to infer the support relationship between objects. All the structured semantics and geometry above are assembled to guide a global optimization for 3D scene modeling. 
Qualitative and quantitative analysis demonstrates the feasibility of our method in understanding and modeling semantics-enriched indoor scenes by evaluating the performance of reconstruction accuracy, computation performance and scene complexity.
\end{abstract}

\keywords{Scene understanding \and Image-based modeling \and Semantic modeling \and Relational reasoning}

\section{Introduction}
\label{sec:introduction}
Understanding indoor environment is of significant impact and has already been applied in the domains of interior design, real estate, etc.
In recent years, 3D scanning and reconstruction of indoor scenes have been intensively explored using various sensors \cite{chen20153d}.
Understanding 3D indoor contents from an RGB image shows its unique significance for our daily-life applications, e.g. 3D digital content generation for social media and content synthesis for virtual reality and augmented reality.

3D scene modeling from a single image is challenging as it requires computers to perform equivalently as human vision to perceive and understand indoor context with only color intensities.
It generally requires for blending various vision tasks \cite{chen20153d} and most of them are still under active development, e.g. object segmentation \cite{bu2016scene}, layout estimation \cite{wei2018understanding} and geometric reasoning \cite{liu2018data}.
Although machine intelligence has reached comparable human-level performance in some tasks (e.g. scene recognition \cite{zhou2018places}), those techniques are only able to represent a fragment knowledge of full scene context.

With the lack of depth clues, prior studies reconstructed indoor scenes from a single image by exploiting shallow image features (e.g. line segments and HOG descriptors \cite{zhang2015single,liu2018data}) or introducing depth estimation \cite{huang2018holistic,nie2018semantic} to search object models. Other works adopt Render-and-Match strategy to obtain CAD scenes with their renderings similar as input images \cite{izadinia2017im2cad}.
However, it is still an unresolved problem when indoor geometry is over-cluttered and complicated.
The reasons are threefold.
First, complicated indoor scenes involve heavily occluded objects, which could cause missing contents in detection \cite{izadinia2017im2cad}.
Second, cluttered environments significantly increase the difficulty of camera and layout estimations, which critically affects the reconstruction quality \cite{lee2017roomnet}.
Third, compared to the large diversity of objects in real scenes, the reconstructed virtual environment is still far from satisfactory (missing small pieces, wrong labeling).
Existing methods have explored the use of various contextual knowledge, including object support relationship \cite{huang2018holistic,nie2018semantic} and human activity \cite{huang2018holistic}, to improve modeling quality.
However, their relational (or contextual) features are hand-crafted and would fail to cover a wide range of objects in cluttered scenes.

\begin{figure*}[!htp]
	\centering
	\includegraphics[width=\textwidth]{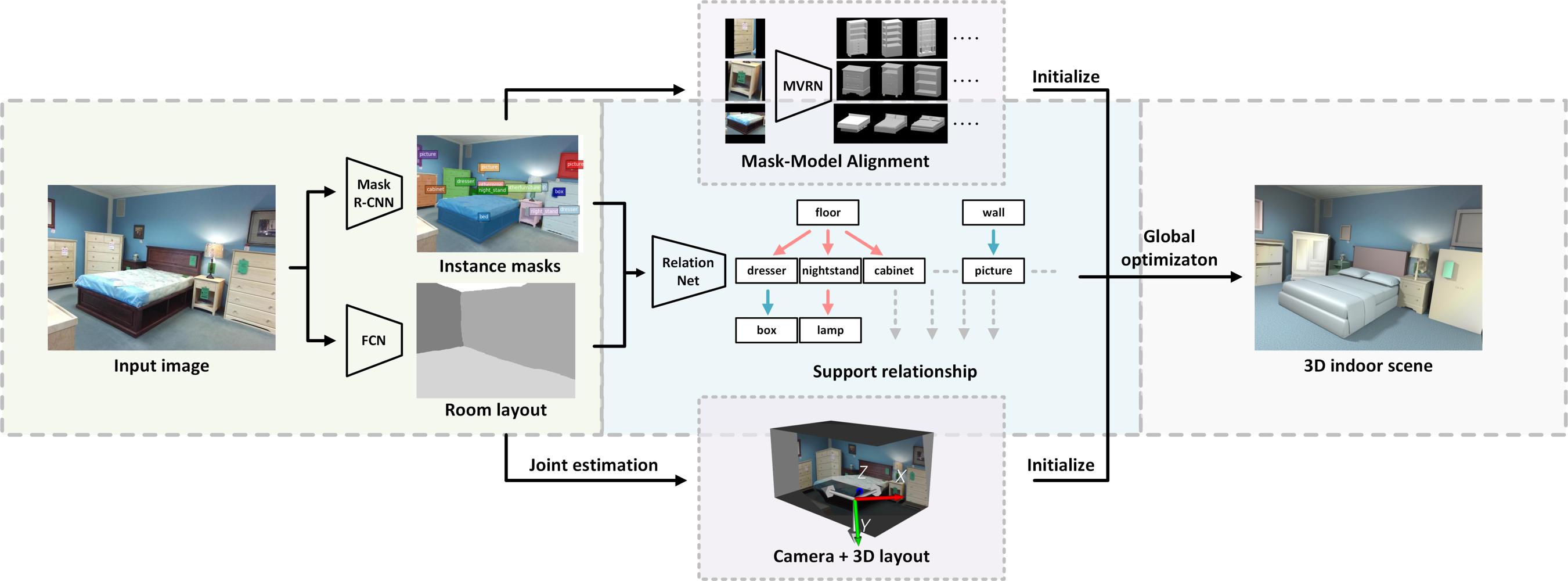}
	\caption{
		Pipeline of indoor scene modeling from a single image. The whole process is divided into three phases: 1. non-relational semantics parsing (e.g. room layout and object masks); 2. support relationship inference; 3. global scene optimization.}
	\label{fig:pipeline}
\end{figure*}

Different from previous works, our work aims at dense scene modeling. We extract and assemble object semantics (i.e. object masks with labels) and geometric contents (i.e. room layout and object models) into structured knowledge after being processed with shallower stacks of neural networks (see Figure~\ref{fig:pipeline}).
All the extracted semantics and geometry above is passed to deeper stacks of networks to infer support relationships between objects, which guides the final 3D scene modeling with global optimization.
We take advantage of these object support relations and improve the modeling performance, in terms of object diversity and accuracy.
We also propose a novel method to jointly estimate the camera parameters and room layout, which helps to improve scene modeling accuracy compared with using the existing methods.
In summary, our contributions can be listed as follows:
\begin{itemize}
	\item a Relation Network for support inference. This network predicts support relationship between indoor objects. It improves the reconstruction quality in the stage of global scene optimization, particularly increasing the accuracy of object placement in occlusions.
	\item a global optimization strategy for indoor scene synthesis. It incorporates the outputs from former networks and iteratively recovers 3D scenes to make them contextually consistent with the scene context. It performs effectively in inferring the shape of heavily occluded objects.
	\item a unified scene modeling system backboned by convolutional neural networks (CNN). With the capability of CNNs in parsing scene contents, latent image features are perceived and accumulated from consecutive networks. It outputs compact indoor context with a shallow-to-deep streamline to automatically generate semantics-enriched 3D scenes. 
\end{itemize}

\section{Related Work}
\label{relw}
\paragraph{\textbf{Indoor Content Parsing}} Capturing indoor contents is a prerequisite for modeling semantic scenes. 
Prior studies have explored scene parsing from various aspects with a single image input, which can be divided into two branches by their targets: 1) semantic detection and 2) geometric reasoning.

In semantic detection, deep learning holds satisfying performance in extracting latent features. It provides high accuracy in acquiring various types of scene semantics, e.g. scene types (bedroom, office or living room, etc.) \cite{cheng2018scene}, object labels (bed, lamp or picture etc.) \cite{wang2018hierarchical}, instance masks \cite{he2017mask} and room layout (locations of wall, floor and ceiling) \cite{lee2017roomnet,ren2016coarse}. 
On these raw semantics, abstract descriptions like scene grammar have been used to summarize them into a hierarchical structure.
It divides indoor contents into groups by functionality or spatial relationship (e.g. affiliation relations) for scene understanding \cite{liu2014creating}. Unlike semantic detection, geometric reasoning refers to capturing spatial clues from images, e.g. depth map \cite{zhang2018deep}, surface normal map \cite{eigen2015predicting} and object geometry (including model retrieval \cite{li2015joint} and reconstruction \cite{wu2018learning}).

However, these scene parsing methods are well tailored for a specific task. We design our holistic scene parsing step on their basis. Semantic and geometric clues are extracted and unified to form structured knowledge for scene understanding and 3D modeling. 
\paragraph{\textbf{Support Inference}} This step is inspired by the research \cite{xu2013sketch2scene} where a freehand sketch drawing is turned into semantically valid, well arranged 3D scenes.
They performed co-retrieval and co-placement of 3D models which are related to each other (e.g. a chair and a desk) by jointly processing the sketched objects in pair.
The success of their work showed the significance of relations between objects, while our work contributes to this importance element by automatically inferring the object support relationship.

Support relationship provides a sort of geometric constraint between indoor objects to build scenes more robustly.
This originates from daily experience that an object requires some support to counteract the gravity.
Support inference from RGB images is an ambiguous problem without knowing the 3D geometry, where occlusions usually make the supporting parts invisible in the field of view.
However, the arrangement of indoor furniture generally follows a set of interior design principles and living habits (e.g. tables are mostly supported by the floor; pictures are commonly on walls).
These latent patterns behind scenes make the support relationship a kind of priors that can be learned by viewing various indoor rooms.
Earlier studies addressed this problem by designing specific optimization problems \cite{silberman2012indoor,xue2015towards} considering both depth clues and image features.
Apart from inferring support relations, many researchers represented the support condition by object stability and indoor safety \cite{zheng2015scene,6618852}.
Moreover, the support relation is also a type of spatial relationship in scene grammar to enhance contextual bindings between objects.
Other approaches implemented support inference to understand scenes with a scene graph \cite{liu2014creating,huang2018holistic}.
However, these methods either require for depth information, or rely on hand-crafted priors or models.
In our work, we formulate the support inference problem into a Visual-Question Answering (VQA) form \cite{antol2015vqa}, where a Relation Network is designed to end-to-end link the object pairs and infer their support relations.

\paragraph{\textbf{Single-view Scene Modeling}} Indoor scene modeling from RGB images can be divided into two branches: 1. layout prediction and 2. indoor content modeling. Based on the Manhattan assumption \cite{coughlan1999manhattan}, layout prediction represents indoor layout with cuboid proposals using line segments \cite{hedau2009recovering} or CNN feature maps \cite{ren2016coarse,lee2017roomnet}.

To reconstruct indoor contents, previous methods adopted cuboid shapes \cite{deng2017amodal,huang2018cooperative} to recover the orientation and placement of target objects without the need for querying CAD model datasets.
However, the geometric details are weak because objects are only represented by a bounding box.
Rather than using cuboid shapes, other methods produced promising results in placement estimation of a single object by aligning CAD models with the object image \cite{lim2014fpm,wu2016single}.
Other methods leveraged shallow features (e.g. line segments, edges and HOG features) \cite{zhang2015single,liu2017indoor,liu2018data} to segment images and retrieve object models, or used a scene dataset as priors to retrieve object locations based on co-occurrence statistics \cite{seeThrough_HuetingReddy_3DV18}.
They either asked for human interaction or hand-crafted priors in parsing scene contents, while semantics and object geometry are learned with our method, allowing the capability of handling extendable object categories.

Recent studies also considered CNNs to detect objects and estimate room layout \cite{izadinia2017im2cad,nie2018semantic} with informative scene knowledge \cite{huang2018holistic}.
Huang et al. \cite{huang2018holistic} estimated depth maps and surface normal maps from RGB images with scene grammar to optimize the object placement. However, depth prediction is sensitive if the input distribution is slightly different from the training data \cite{nie2018semantic}.
Instead of tailoring scene grammar to improve the reconstruction results, we incorporates the relational reasoning in our process to infer the object relationship with a Relation Network.
A parallel development \cite{izadinia2017im2cad} followed a Render-and-Match strategy to optimize object locations and orientations, which did not involve any depth clues and other relational constraints.
CAD scenes are iterated until their renderings are sufficiently close to the input images.
In contrast, our method does not refer to extra depth prediction and scene rendering iterations, which significantly boosts the computing performance.
The scene modeling is built on a unified CNN architecture.
Intermediate semantic and geometric clues are parsed and accumulated by sub-modules, and reorganized with support relations to guide the scene modeling.

\section{Overview}\label{overview}
Our framework is built on the hypothesis that, features produced in each phase could be accumulated to feed into the consequent networks for deeper scene understanding.
This process is divided into three phases, as illustrated in Figure~\ref{fig:pipeline}.
The first phase obtains non-relational semantics (i.e.room layout, object masks and labels) and retrieves a small set of 3D object candidates from a large model library (Section~\ref{contextparsing}).
This part takes advantage of a number of recent development in computer vision communities. We tailored a selection of methods to precondition the non-relational features for solving the scene modeling problem in later two phases.

In the second phase, we introduce a Relation Network to infer support relationships between objects (Section~\ref{sec:relation}). 
This relational semantics offers physical constraints to organize those non-relational information into a reasonable contextual structure for 3D modeling.

The third phase assembles the geometric contents to model the 3D scene contextually consistent with these relational and non-relational semantics (Section~\ref{scenemodeling}). 
The 3D room layout and camera orientation are jointly estimated to ensure their consistence. It provides two coordinate systems (the room coordinate system and the camera coordinate system) for the global optimization in scene modeling and refinement.

\section{Non-Relational Semantics Parsing}\label{contextparsing}
\paragraph{\textbf{2D Layout Estimation}} Layout estimation provides room boundary geometry (i.e. the location of the floor, ceiling and walls). Using CNNs to produce layout features, current works \cite{ren2016coarse,lee2017roomnet} generally ask for camera parameters to estimate vanishing points for layout proposal decision. 
We adopt the Fully Convolutional Network (FCN) from \cite{ren2016coarse} to extract the layout edge map and label map. These feature maps present a coarse prediction of 2D layout (see Appendix~\ref{appendix_results}). 
An accurate 3D layout is jointly estimated along with camera parameters for further scene modeling (see Section~\ref{sec:layout}).

\begin{figure}[!ht]
	\centering
	\includegraphics[width=0.7\linewidth]{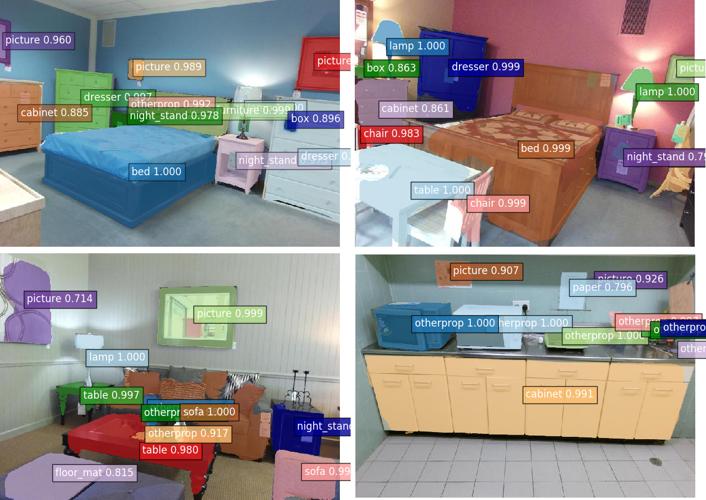}
	\caption{Instance segmentation samples}
	\label{fig:seg}
\end{figure}

\paragraph{\textbf{Scene Segmentation}} We segment images at the instance-level to obtain object category labels and corresponding 2D masks. 
Object masks present meaningful clues to initialize object sizes, 3D locations and orientations. 
Particularly, we introduce the Mask R-CNN \cite{he2017mask} to capture object masks with instance segmentation. 
We customize the Mask R-CNN backboned by ResNet-101 \cite{he2016deep}, with the weights pre-trained on the MSCOCO dataset \cite{matterport_maskrcnn_2017}. 
It is fine-tuned on the NYU v2 dataset \cite{silberman2012indoor} which contains 1,449 densely labeled indoor images covering 37 common and 3 `other' categories. 
(The training strategy is detailed in Appendix~\ref{appendix_seg}).
Since object masks act significantly in the latter stages, we append Mask R-CNN with the Dense Conditional Random Field (DCRF) \cite{krahenbuhl2011efficient} to merge overlaps and improve mask edges. 
Besides, wall, floor and ceiling masks are removed as they are precisely decided in the 2D layout estimation. 
Segmentation samples are shown in Figure~\ref{fig:seg}.

\begin{figure}[!ht]
	\centering
	\includegraphics[width=\linewidth]{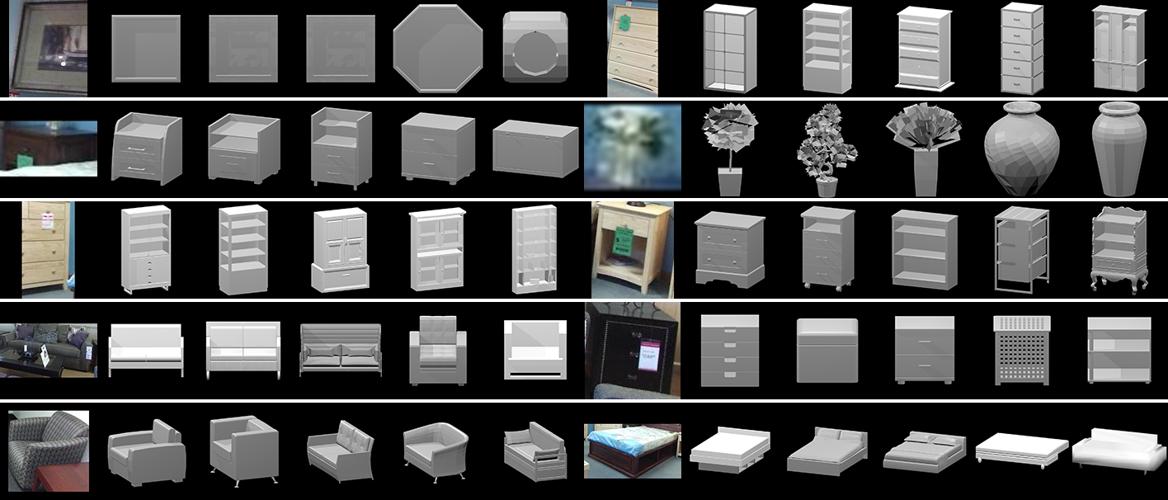}
	\caption{CAD model candidates. For each object image, we search our model dataset and output five similar candidates for scene modeling.}
	\label{fig:model_retrieval}
\end{figure}

\paragraph{\textbf{Model Retrieval}}\label{sec:model_ret} This task is to retrieve CAD models with the most similar appearance to the segmented object images.
A Multi-View Residual Network (MVRN) pretrained on ShapeNet \cite{chang2015shapenet} is introduced for shape retrieval.
Similar with \cite{izadinia2017im2cad,huang2018holistic}, we align and render each model from 32 viewpoints (two elevation angles at 15 and 30 degrees, and 16 uniform azimuth angles) for appearance-based matching.
A Multi-View Convolutional Network \cite{su2015multi} backboned with ResNet-50 is adopted as feature extractors to view CAD models from different viewpoints.
This type of architecture is designed to mimic human eyes by observing an object from multiple viewpoints to recognize the object shape. 
Deep features from a single view is represented by a 2048-dimensional vector (i.e. the last layer size of ResNet-50).
This compact descriptor enables us to match models efficiently in the vector space.
The similarity between an image and a model can be measured by the cosine distance: $\max_{i\in[1,32]}\cos(\bm{f}, \bm{f}^{\mathrm{m}}_{i}), \bm{f},\bm{f}^{\mathrm{m}}_{i}\in\mathcal{R}^{2048}$, where $\bm{f}$ and $\bm{f}^{\mathrm{m}}_{i}$ respectively denote the shape descriptor of the object image and a rendering of the model.
The model set construction and training strategy are detailed in Appendix~\ref{appendix_model_ret}. Furthermore, we fine-tune the orientation of matched models with ResNet-34 (see Discussions). 
Figure~\ref{fig:model_retrieval} shows some matched samples on our model set. 
Top-5 candidates are selected for global scene optimization in Section~\ref{scenemodeling}.

\begin{figure*}[h]
	\centering
	\includegraphics[width=\linewidth]{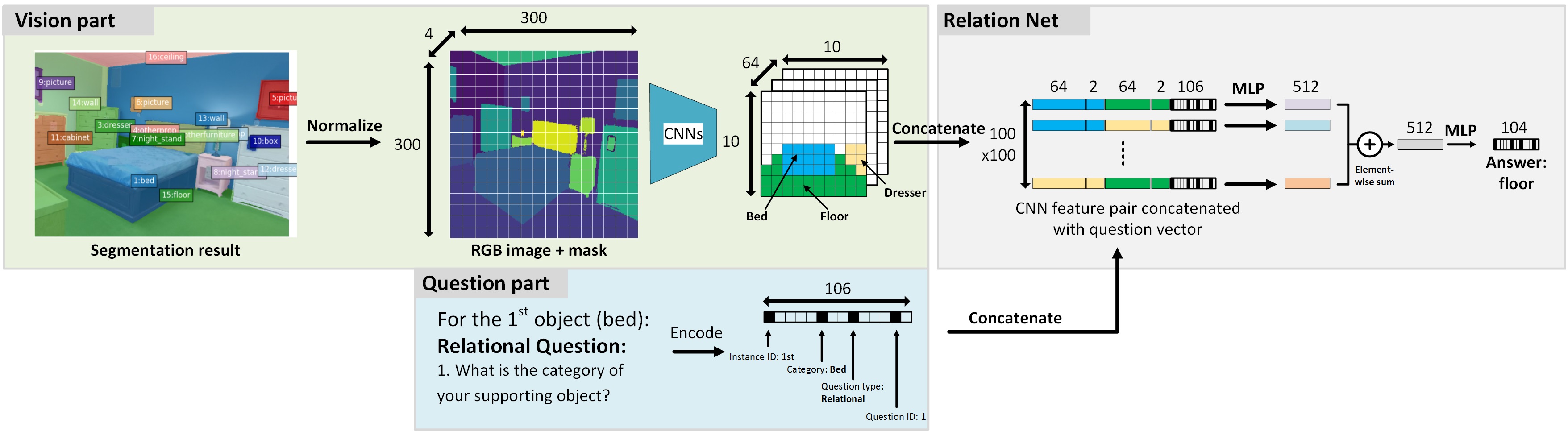}
	\caption{Relation Network for support inference. The whole architecture consists of three parts. The vision part and the question part are responsible for encoding object images and related questions separately, and the Relation Network answers these questions based on the image features.}
	\label{fig:relationalNet}
\end{figure*}

\section{Relational Reasoning}
\label{sec:relation}
Section~\ref{contextparsing} dedicates to parsing indoor scenes into non-relational contents.
We here aim to extract relational clues from these upstream outputs to conclude support relationships between objects.
This relationship serves as physical constraints to guide scene modeling. 

As assumed in existing works \cite{silberman2012indoor,wong2015smartannotator}, two support types are considered in our paper (i.e. support from behind, e.g. on a wall, or below, e.g. on a table). 
Every object except layout instances (i.e. wall, ceiling and floor) must be supported by another one. 
For objects which are supported by hidden instances, we treat them as being supported by layout instances. 

Unlike non-relational semantics, relational context asks for not only the object property features, but also the contextual link between object pairs.
Thus, a key is to combine the object feature pairs with specific task descriptions for support reasoning.
It can be intuitively formulated as a Visual Question Answering (VQA) manner \cite{antol2015vqa,santoro2017simple}: given the segmentation results, which instance is supporting object A?
Is it supported from below or behind?
With this insight, we configure a Relation Network to answer these support relationship questions by linking image features.
Our network is designed as shown in Figure~\ref{fig:relationalNet}.
The upstream of the Relation Network consists of two parts which encode visual images (with masks) and questions respectively.

\begin{figure}[!ht]
	\centering
	\includegraphics[width=0.7\linewidth]{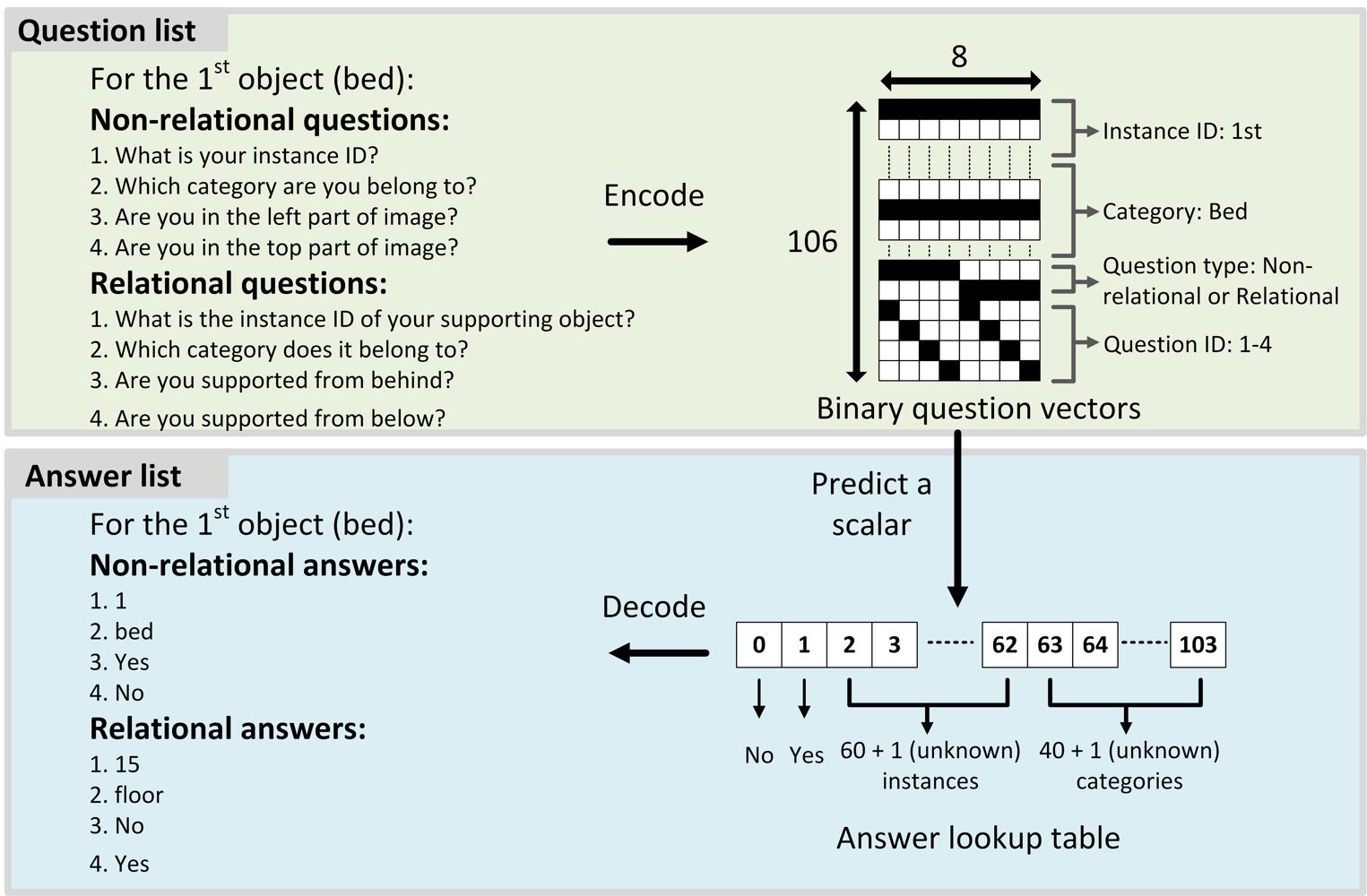}
	\caption{Questions and answers for training}
	\label{fig:VQA_questions}
\end{figure}

In the \textbf{Vision} part, the RGB image (color intensities, 3-channel) is normalized to $\left[0,1\right]$ and appended with its mask (instance labels, 1-channel), followed by a scale operation to a 300x300x4 matrix. We then generate 10x10x64 CNN feature vectors after convolutional operations. In the \textbf{Question} part, for each object instance, we customize our relational reasoning by answering two groups of questions: non-relational and relational; four questions for each. Taking the bed in Figure~\ref{fig:relationalNet} as an example, the related questions and corresponding answers are encoded as shown in Figure~\ref{fig:VQA_questions}. We design the four relational questions for support inference, and the other four non-relational questions as regularization terms to make our network able to identify the target object we are querying. In our implementation, we train the network on NYU v2 \cite{silberman2012indoor}. In a single image, maximal 60 indoor instances with 40 categories are considered. Therefore, for the $i$-th object which belongs to the $j$-th category, we encode the $k$-th question from the $m$-th group to a 106-d (106=60+40+4+2) binary vector. 

The outputs of the \textbf{Vision} and the \textbf{Question} parts are concatenated. We represent the 10x10x64 CNN features by 100 of 64-d feature vectors, and form all possible pairs of these feature vectors into 100x100 pairs. The 100x100 feature pairs are appended with their 2D coordinates (2-d) and exhaustively concatenated with the encoded question vector (106-d), then go through two multi-layer perceptrons to answer the questions (see network specifications in Appendix~\ref{appendix_rel}). For each question, the Relation Network outputs a scalar between 0 and 103. We decode it into a human-language answer by indexing the lookup table as illustrated in Figure~\ref{fig:VQA_questions}. The correct rate on the testing dataset of NYU v2 reaches 80.62\% and 66.82 \% on non-relational and relational questions respectively.

In our experiment, we observe that the numbering of instance masks is randomly given from the object segmentation, which undermines the network performance on the first relational question (see Figure~\ref{fig:VQA_questions}). In our implementation, we use the last three relational questions to predict the category of the supporting object and the support type, and keep the first one as a regularization term. The exact supporting instance can be identified by maximizing the prior supporting probability between the target object and its neighbors:
\begin{equation}
\label{eqn:02}
\mathrm{O}_{j^{*}}=\argmax_{\mathrm{O}_{j}\in\mathcal{N}(\mathrm{O}_{i})} \, \mathrm{P}(\mathcal{C}(\mathrm{O}_{j}) | \mathcal{C}(\mathrm{O}_{i}), \mathrm{T}_{k}), \,\mathcal{C}(\mathrm{O}_{j}) \in \mathcal{SC}(\mathrm{O}_{i}),
\end{equation}
where $\mathrm{O}_{i}$ and $\mathcal{N}(\mathrm{O}_{i})$ respectively denote the $i$-th object and its neighboring instances (layout instances are neighbors to all objects).
$\mathcal{C}(\mathrm{O}_{j})$ represents the category label of object $\mathrm{O}_{j}$. $\mathcal{SC}(\mathrm{O}_{i})$ indicates the top-5 (in our experiment) category candidates of $\mathrm{O}_{i}$'s supporting object, and $\mathrm{T}_{k}$ denotes the support type, $k=1,2$. Hence $\mathrm{P}(\mathcal{C}(\mathrm{O}_{j}) | \mathcal{C}(\mathrm{O}_{i}), \mathrm{T}_{k})$ means the probability of $\mathcal{C}(\mathrm{O}_{j})$ supporting $\mathcal{C}(\mathrm{O}_{i})$ by $\mathrm{T}_{k}$. The prior probability $\mathrm{P}$ is obtained by counting from the training data (see Appendix~\ref{appendix_prior_height} for details). The supporting instance is represented by $\mathrm{O}_{j^{*}}$. This process can improve the testing accuracy on the four relational questions by a large margin (from 66.82\% to 82.74\%).

\section{Global Scene Optimization}\label{scenemodeling}
The final process is composed of two steps: scene initialization and contextual refinement.
The first step initializes camera, 3D layout and object properties. 
The second step involves an iterative refinement to pick correct object CAD models and fine-tune their sizes, locations and orientations with support relation constraints.

\subsection{Scene Initialization}
\begin{figure*}[h!]
	\centering
	\includegraphics[width=\textwidth]{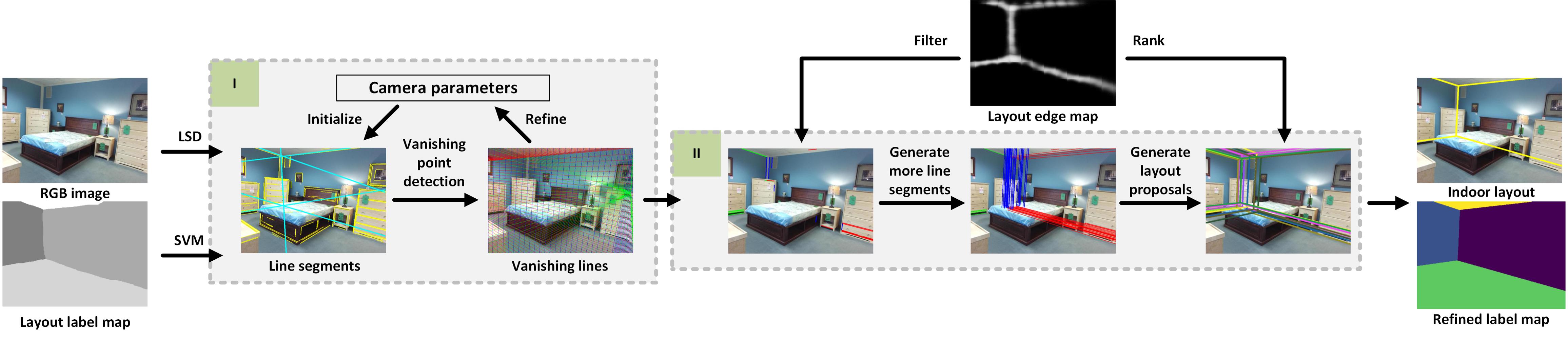}
	\caption{
		Camera-layout joint estimation. The camera parameters and vanishing points are jointly optimized in Part I, which leads to generate room layout proposals in Part II. The optimal layout is decided by the maximal probability score in layout edge map.}
	\label{fig:layout_estimation}
\end{figure*}

\paragraph{\textbf{Camera-layout Joint Estimation}}\label{sec:layout}
The camera-layout estimation is illustrated in Figure~\ref{fig:layout_estimation}. 
We jointly estimate camera parameters and a refined room layout by minimizing the angle deviations between the layout lines and vanishing lines in images (see Part I in Figure~\ref{fig:layout_estimation}). 
We firstly detect line segments from both the original image 
and the layout label map using LSD \cite{von2012lsd} and support vector machine (SVM) respectively. 
With the initialized camera parameters, orthogonal vanishing points are detected with the strategy proposed by \cite{lu20172}. 
The quality of vanishing points is scored by the count and length of the line segments they cross through. 
Longer line segments (like layout lines) would contribute more and guide the orthogonal vanishing lines in alignment with room orientation (see Part I in Figure~\ref{fig:layout_estimation}). 
However, an improper camera initialization, particularly in cluttered environments, would often cause faulty estimation of 3D room layout \cite{huang2018holistic}. 
We include iterations to improve the camera parameters from the detected line segments and produce a refined room layout simultaneously.

We denote the three orthogonal vanishing points by $\{\mathbf{vp}_{i}\}$, and the line segment set that (nearly) crosses through $\mathbf{vp}_{i}$ as $\mathcal{L} ( \mathbf{vp}_{i} ), i=1,2,3$. Both of them are expressed by homogeneous coordinates. 
Similar to K-Means clustering, for the $i$-th cluster $\mathcal{L} (\mathbf{vp}_{i})$, we re-estimate a new vanishing point $\mathbf{vp}^{*}_{i}$ by decreasing its distances to line segments in $\mathcal{L} (\mathbf{vp}_{i})$. This problem can be formulated as:
\begin{equation}
\label{eqn:01}
\begin{aligned}
& \mathbf{vp}^{*}_{i} = \argmin_{\mathbf{vp}_{i}} \, \bm{\varepsilon}^{\mathrm{T}}\bm{\varepsilon},\\
& [\bm{l}_{1}, \bm{l}_{2}, ..., \bm{l}_{N_{i}}]^{\mathrm{T}}\mathbf{vp}_{i}=\bm{\varepsilon}, i=1,2,3,
\end{aligned}
\end{equation}
where $\bm{l}_{k}$ denotes the coordinates of a line segment in cluster $\mathcal{L} ( \mathbf{vp}_{i} )$, $k=1,2,...,N_{i}$. $N_{i}$ is the capacity of $\mathcal{L} ( \mathbf{vp}_{i} )$. We solve it with the eigen decomposition to obtain the eigen vector corresponding to the smallest eigen value of $[\bm{l}_{1}, \bm{l}_{2}, ..., \bm{l}_{N_{i}}]^{\mathrm{T}}[\bm{l}_{1}, \bm{l}_{2}, ..., \bm{l}_{N_{i}}]$ as the updated $\mathbf{vp}_{i}$. 
After that, camera parameters can be updated with the renewed vanishing points by \cite{kovsecka2002video}.
With this strategy, the vanishing points and camera parameters can be jointly optimized as each of them iteratively converges. 

\begin{figure}
	\centering
	\begin{subfigure}[t]{0.23\textwidth}
		\includegraphics[width=\textwidth]  
		{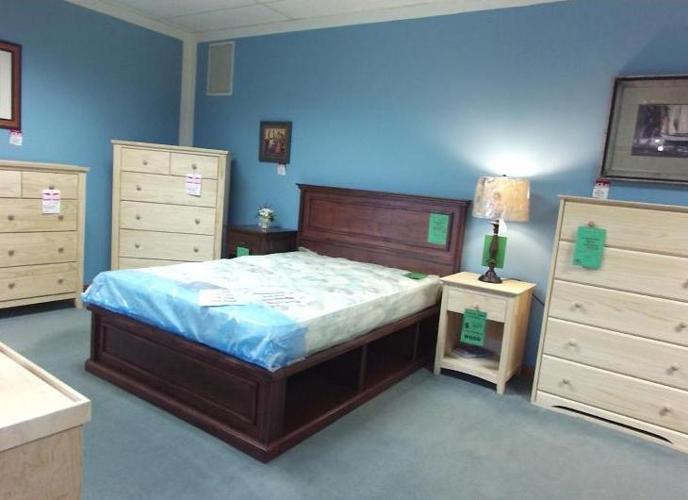}
		\includegraphics[width=\textwidth]
		{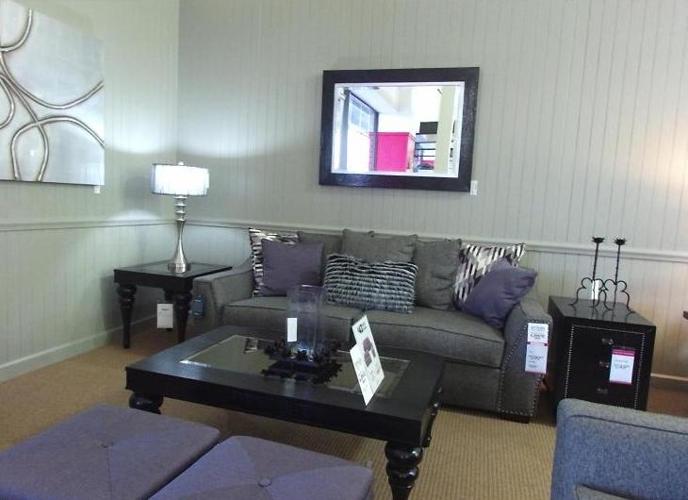}
	\end{subfigure}
	\begin{subfigure}[t]{0.23\textwidth}
		\includegraphics[width=\textwidth]  
		{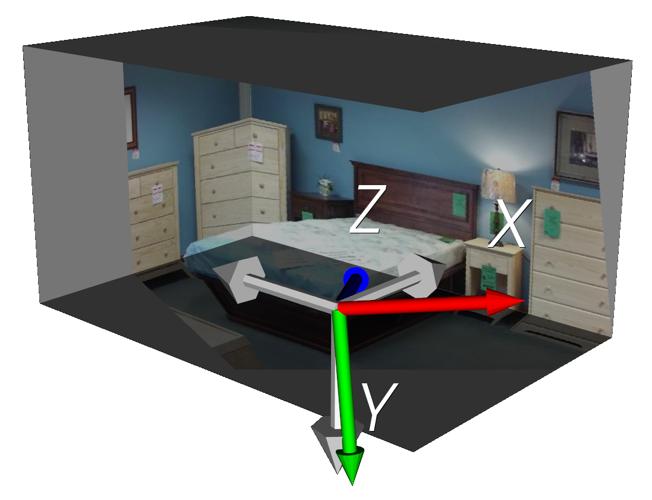}
		\includegraphics[width=\textwidth]
		{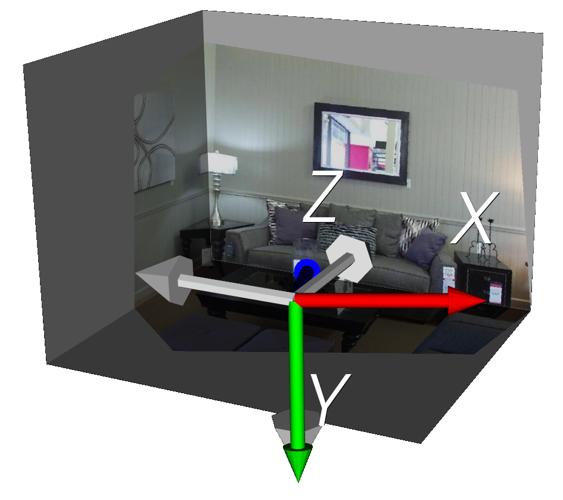}
	\end{subfigure}
	\begin{subfigure}[t]{0.23\textwidth}
		\includegraphics[width=\textwidth]  
		{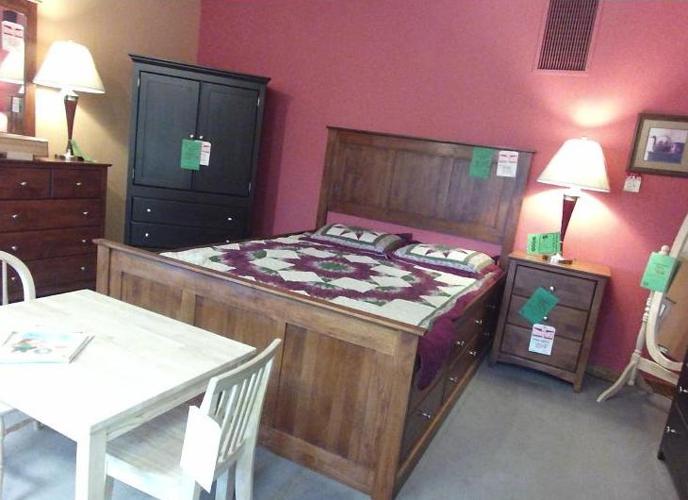}
		\includegraphics[width=\textwidth]
		{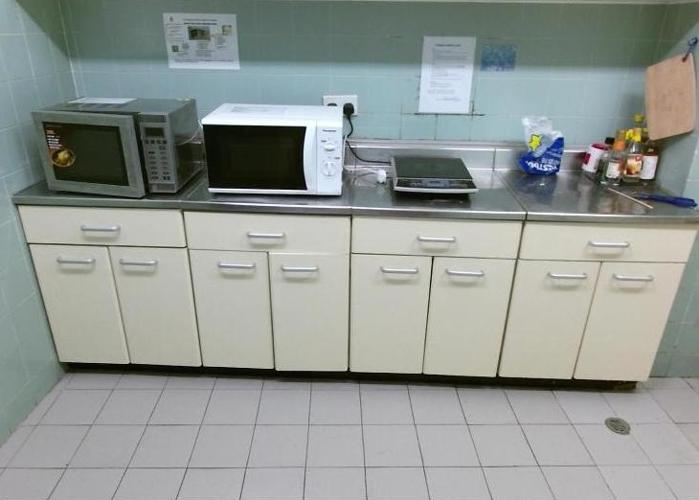}
	\end{subfigure}
	\begin{subfigure}[t]{0.23\textwidth}
		\includegraphics[width=\textwidth]  
		{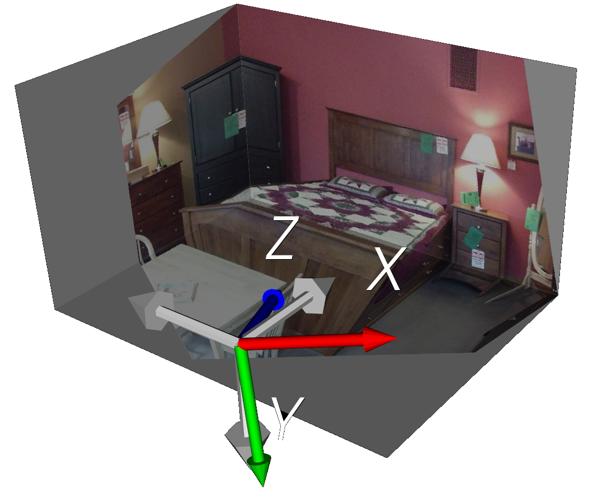}
		\includegraphics[width=\textwidth]
		{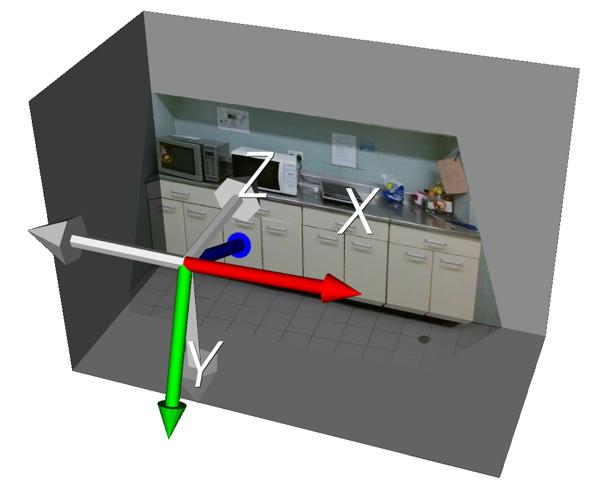}
	\end{subfigure}
	\caption{3D room layout with camera orientation (left: original image, right: 3D layout). The colored arrows represent the camera orientation. The gray arrows respectively point at the floor and walls, which indicates the room layout orientation.}
	\label{fig:3Dlayout}
\end{figure}

To obtain the optimal indoor layout (see Part II in Figure~\ref{fig:layout_estimation}), the line segments that are not located in the layout edge map (high-intensity area) are removed, and we infer more line segments by connecting vanishing points with intersections of line segments from different clusters. More layout proposals can be generated by extensively combining these line segments (see this work \cite{ren2016coarse} for more details). We use the layout edge map to score each pixel in layout proposals and obtain the optimal one with the maximal sum. As the vanishing points provide the room orientation \cite{lu20172}, we fit the indoor layout using a 3D cuboid, with the position of a room corner and layout sizes as optimization variables \cite{hedau2009recovering}. 
Then the camera intrinsic and extrinsic parameters can be estimated. 
Samples of 3D room layout with calibrated cameras are shown in Figure~\ref{fig:3Dlayout}.

\begin{figure}[!ht]
	\centering
	\includegraphics[width=0.7\linewidth]{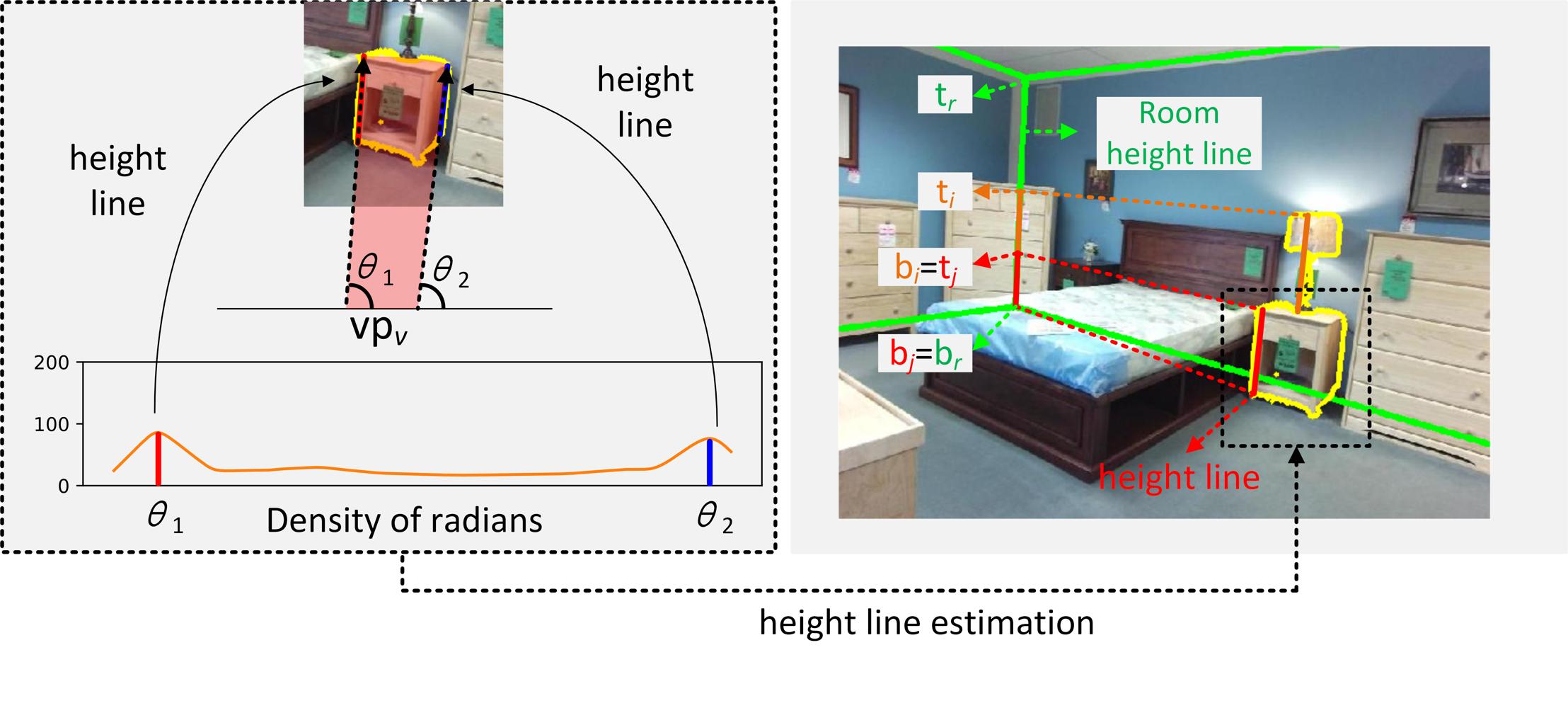}
	\caption{Single-view geometry for object height estimation}
	\label{fig:svg}
\end{figure}

\paragraph{\textbf{Model Initialization}}\label{sec:georet}
Model retrieval (see Section \ref{sec:model_ret}) provides CAD models and orientations for indoor objects. In this part, we introduce single-view geometry combining with support relationship to estimate object sizes and positions with considering object occlusions. The room layout and vanishing points obtained in Section \ref{sec:layout} are used to measure the height of each object. The whole process is illustrated in Figure~\ref{fig:svg}.

Taking the nightstand and lamp in Figure~\ref{fig:svg} as examples, the object $\mathrm{O}_{i}$ (lamp) is supported by $\mathrm{O}_{j}$ (nightstand) from below.
We denote the 2D mask of $\mathrm{O}_{j}$ by $\mathbf{M}_{j}$.
$\mathbf{vp}_{\mathrm{v}}\in\mathcal{R}^{2}$ is the vertical vanishing point on the image plane. For $\mathbf{M}_{j}$, we get its height line by scanning the mask boundary with rays originated from $\mathbf{vp}_{\mathrm{v}}$ (see Figure~\ref{fig:svg}(left)). Each ray connects a pixel on the mask boundary with $\mathbf{vp}_{\mathrm{v}}$. We estimate the Gaussian kernel density of the radian of these rays, and extract the rays whose radian is a local maxima in density. The `local maximal' ray that holds the longest intersection with the mask boundary is selected, and the longest intersection is taken as the optimal height line of $\mathrm{O}_{j}$.

To estimate the real height of objects, we introduce single-view geometry for height measurement (see Figure~\ref{fig:svg}(right)). Specifically, we take the room height line as the reference, and map object's height line onto the reference through the vanishing lines. For $\mathrm{O}_{i}$ (lamp), we denote its top and bottom of the mapped height line by $t_{i}$ and $b_{i}$ respectively. $t_{r}$ and $b_{r}$ respectively indicate the top and the bottom of the room height line. The height of $\mathrm{O}_{i}$ can be calculated by the cross ratio \cite{criminisi2000single}:

\begin{equation}
\label{eqn:03}
\begin{aligned}
H_{i} &= A_{i} - A_{j}, \\
\frac{A_{i}}{H_{r}} &= \frac{\|t_{i}-b_{r}\|}{\|t_{r}-b_{r}\|}\cdot\frac{\|\mathbf{vp}_{\mathrm{v}}-t_{r}\|}{\|\mathbf{vp}_{\mathrm{v}}-t_{i}\|},
\end{aligned}
\end{equation}
where $A_{i}$ and $A_{j}$ respectively denote the top altitude of $\mathrm{O}_{i}$ and $\mathrm{O}_{j}$ (i.e. the real height of $\overrightarrow{t_{i}b_{r}}$ and $\overrightarrow{t_{j}b_{r}}$). $\mathrm{O}_{j}$ is supporting $\mathrm{O}_{i}$ from below. Thus $H_{i}$ is the real height of $\mathrm{O}_{i}$. $H_{r}$ is the real height of the room (i.e. the real height of $\overrightarrow{t_{r}b_{r}}$) and  $\|*\|$ represents the Euclidean distance. We use this formula to recursively get the real height of $\mathrm{O}_{i}$ from the difference between the top altitude of $\mathrm{O}_{i}$ and its supporting object $\mathrm{O}_{j}$. Rather than to address their real height individually, this recursive strategy asks for solving equations following the supporting order. It brings us benefits to verify the support type and solve occlusion problems. For example, the support type should be `support from below' if $H_{i}$ is larger than zero. Moreover, the bottom of an object ($b_{i}$) is usually invisible when it is occluded or not segmented out. While in practice, $b_{i}$ is at the same altitude with $t_{j}$ if $\mathrm{O}_{j}$ is supporting $\mathrm{O}_{i}$ from below. We replace $b_{i}$ with $t_{j}$ in calculations to estimate the real height of each object.

Unlike the `support from below' scenarios where objects are stacked from the floor following the vertical direction, for objects that are supported from behind, the supporting surfaces are not guaranteed with a fixed normal direction. It would be much more complicated to get a closed-form solution. If $\mathrm{O}_{i}$ is supported by walls (like pictures), we can still get an accurate estimate by Equation \ref{eqn:03} (i.e. height difference between $\overrightarrow{t_{i}b_{r}}$ and $\overrightarrow{b_{i}b_{r}}$). While for other cases (e.g. objects are supported by unknown surfaces), we still use this solution to get a rough estimate first. To ensure a reasonable height estimate, we parse the ScanNet \cite{dai2017scannet} to generate a prior height distribution for each object category and replace those unreasonable estimates with the statistically average (see Appendix~\ref{appendix_prior_height} for details).

So far we have obtained the height estimate of each object and its altitude relative to the floor. With the room geometry and the camera parameters obtained in Section \ref{sec:layout}, the 3D location of objects can be estimated using the perspective relation between object masks and its spatial position, we refer readers to this work \cite{choi2015indoor} for more details.

\subsection{Contextual Refinement}
When a room is full of clutter, there could still exist errors in scene initialization, and the aforementioned processes may not be sufficient to solve the scene modeling toward satisfaction. Therefore, a contextual refinement is adopted to fine-tune the CAD models and orientations from candidates (see Section \ref{sec:model_ret}). It refines their initial 3D size and position to make the reconstructed scene consistent in semantic and geometric meaning with the indoor context. 
We formulate this into an optimization problem:
\begin{equation}
\label{eqn:04}
\begin{aligned}\
& \max_{\theta_{i}, \bm{S}_{i}, \bm{O}_{i}, \bm{p}_{i}}\, \mathtt{IoU}\{\mathtt{Proj}[\bm{R}(\theta_{i})\cdot\bm{S}_{i}\cdot\bm{O}_{i}+\bm{p}_{i}],\,\mathbf{M}_{i}\}, \\
& \bm{R}({\theta_{i}})=
\begin{bmatrix}
\cos(\theta_{i}) & -\sin(\theta_{i}) & 0\\
\sin(\theta_{i}) & \cos(\theta_{i}) & 0\\
0 & 0 & 1
\end{bmatrix}, \,
\bm{S}_{i}=
\begin{bmatrix}
s_{i,1} & 0 & 0\\
0 & s_{i,2} & 0\\
0 & 0 & 1
\end{bmatrix}\cdot s_{i,3}, \\
& \bm{p}_{i} = 
\begin{bmatrix}
p_{i,1}, & p_{i,2}, & p_{i,3}
\end{bmatrix}^{\mathrm{T}},\, i=1,2,...,N.
\end{aligned}
\end{equation}
$\bm{O}_{i}$ indicates 3D points in a model candidate of the $i$-th object. All CAD models are initially aligned and placed at the origin of the room coordinate system with the horizontal plane parallel to the floor. $\bm{S}_{i}$ is an anisotropic scaling matrix to control the 3D size of $\bm{O}_{i}$. $ \bm{R}(\theta_{i})$ and $\bm{p}_{i}$ are designed to adjust its orientation and position. $\mathtt{Proj[*]}$ denotes the perspective projection to map coordinates from the room coordinate system to the image plane. $\mathtt{IoU}[*]$ is the Intersection over Union operator. $\mathbf{M}_{i}$ represents the segmented mask of the $i$-th object. Therefore, the target of our contextual refinement is to decide the CAD models $\{\bm{O}_{i}\}$ with orientations $\{\theta_{i}\}$, and adjust their size $\{\bm{S}_{i}\}$ and position $\{\bm{p}_{i}\}$ to make the 2D projections of those reconstructed objects approximate to our segmentation results. $i=1,2,...,N$ and $N$ indicates the count of segmented objects. We implement the scene refinement with a recursive strategy following the support relation constraints.

\paragraph{\textbf{Support constraints from below}}
For $\bm{O}_{i}$ that is supported by $\bm{O}_{j}$ from below, we ask for the geometric center of $\bm{O}_{i}$ falling inside the supporting surface, and the bottom of $\bm{O}_{i}$ attached above the surface:

\begin{subequations}\label{cons:1}
	\begin{align}
	[\bm{R}(\theta_{i})\cdot\bm{S}_{i}\cdot{\bm{O}_{i}}+\bm{p}_{i}]^{\mathrm{c}}_{x,y}&\geqslant\min[\bm{O}_{j}]_{x,y},\\
	[\bm{R}(\theta_{i})\cdot\bm{S}_{i}\cdot{\bm{O}_{i}}+\bm{p}_{i}]^{\mathrm{c}}_{x,y}&\leqslant\max[\bm{O}_{j}]_{x,y},\\ \min[\bm{R}(\theta_{i})\cdot\bm{S}_{i}\cdot{\bm{O}_{i}}+\bm{p}_{i}]_{z|x,y}&\geqslant\max[\bm{O}_{j}]_{z|x,y},
	\end{align}
\end{subequations}

where $[*]^{c}_{x,y}$ indicates the horizontal coordinate $(x,y)$ of the geometric center, and $[*]_{z|x,y}$ is the altitude value at $(x, y)$.

\paragraph{\textbf{Support constraints from behind}} If $\bm{O}_{i}$ is supported by $\bm{O}_{j}$ from behind, we let $\bm{O}_{i}$ to be attached on a side surface of $\bm{O}_{j}$'s bounding box. Thus we do not ask for the orientation of $\bm{O}_{i}$ as it is consistent with the supporting surface. Considering there are four rectangular side surfaces, for each one, we build a local coordinate system $(\bm{o}^{k}_{j},\bm{e}^{k,1}_{j},\bm{e}^{k,2}_{j})$ on a vertex $\bm{o}^{k}_{j}$ and a pair of orthogonal edges $(\bm{e}^{k,1}_{j},\bm{e}^{k,2}_{j})$ on these rectangles. $k\in[1,2,3,4]$ indicates one of the four side surfaces, which is decided by solving the target function (\ref{eqn:04}). Support constraints from behind can be written as:
\begin{subequations}\label{cons:2}
	\begin{align}
	\label{cons:2_1}
	&0\leqslant(\bm{c}_{i}-\bm{o}^{k}_{j})^{\mathrm{T}}\cdot\bm{e}^{k,m}_{j}\leqslant\|\bm{e}^{k,m}_{j}\|^{2},\, m=1,2,\\
	\label{cons:2_2}
	&2(\bm{c}_{i}-\bm{o}^{k}_{j})^{\mathrm{T}}\cdot\bm{n}^{k}_{j}=\mathtt{range}[(\bm{R}(\theta_{i})\cdot\bm{S}_{i}\cdot{\bm{O}_{i}})^{\mathrm{T}}\cdot\bm{n}^{k}_{j}],\\
	\intertext{where}
	\label{cons:2_3}
	&\bm{c}_{i}=[\bm{R}(\theta_{i})\cdot\bm{S}_{i}\cdot{\bm{O}_{i}}+\bm{p}_{i}]^{\mathrm{c}},\\
	\label{cons:2_4}
	&\bm{n}^{k}_{j}=\bm{e}^{k,1}_{j}\times\bm{e}^{k,2}_{j}/\|\bm{e}^{k,1}_{j}\times\bm{e}^{k,2}_{j}\|.
	\end{align}
\end{subequations}
$\bm{c}_{i}$ is the geometric center of the updated $\bm{O}_{i}$.  $\bm{n}^{k}_{j}$ denotes the surface normal (see (\ref{cons:2_3}) and (\ref{cons:2_4})). Hence, (\ref{cons:2_1}) shows that the projection of $\bm{c}_{i}$ along $\bm{n}^{k}_{j}$ should fall inside the supporting surface. $\mathtt{range}[x]$ means $x_{\mathrm{max}}-x_{\mathrm{min}}$. Therefore, (\ref{cons:2_2}) implies that the distance between $\bm{c}_{i}$ and the surface should be a half of the object's size along the direction of $\bm{n}^{k}_{j}$. This is to secure the attachment of $\bm{O}_{i}$ onto the supporting surface. The only difference from constraint (\ref{cons:1}) is that the  optimization of object orientation turns to choosing a correct supporting surface.

\begin{figure*}[h]
	\centering
	\includegraphics[width=\linewidth]{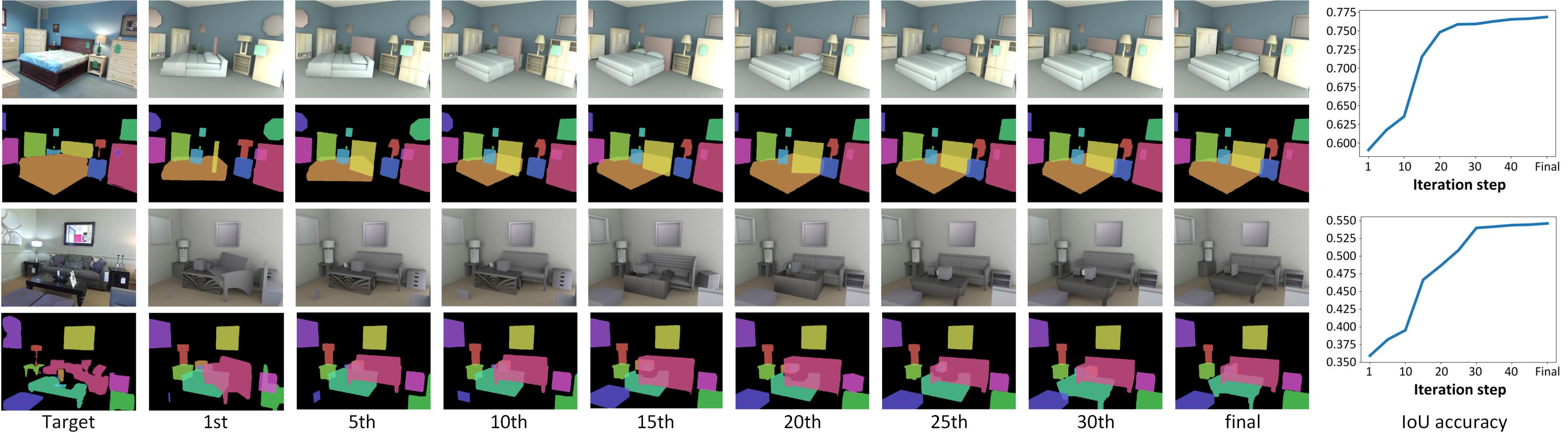}
	\caption{Scene modeling with contextual refinement. The leftmost column presents the original RGB images and the corresponding segmentation. The median part shows the scene modeling results by iterations. The rightmost column illustrates the iteration trajectory of $\mathtt{IoU}$ values correspondingly.}
	\label{fig:convergence}
\end{figure*}

To solve the target function (\ref{eqn:04}), we adopt the exhaustive grid search to decide the exact $\{\bm{O}_{i}\}$ and $\{\theta_{i}\}$. For each grid, BOBYQA method \cite{powell2009bobyqa} is used to refine $\{\bm{S}_{i}\}$ and $\{\bm{p}_{i}\}$. We illustrate the convergence trajectory in Figure~\ref{fig:convergence}. The results demonstrate that the real height of every objects can be initially estimated before iterative refinement, even though there are heavy occlusions or objects that are not fully segmented. From the IoU curve, 30 iterations for model fine-tuning are enough to recover a whole scene.

\section{Experiments and Analysis}
\label{sec:exp_dist}
We present both qualitative and quantitative evaluation of our method with the NYU v2 \cite{silberman2012indoor} and SUN RGB-D dataset \cite{song2015sun}.
All tests are implemented with Python 3.5 on a desktop PC with one TITAN XP GPU and 8 Intel Xeon E5 CPUs.
Parameters and network configurations are detailed in Appendix~\ref{appendix_a}.

\subsection{Performance Analysis}
We record the average time consumption of each phase for 654 testing samples of NYU v2 (see Table~\ref{time_cons}).
The time cost in modeling a whole scene is related to its complexity.
It is expected that modeling a cluttered room with more items costs more time.
Object-specific tasks (segmentation, model retrieval) are processed in parallel.
On average, it takes 2-3 minutes to process a indoor room of reasonable complexity containing up to 20 detected objects.

\begin{table}[ht]
	\begin{center}
		\caption{Average time consumption (in seconds) of (1) 2D segmentation + DCRF refining, (2) model retrieval, (3) support inference, (4) camera-layout joint estimation, (5) model initialization and (6) scene modeling. 30 iterations are used in the contextual refinement, and the average number of detected objects is 16 over the 654 testing images.}
		\label{time_cons}
		\begin{tabular}{l c c c c c c c}
			\hline
			Phase             & (1)  & (2)   & (3) & (4) & (5) & (6) & Total\\
			\hline
			Time elapsed      & 9.87 & 9.72 & 2.08 & 25.53 & 0.95 & 69.68 & 117.84\\
			\hline
		\end{tabular}
	\end{center}
\end{table}

\begin{figure}[!ht]
	\centering
	\begin{subfigure}[t]{0.15\textwidth}
		\includegraphics[width=\textwidth]  
		{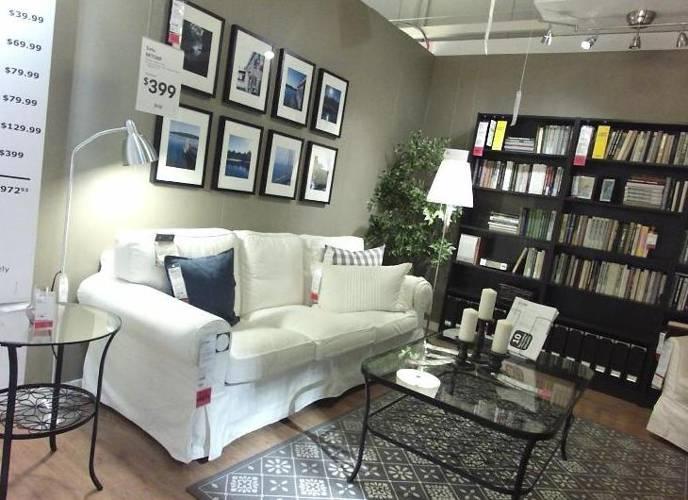}
		\includegraphics[width=\textwidth]
		{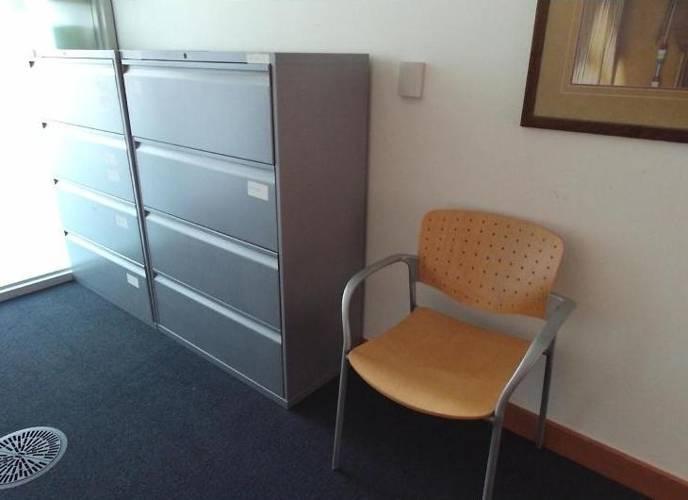}
		\includegraphics[width=\textwidth]
		{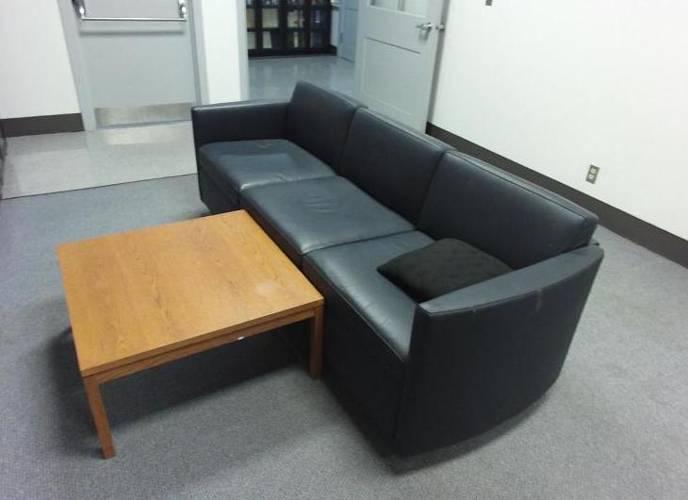}
		\includegraphics[width=\textwidth]
		{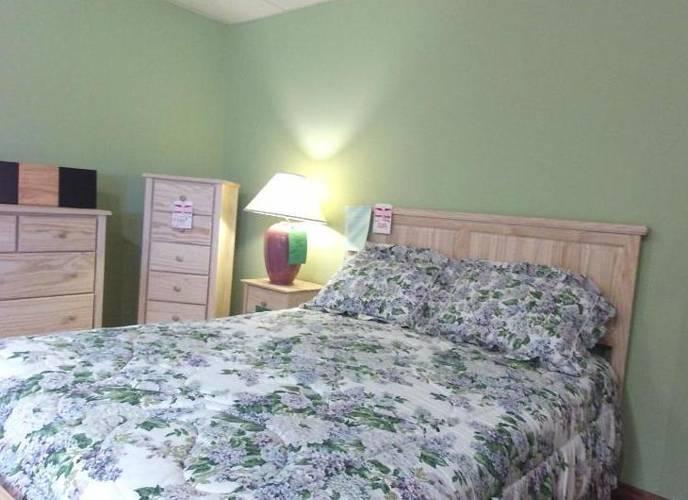}
		\includegraphics[width=\textwidth]
		{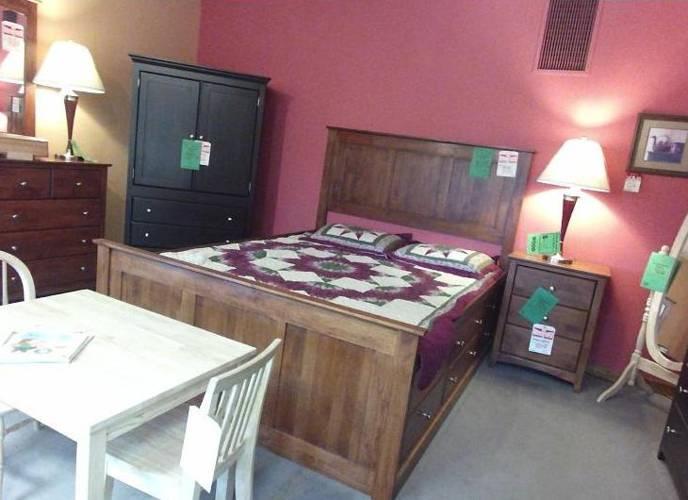}
		\includegraphics[width=\textwidth]
		{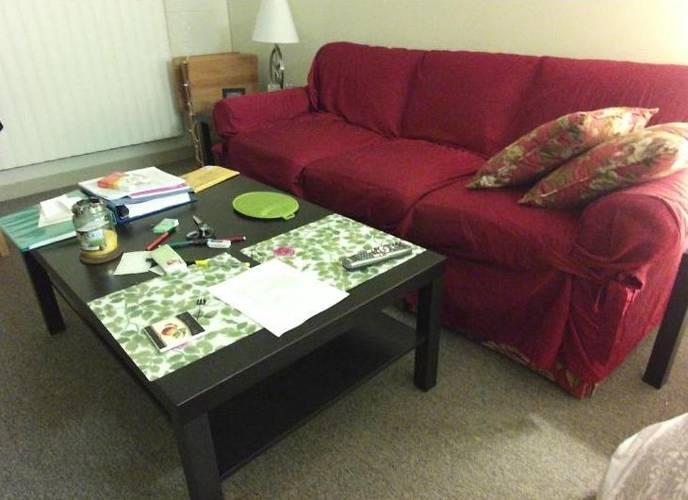}
		\includegraphics[width=\textwidth]
		{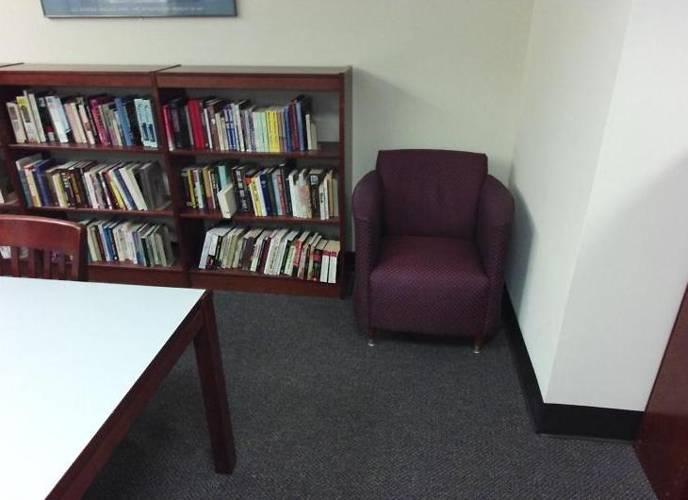}
	\end{subfigure}
	\begin{subfigure}[t]{0.15\textwidth}
		\includegraphics[width=\textwidth]  
		{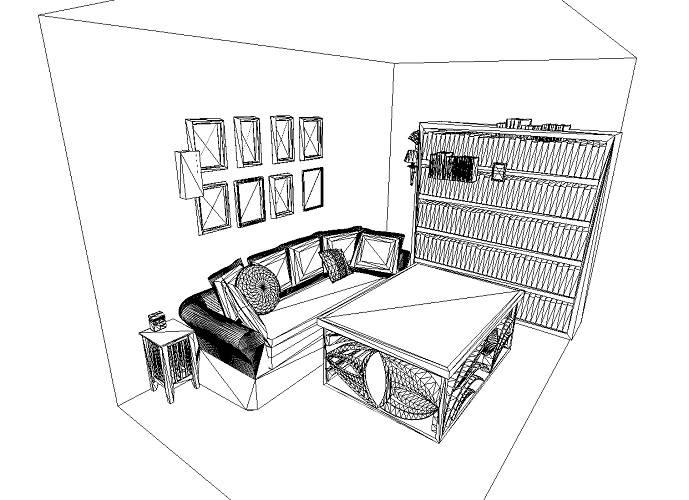}
		\includegraphics[width=\textwidth]
		{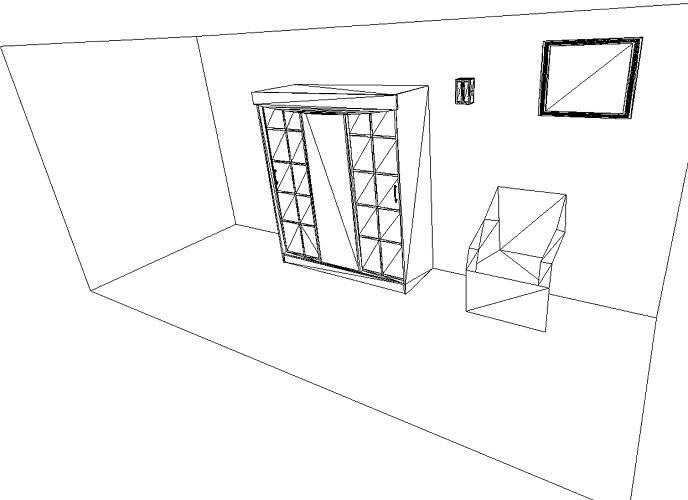}
		\includegraphics[width=\textwidth]
		{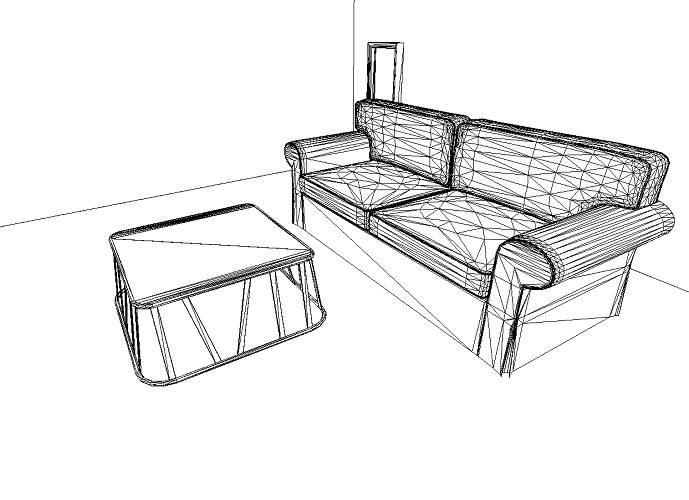}
		\includegraphics[width=\textwidth]
		{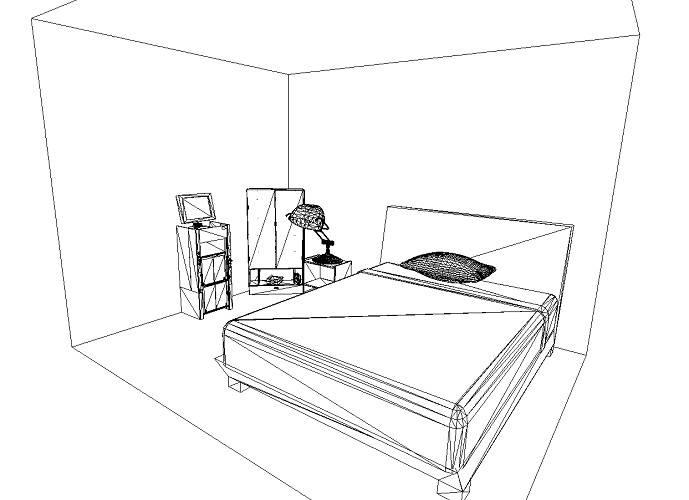}
		\includegraphics[width=\textwidth]
		{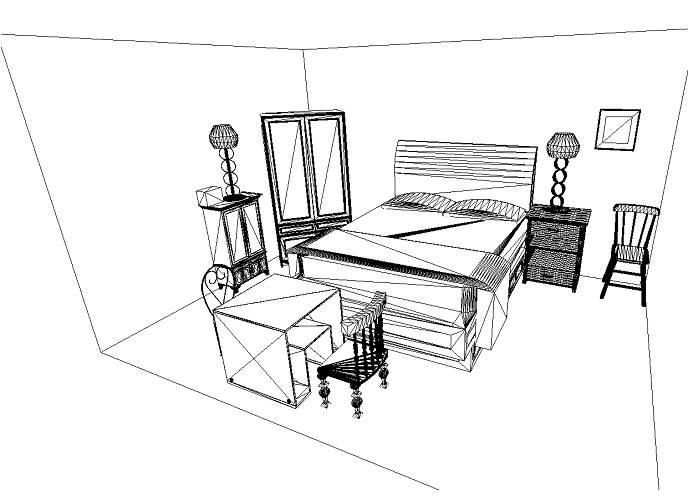}
		\includegraphics[width=\textwidth]
		{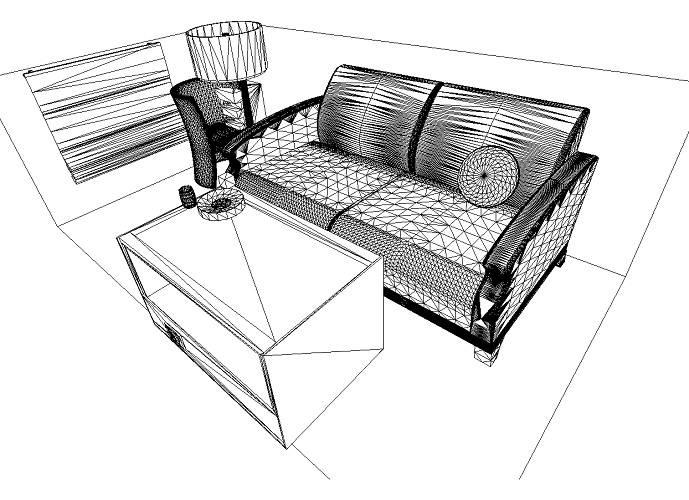}
		\includegraphics[width=\textwidth]
		{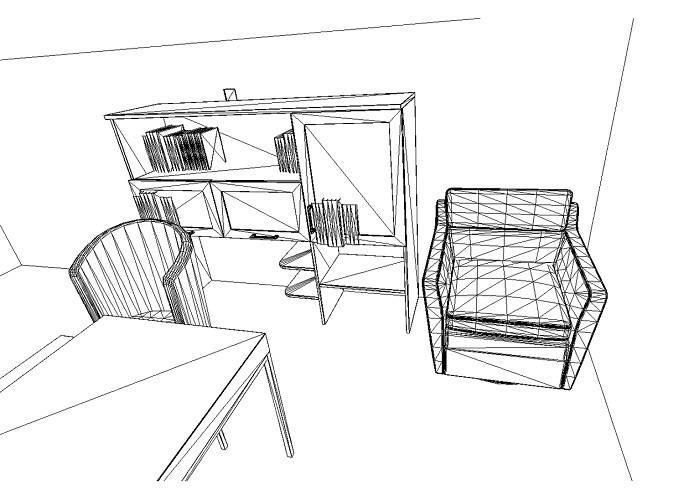}
	\end{subfigure}
	\begin{subfigure}[t]{0.15\textwidth}
		\includegraphics[width=\textwidth]  
		{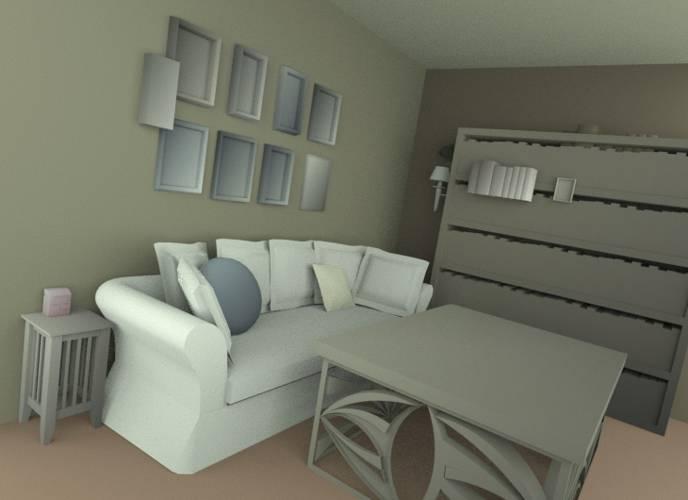}
		\includegraphics[width=\textwidth]
		{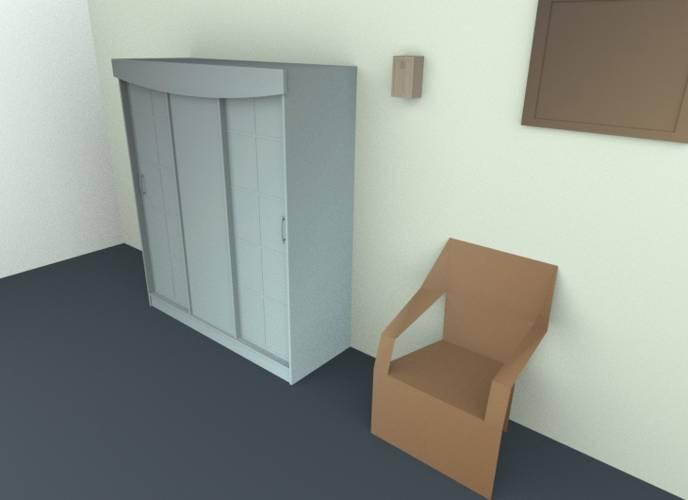}
		\includegraphics[width=\textwidth]
		{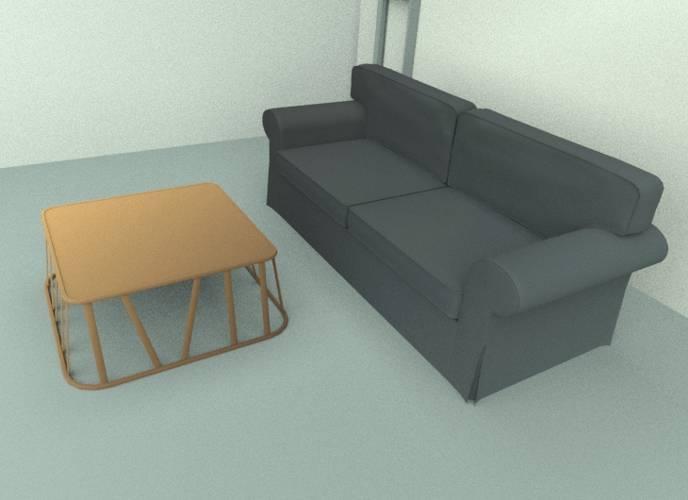}
		\includegraphics[width=\textwidth]
		{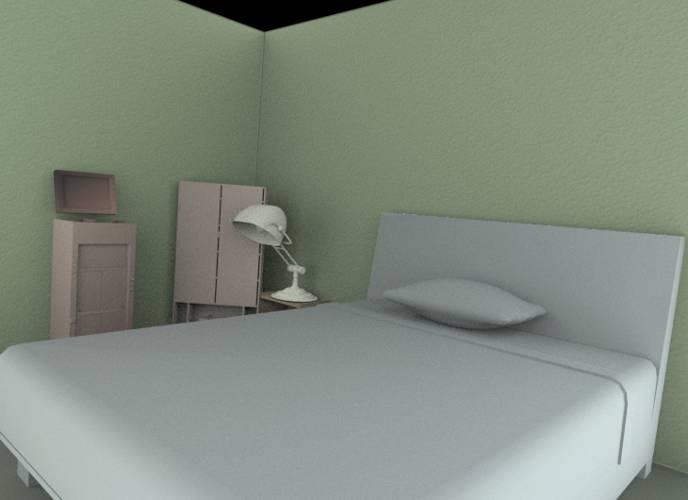}
		\includegraphics[width=\textwidth]
		{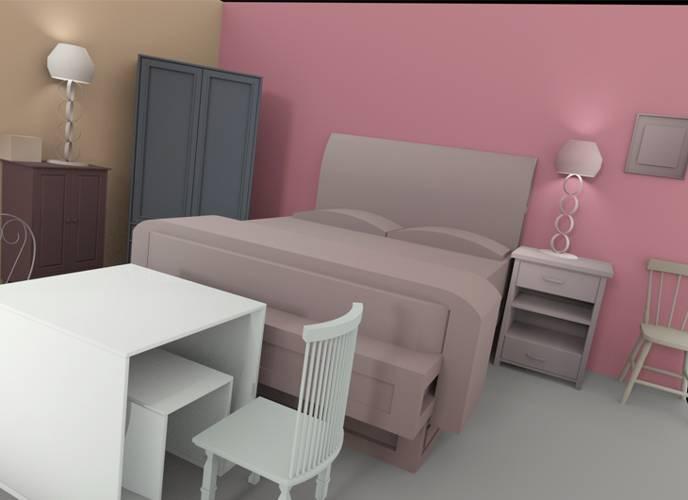}
		\includegraphics[width=\textwidth]
		{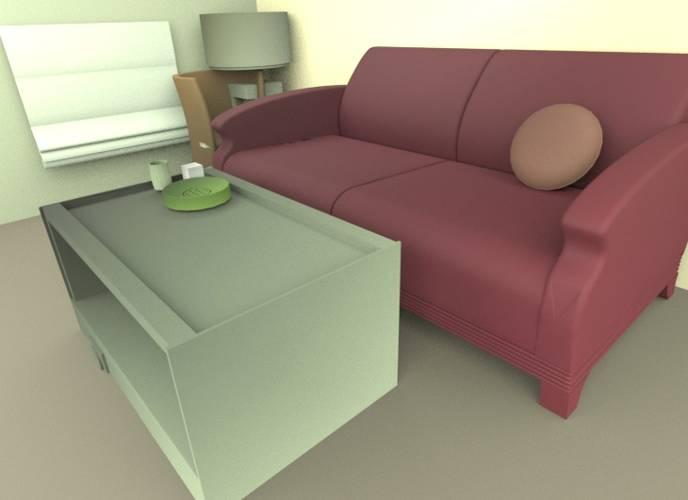}
		\includegraphics[width=\textwidth]
		{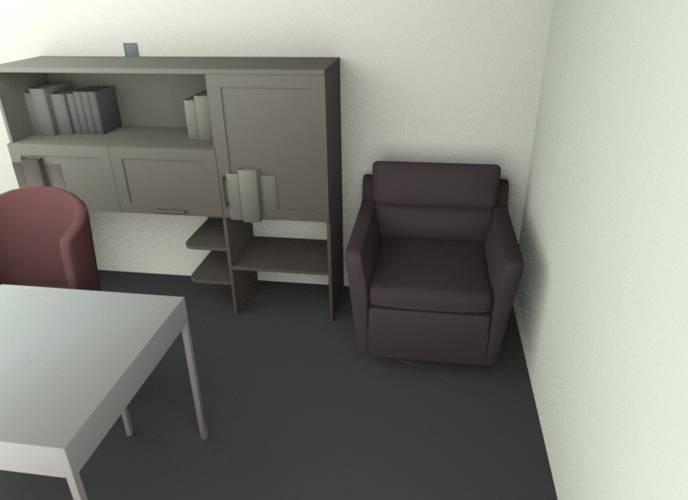}
	\end{subfigure}
	\rulesep
	\begin{subfigure}[t]{0.15\textwidth}
		\includegraphics[width=\textwidth]  
		{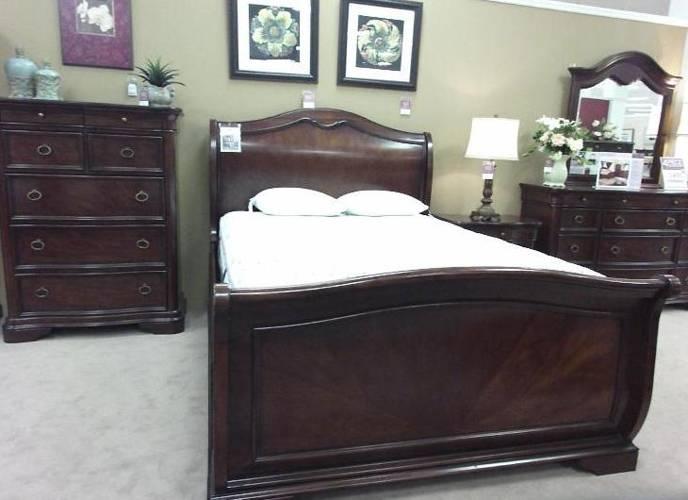}
		\includegraphics[width=\textwidth]
		{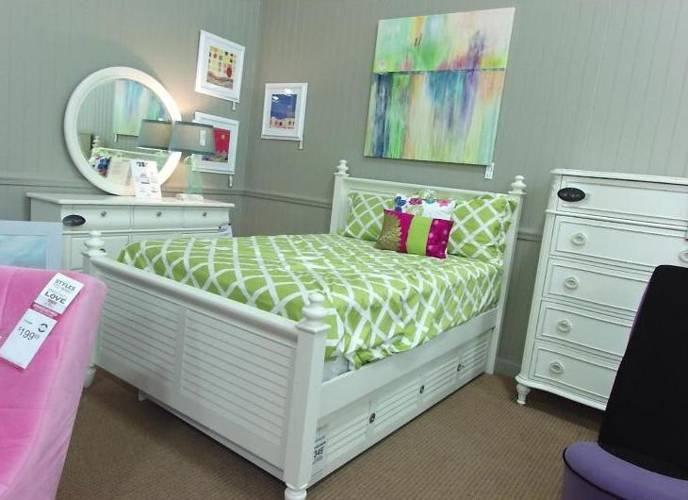}
		\includegraphics[width=\textwidth]
		{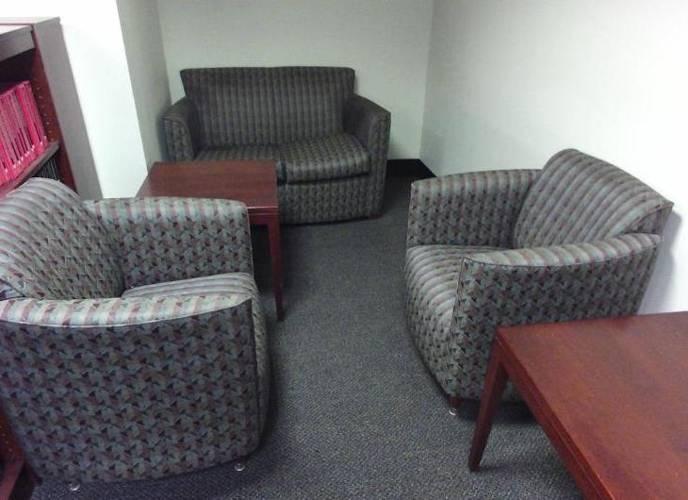}
		\includegraphics[width=\textwidth]
		{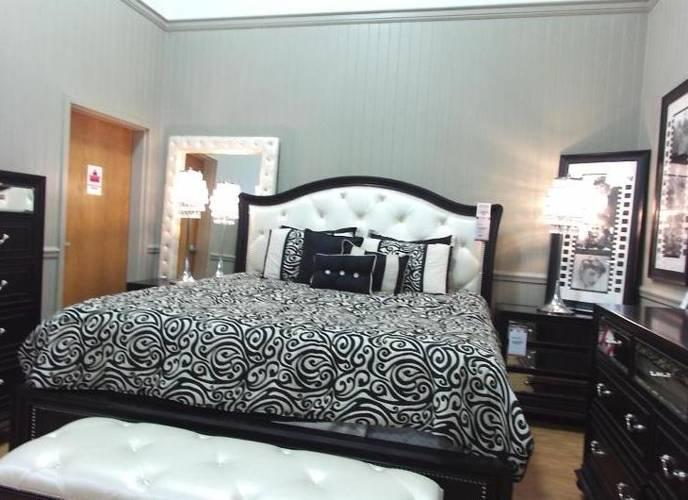}
		\includegraphics[width=\textwidth]
		{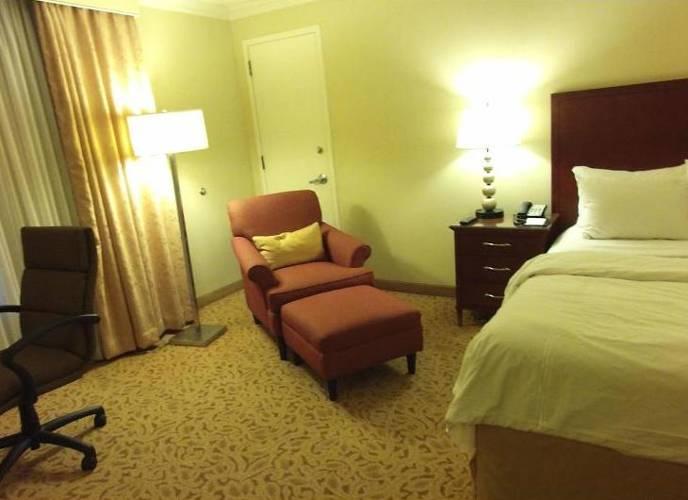}
		\includegraphics[width=\textwidth]  
		{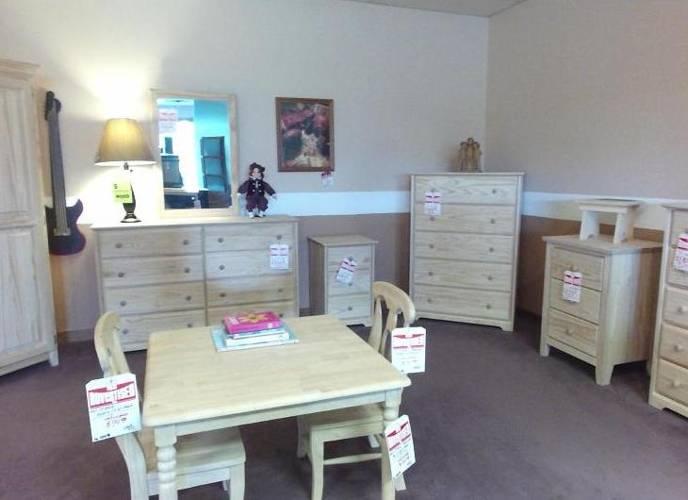}
		\includegraphics[width=\textwidth]
		{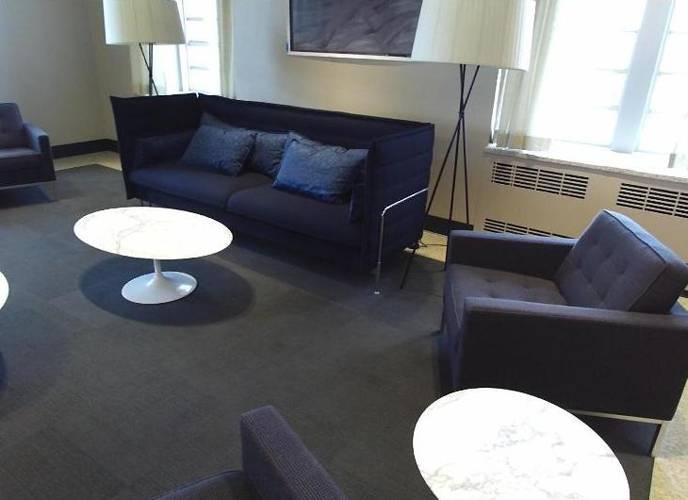}
	\end{subfigure}
	\begin{subfigure}[t]{0.15\textwidth}
		\includegraphics[width=\textwidth]
		{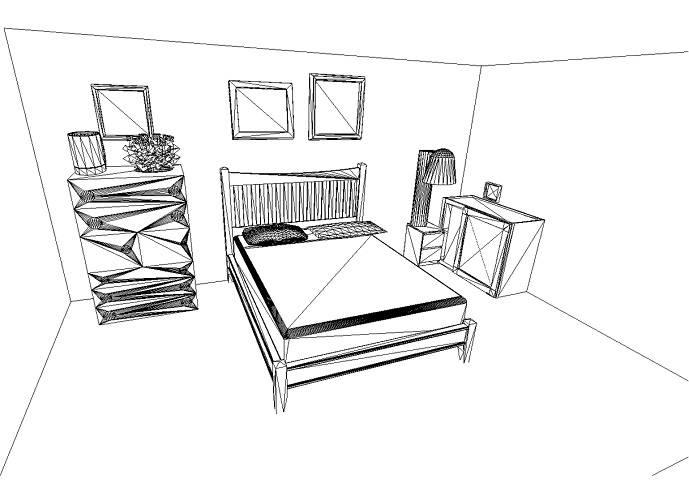}
		\includegraphics[width=\textwidth]  
		{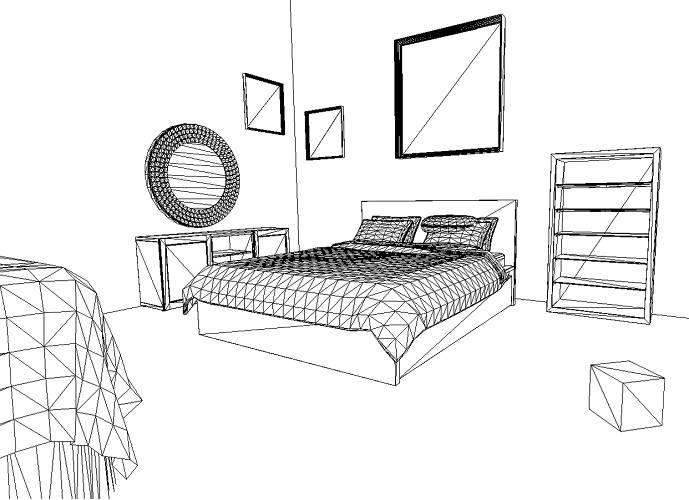}
		\includegraphics[width=\textwidth]
		{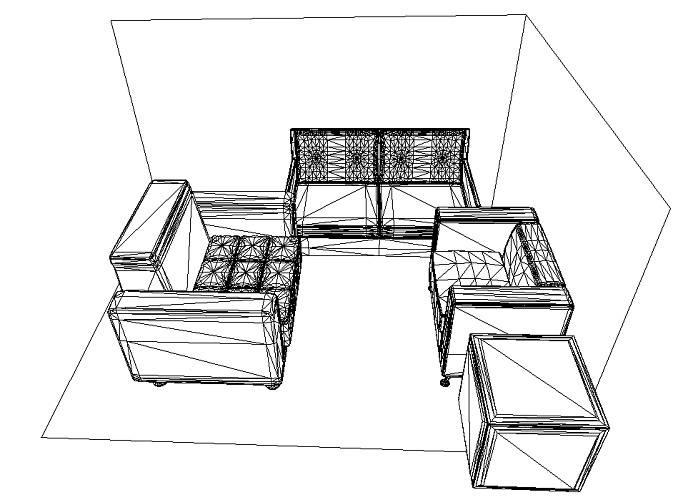}
		\includegraphics[width=\textwidth]
		{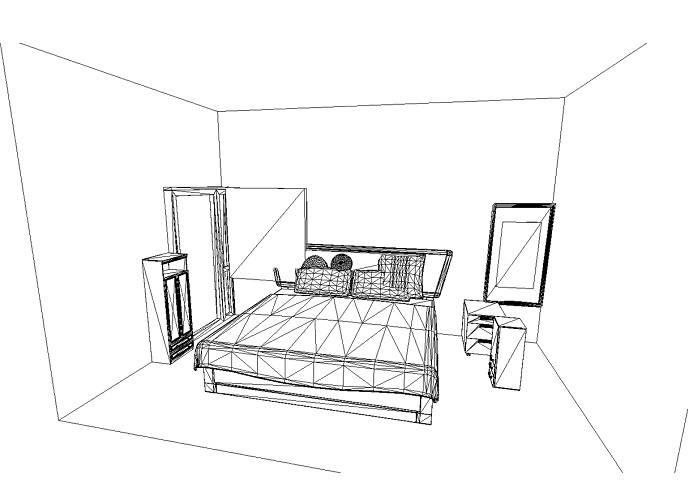}
		\includegraphics[width=\textwidth]
		{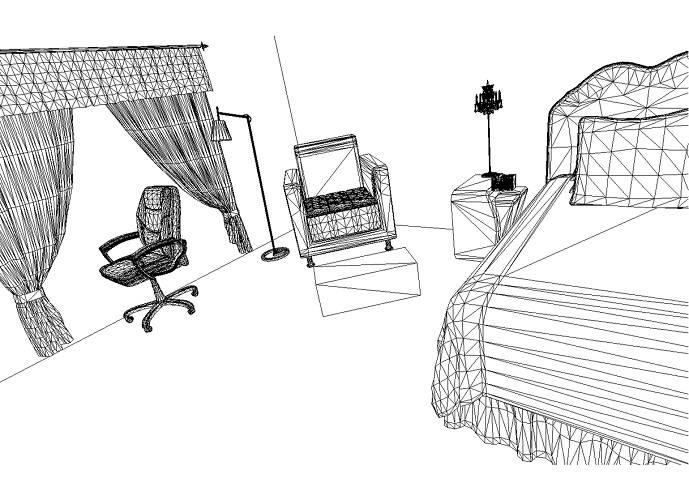}
		\includegraphics[width=\textwidth]  
		{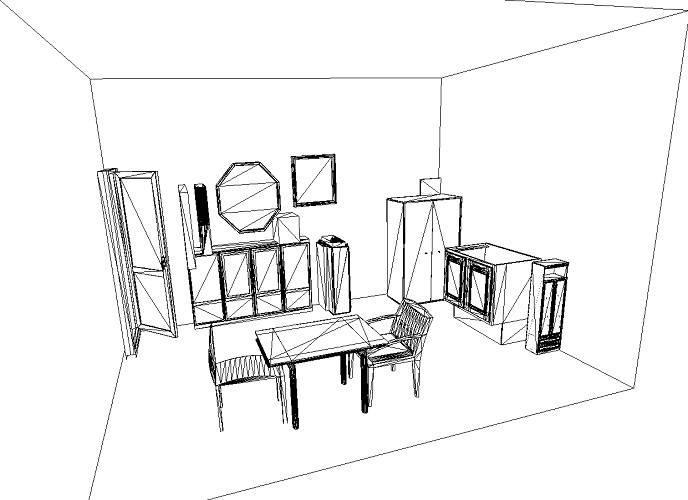}
		\includegraphics[width=\textwidth]
		{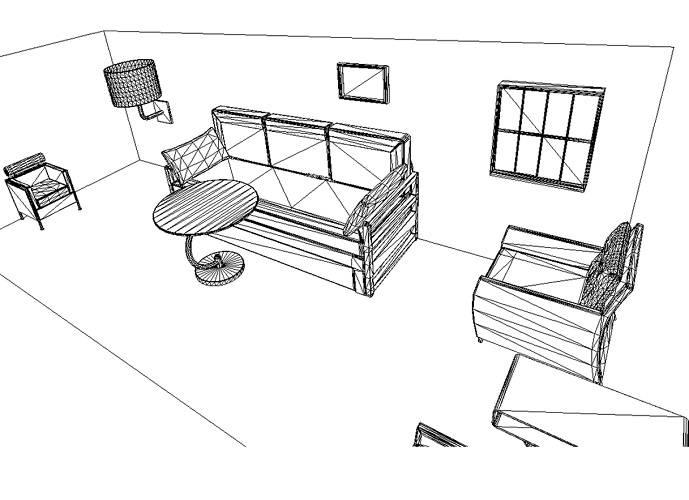}
	\end{subfigure}
	\begin{subfigure}[t]{0.15\textwidth}
		\includegraphics[width=\textwidth]
		{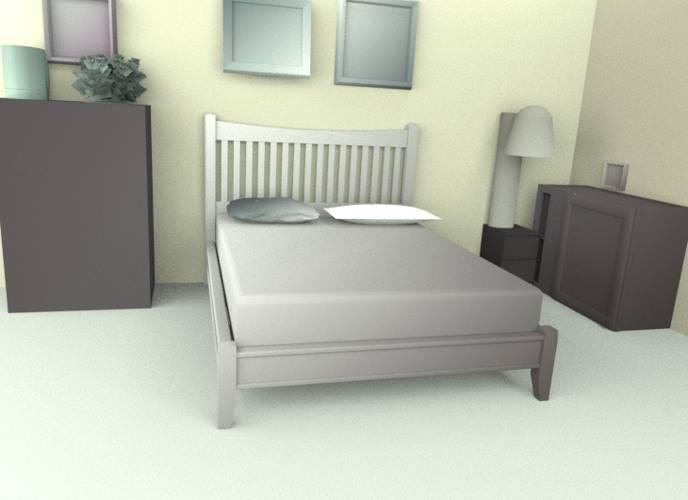}
		\includegraphics[width=\textwidth]  
		{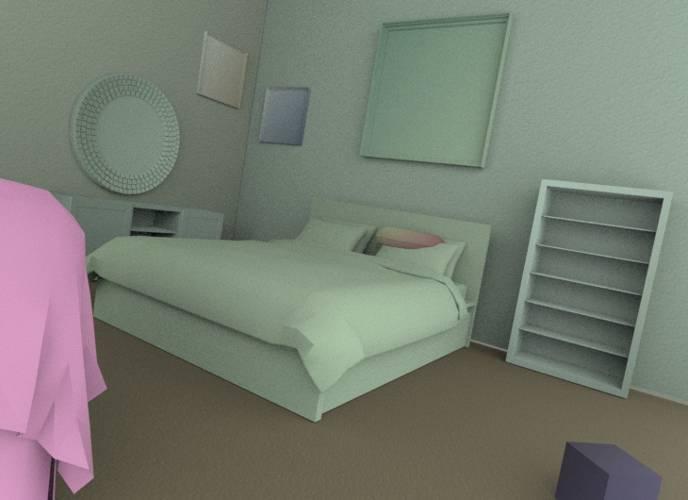}
		\includegraphics[width=\textwidth]
		{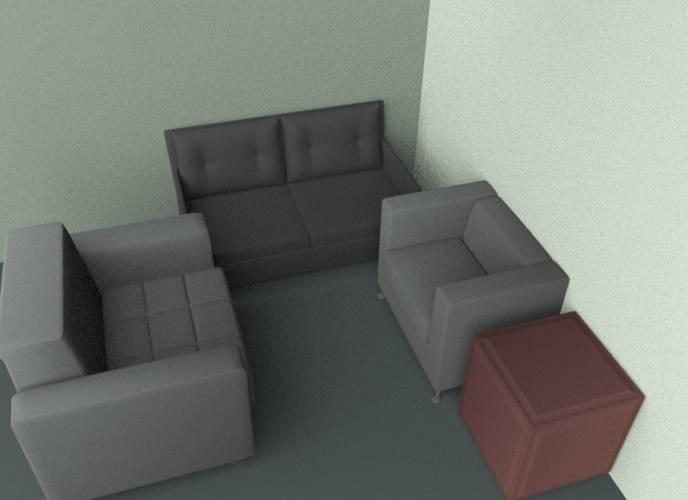}
		\includegraphics[width=\textwidth]
		{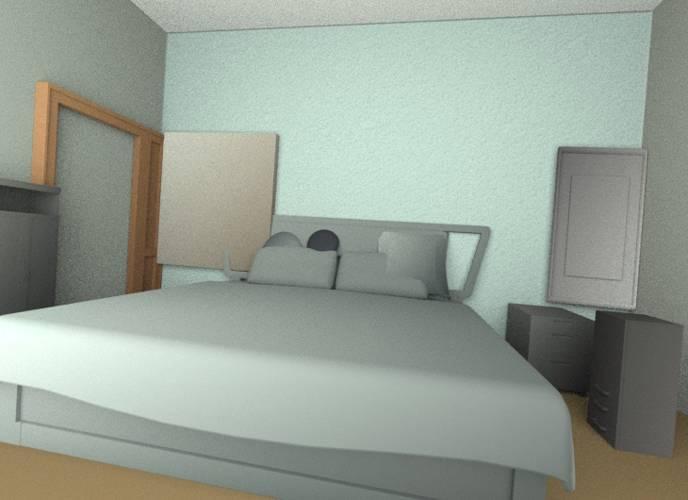}
		\includegraphics[width=\textwidth]
		{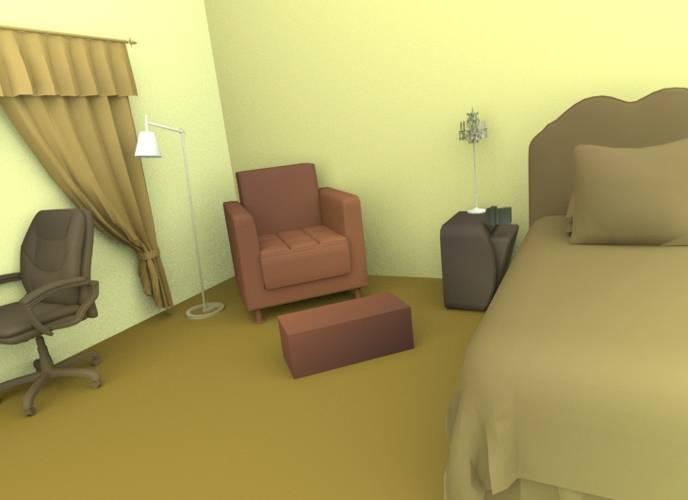}
		\includegraphics[width=\textwidth]  
		{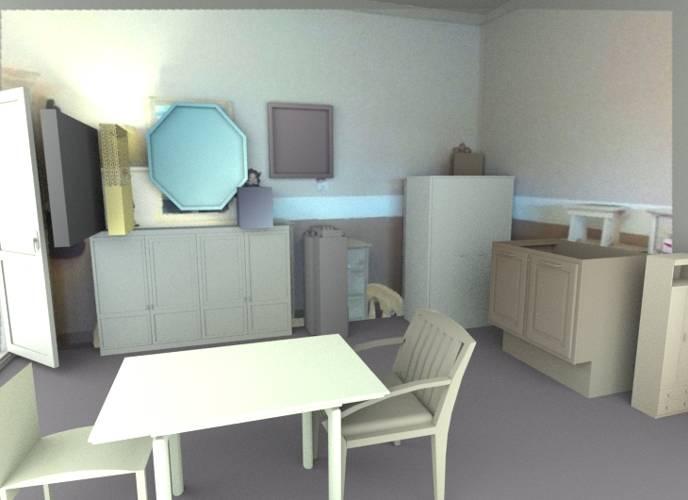}
		\includegraphics[width=\textwidth]
		{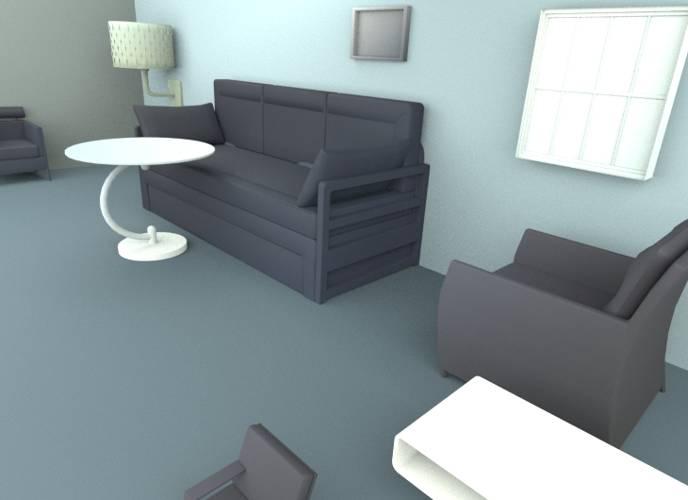}
	\end{subfigure}
	\caption{Scene modeling samples on the SUN RGB-D dataset. Each sample consists of an original image (left), the reconstructed scene (raw mesh, middle) and the rendered scene with our estimated camera parameters (right).}
	\label{fig:sceneret}
\end{figure}

\subsection{Qualitative Evaluation}
Figure~\ref{fig:sceneret} illustrates part of modeling results with different room types and various complexity (randomly picked from the SUN RGB-D dataset, see intermediate results and more samples in Appendix~\ref{appendix_results}). The results demonstrate that the detected objects are organized to make the overall presentation consistent with the original images (e.g., object orientation, position and support relationships). The same camera model as the one estimated from each input image is used in rendering, showing both the room layout and camera are reliably recovered with our joint estimation. 
Benefited from the robust support inference, objects that are heavily occluded or partly visible in the image are predicted with a plausible size.

\begin{figure}[!ht]
	\centering
	\begin{subfigure}[t]{0.155\textwidth}
		\includegraphics[width=\textwidth]  
		{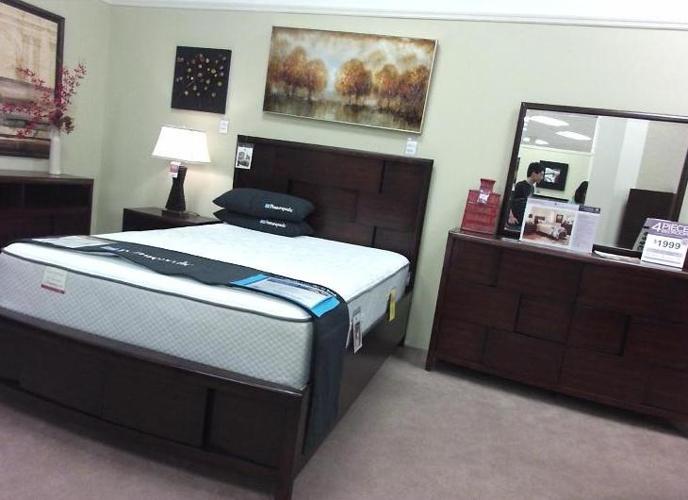}
		\includegraphics[width=\textwidth]
		{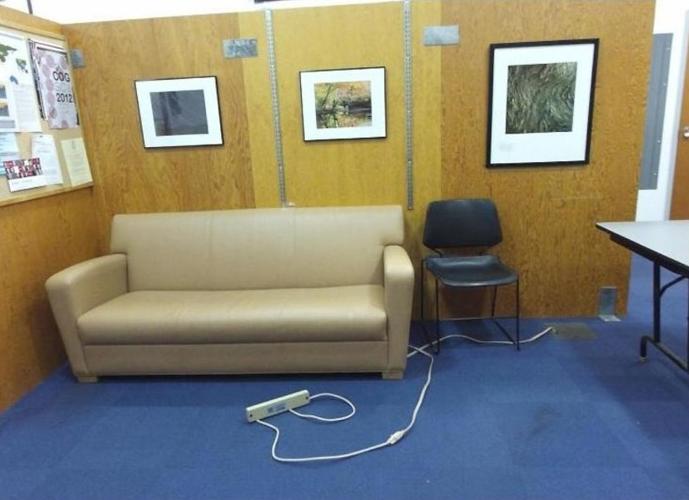}
		\includegraphics[width=\textwidth]
		{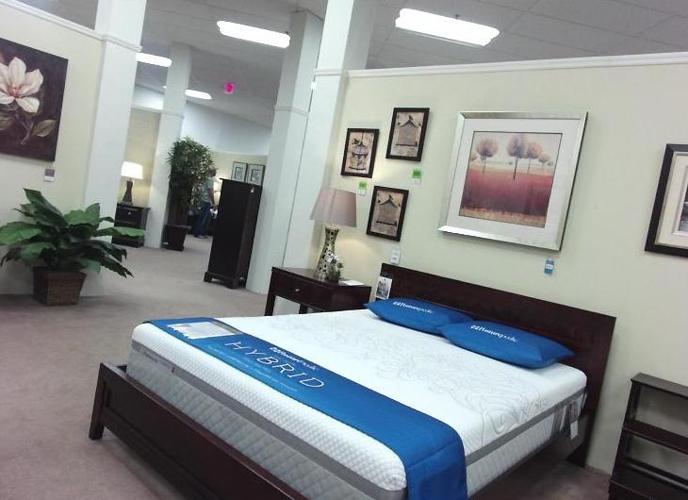}
		\includegraphics[width=\textwidth]
		{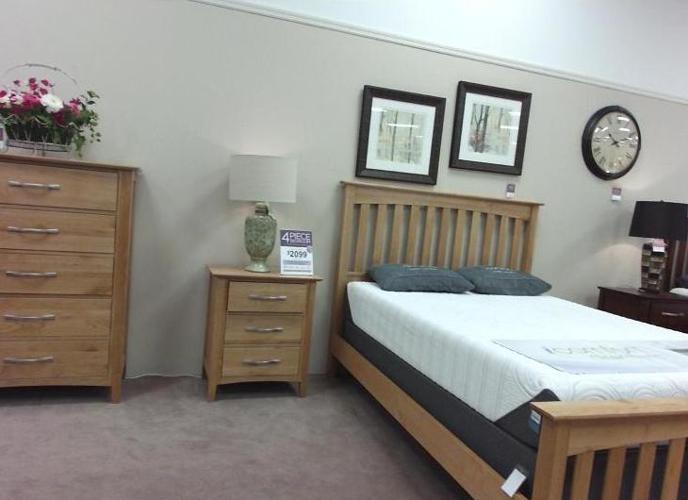}
		\includegraphics[width=\textwidth]
		{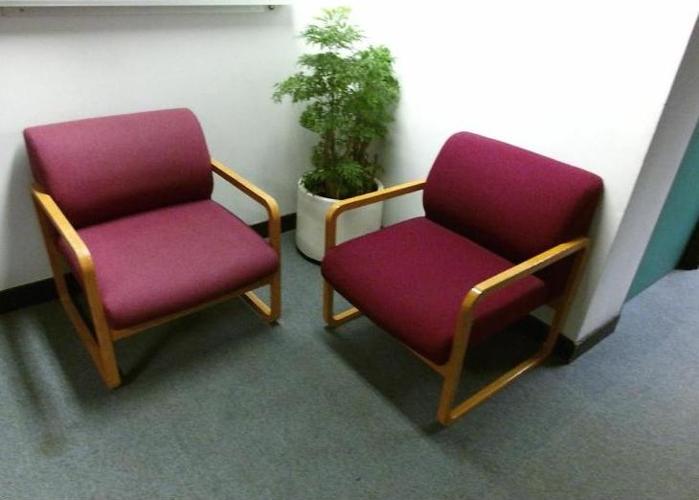}
		\includegraphics[width=\textwidth]
		{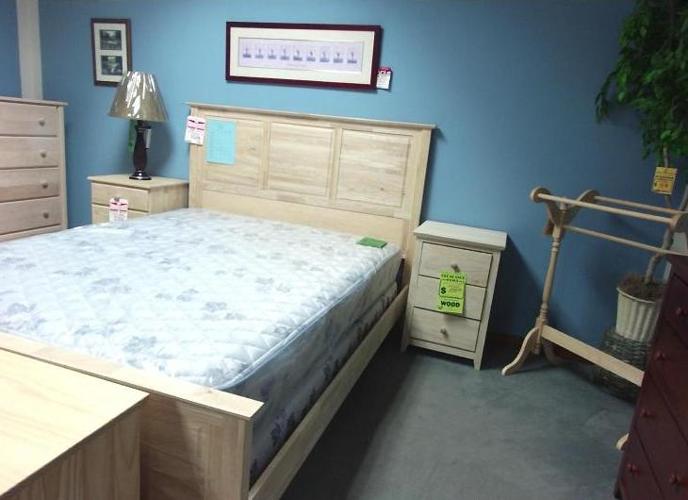}
		\includegraphics[width=\textwidth]
		{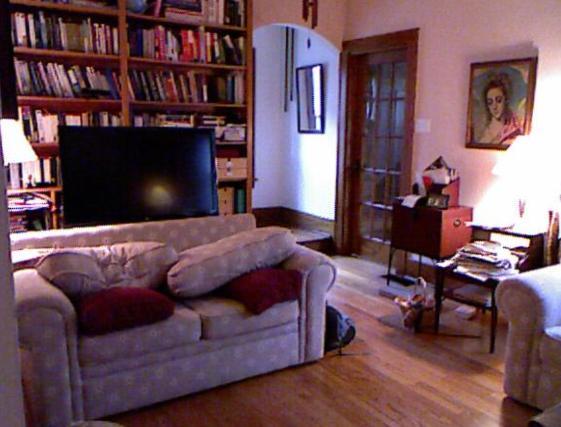}
		\caption{}
		\label{comp:c1}
	\end{subfigure}
	\begin{subfigure}[t]{0.155\textwidth}
		\includegraphics[width=\textwidth]  
		{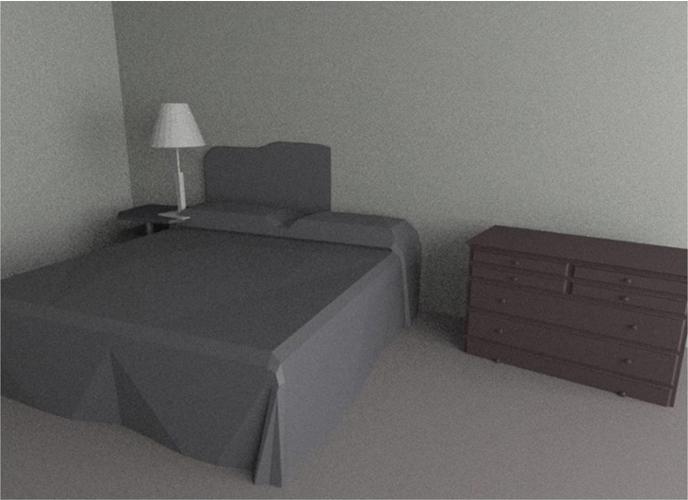}
		\includegraphics[width=\textwidth]
		{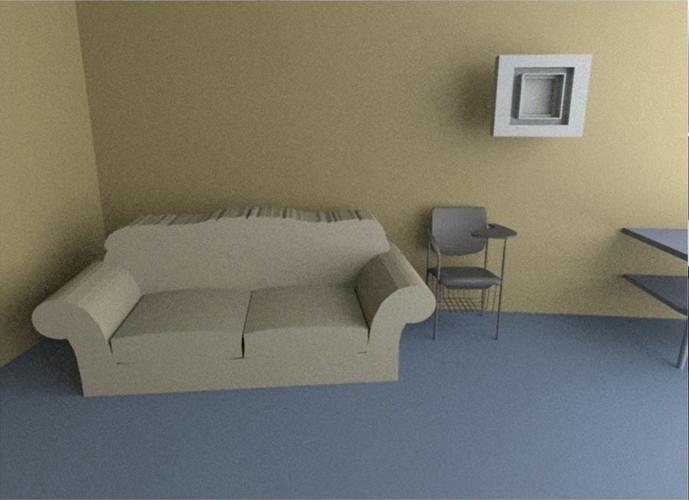}
		\includegraphics[width=\textwidth]
		{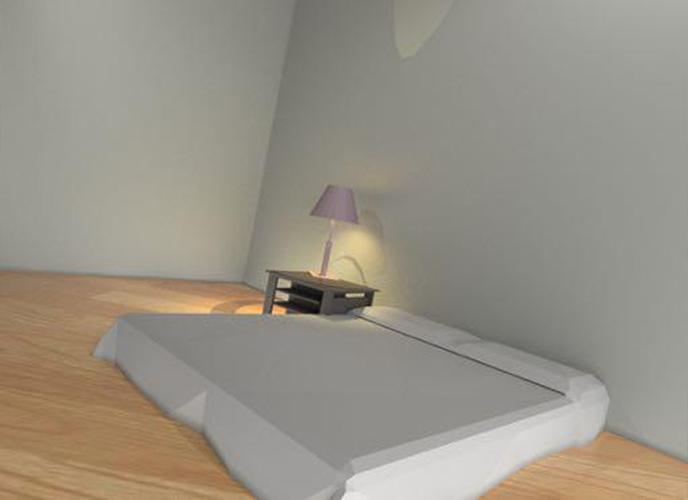}
		\includegraphics[width=\textwidth]
		{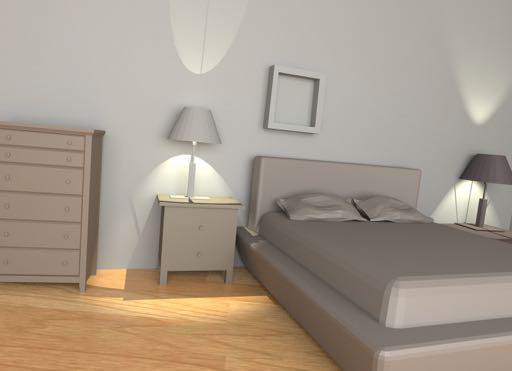}
		\includegraphics[width=\textwidth]
		{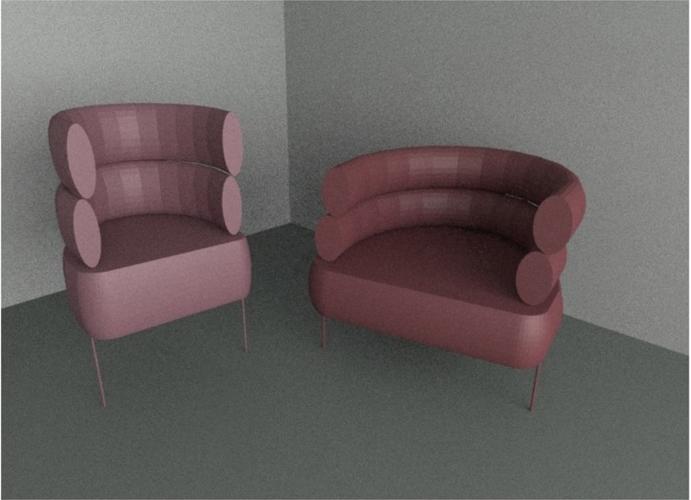}
		\includegraphics[width=\textwidth]
		{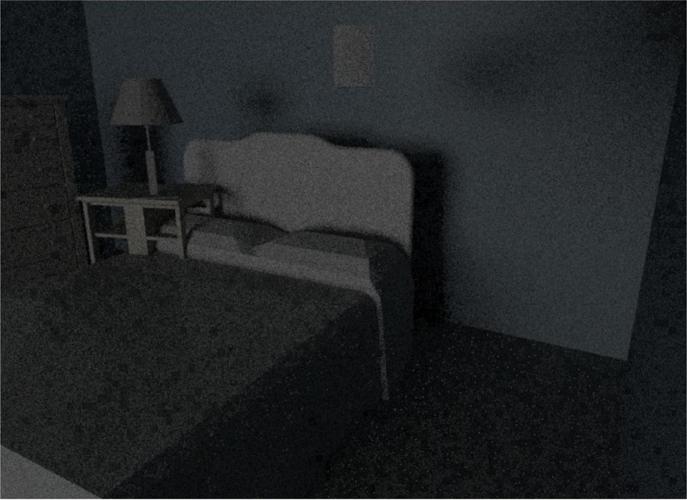}
		\includegraphics[width=\textwidth]
		{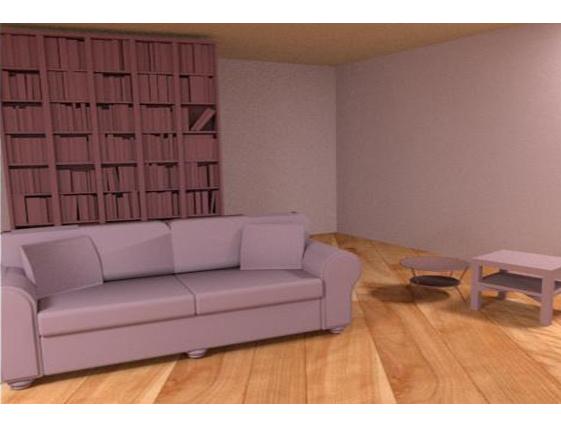}
		\caption{}
	\end{subfigure}
	\begin{subfigure}[t]{0.155\textwidth}
		\includegraphics[width=\textwidth]  
		{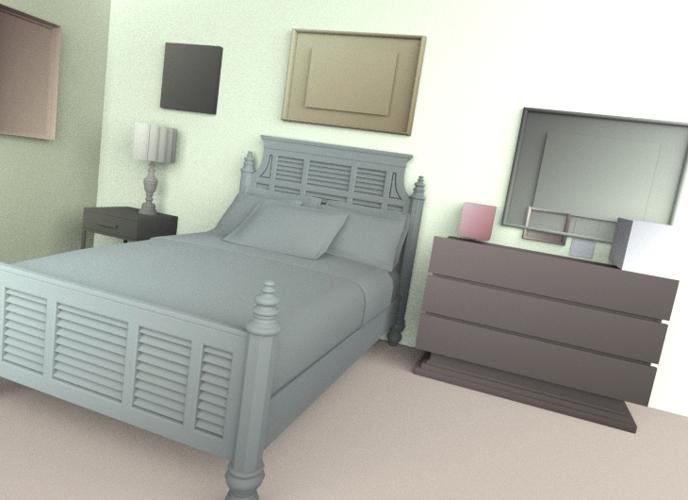}
		\includegraphics[width=\textwidth]
		{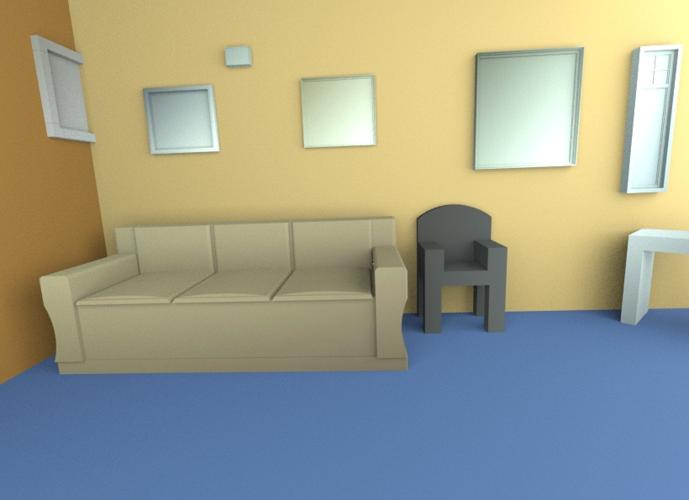}
		\includegraphics[width=\textwidth]
		{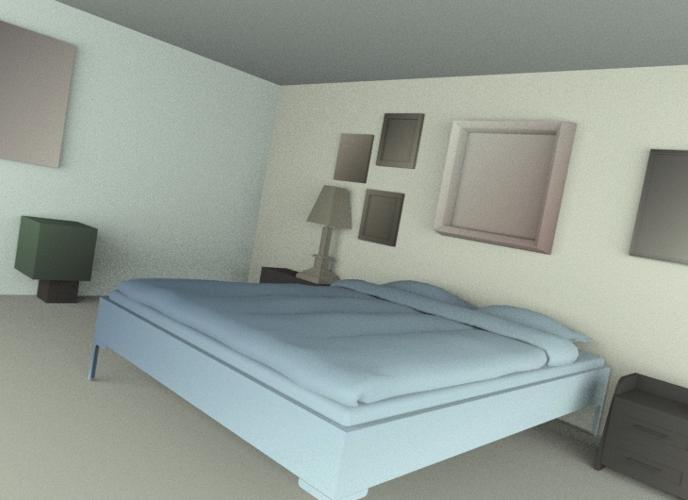}
		\includegraphics[width=\textwidth]
		{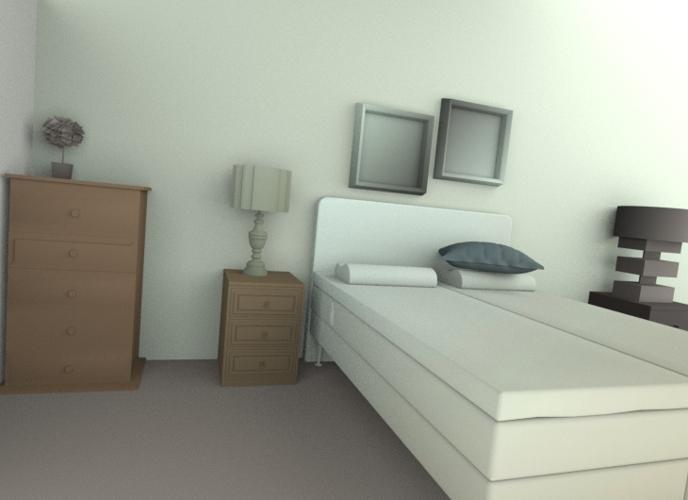}
		\includegraphics[width=\textwidth]
		{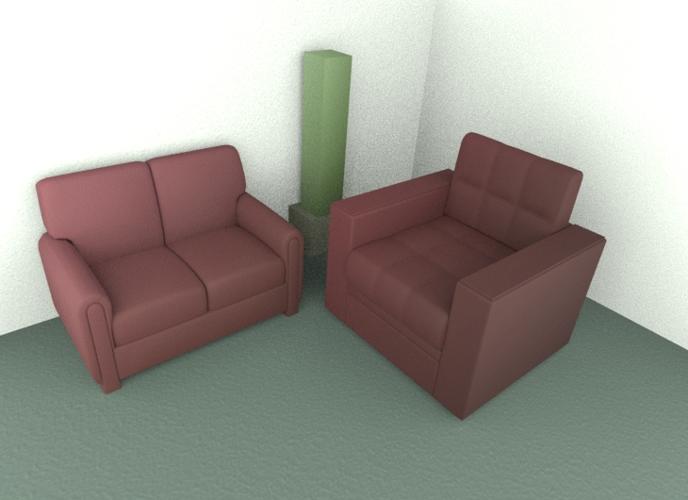}
		\includegraphics[width=\textwidth]
		{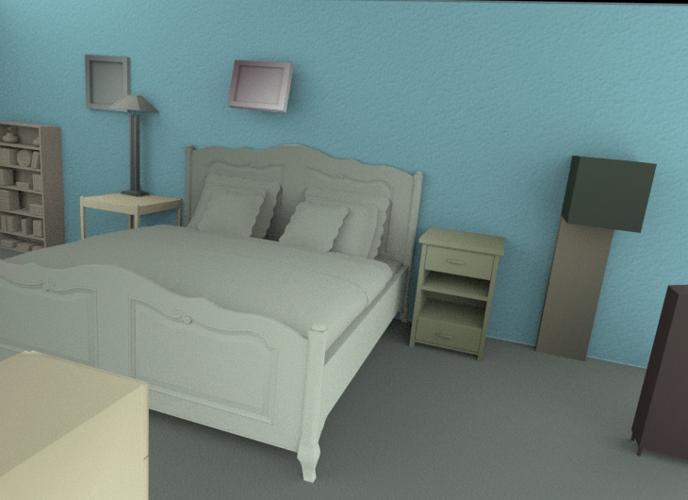}
		\includegraphics[width=\textwidth]
		{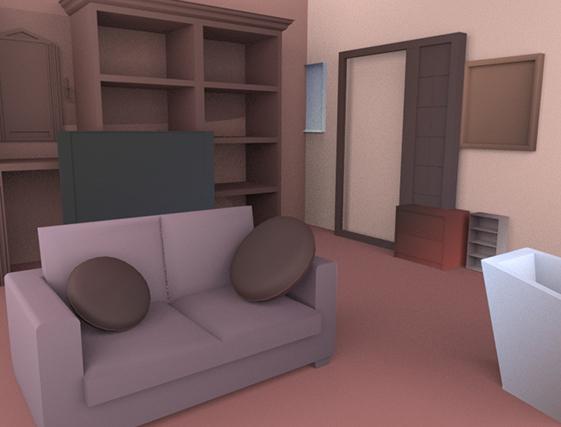}
		\caption{}
	\end{subfigure}\rulesep
	\begin{subfigure}[t]{0.155\textwidth}
		\includegraphics[width=\textwidth]
		{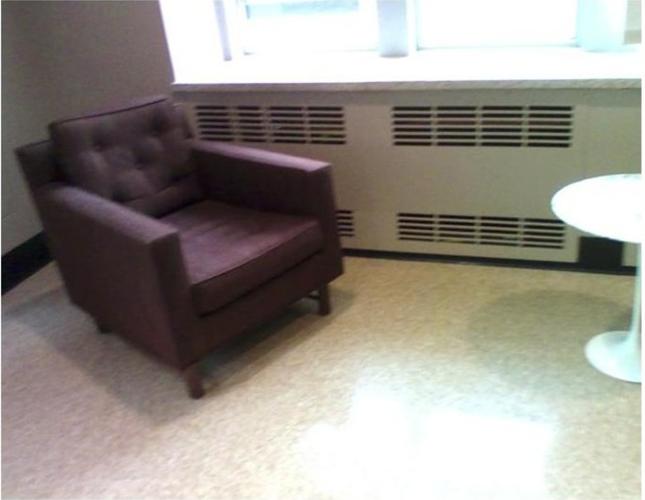}
		\includegraphics[width=\textwidth]  
		{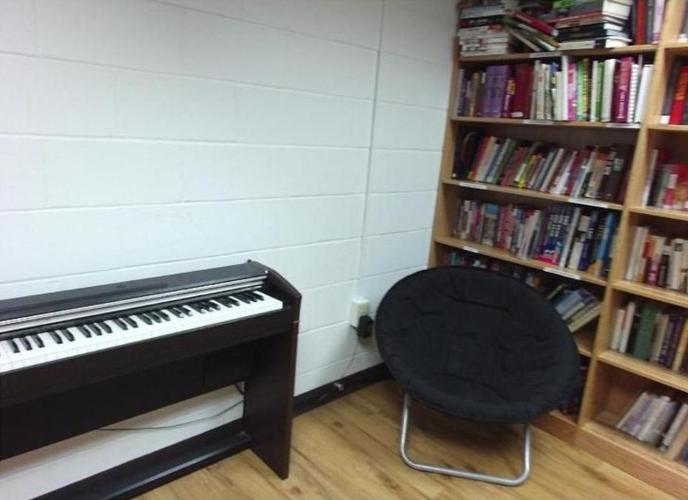}
		\includegraphics[width=\textwidth]
		{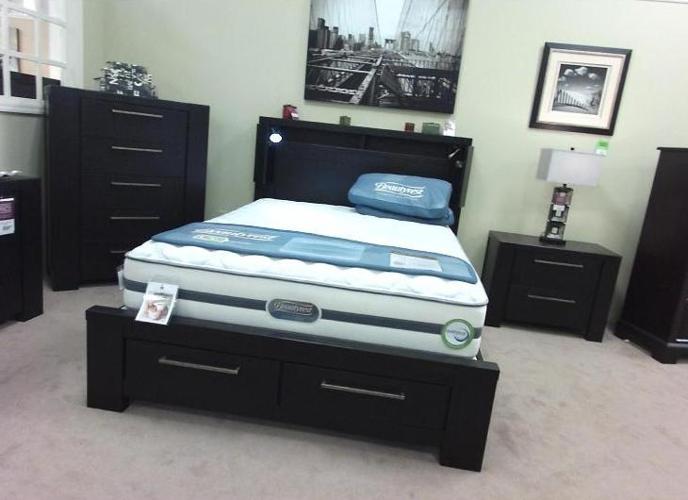}
		\includegraphics[width=\textwidth]
		{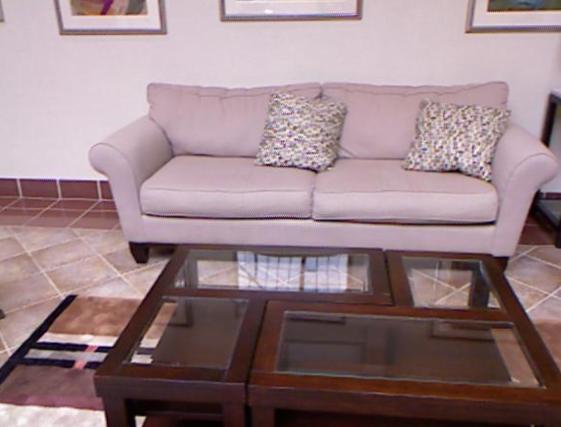}
		\includegraphics[width=\textwidth]
		{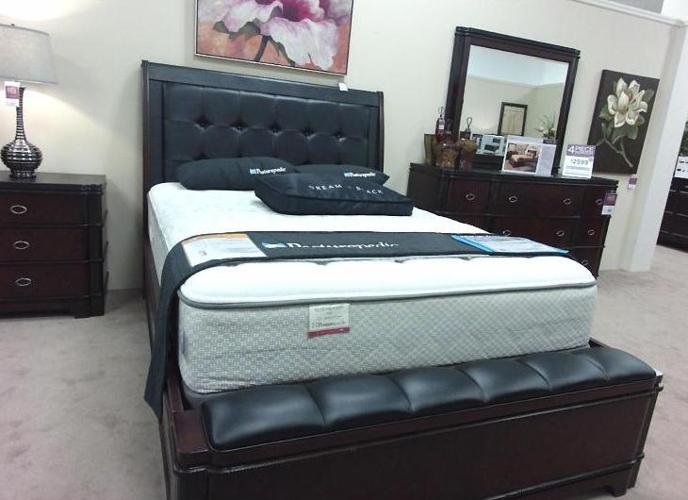}
		\includegraphics[width=\textwidth]
		{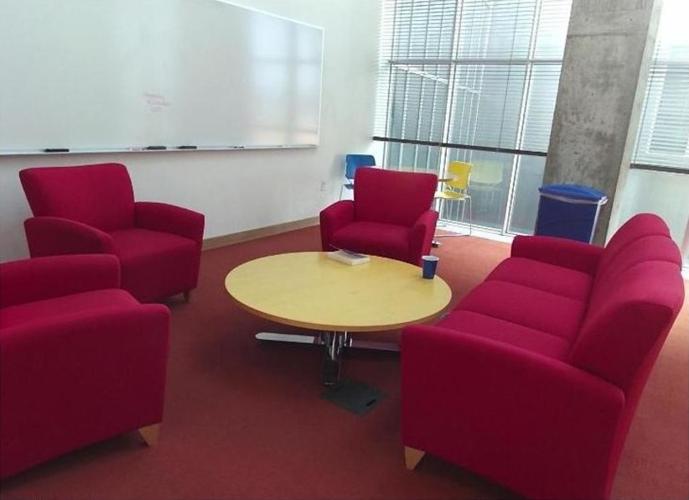}
		\includegraphics[width=\textwidth]
		{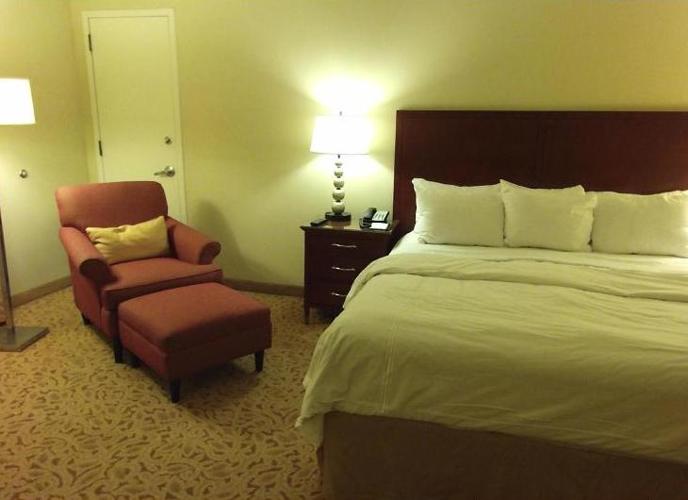}
		\caption{}
		\label{comp:c2}
	\end{subfigure}
	\begin{subfigure}[t]{0.155\textwidth}
		\includegraphics[width=\textwidth]
		{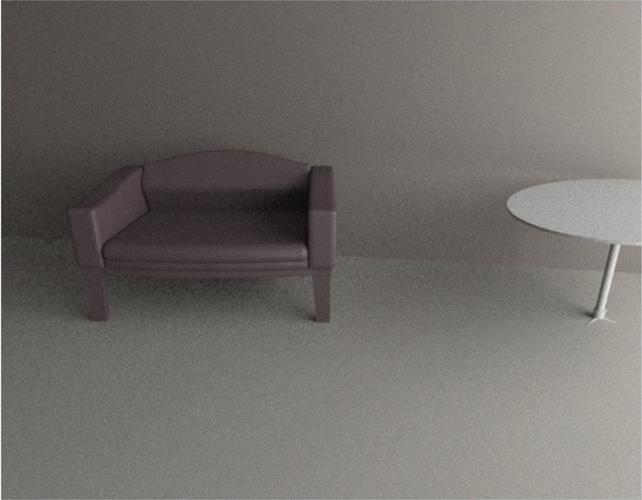}
		\includegraphics[width=\textwidth]  
		{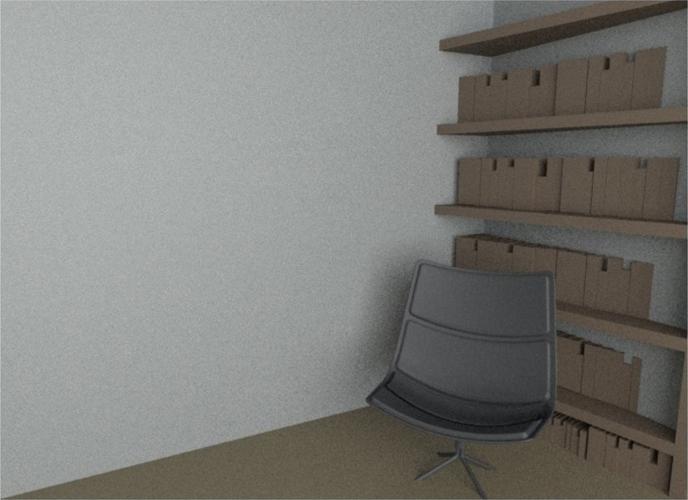}
		\includegraphics[width=\textwidth]
		{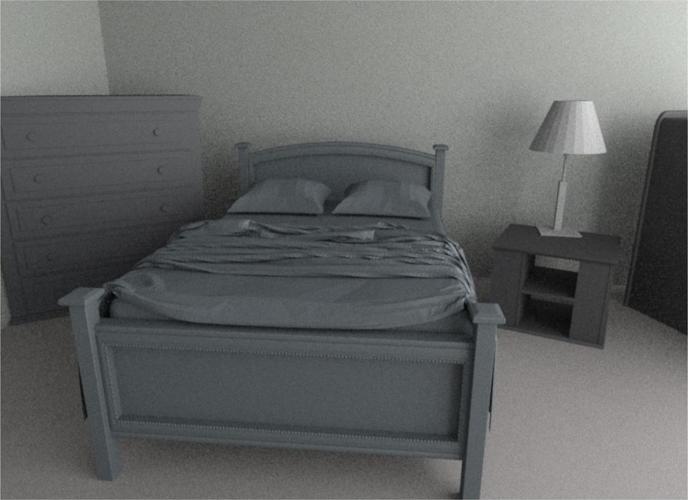}
		\includegraphics[width=\textwidth]
		{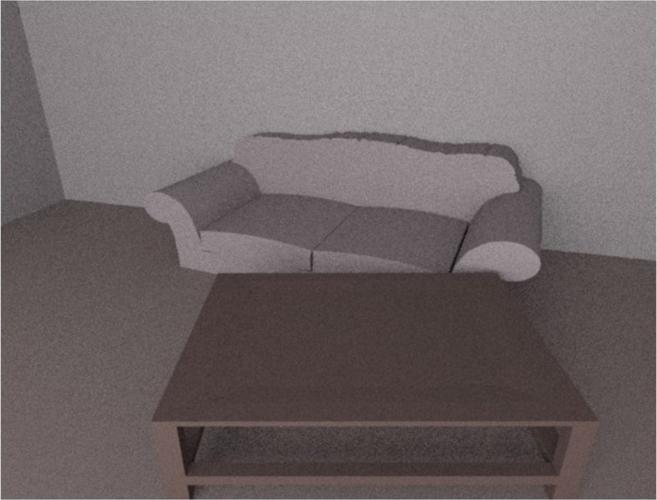}
		\includegraphics[width=\textwidth]
		{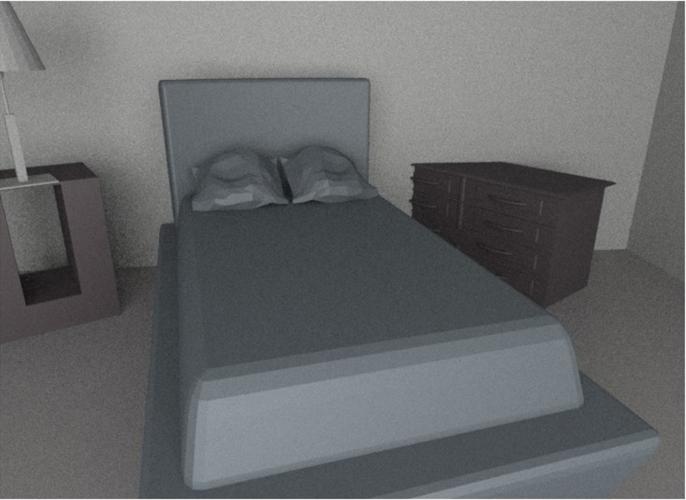}
		\includegraphics[width=\textwidth]
		{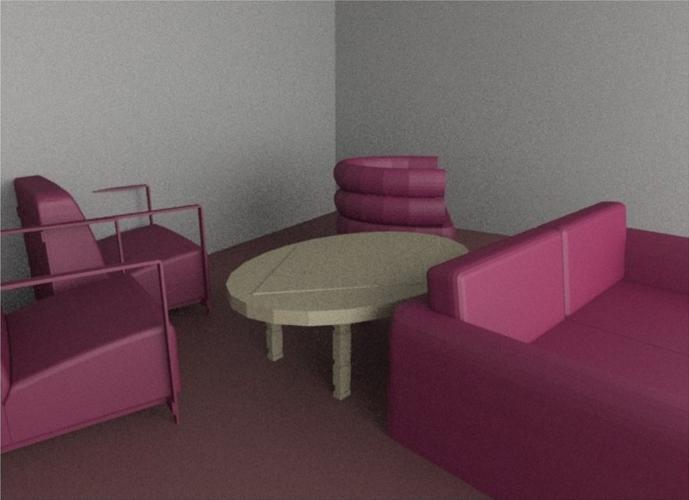}
		\includegraphics[width=\textwidth]
		{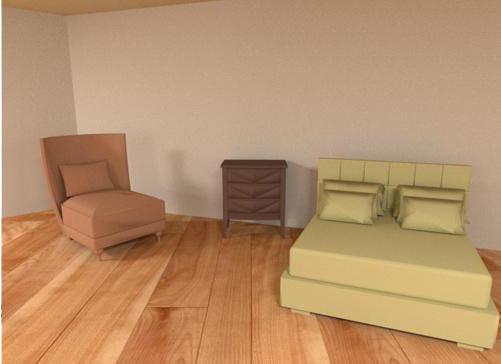}
		\caption{}
	\end{subfigure}
	\begin{subfigure}[t]{0.155\textwidth}
		\includegraphics[width=\textwidth]
		{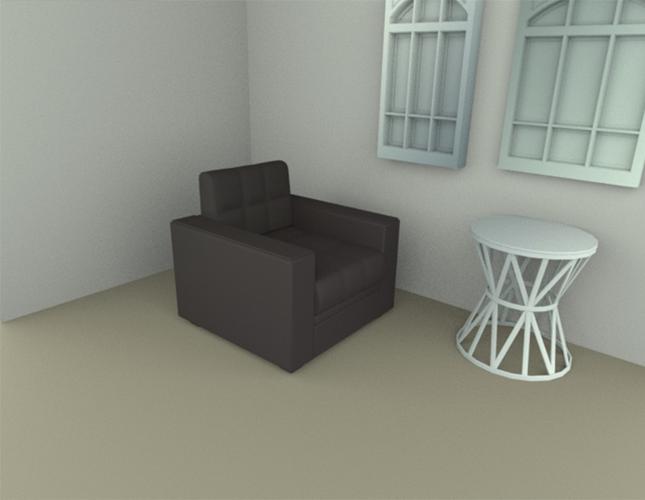}
		\includegraphics[width=\textwidth]  
		{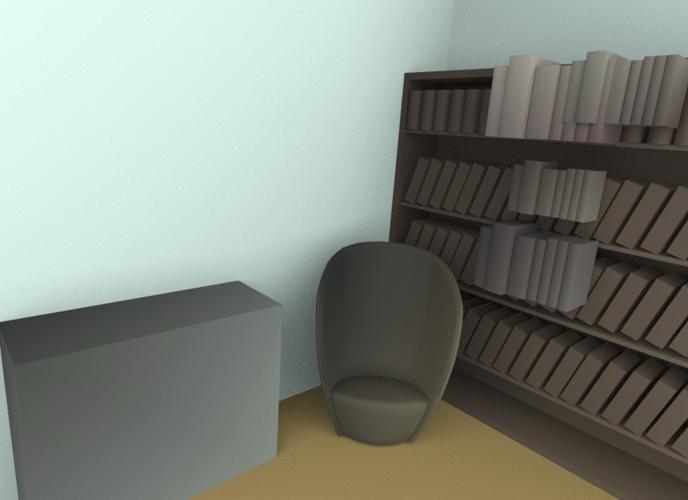}
		\includegraphics[width=\textwidth]
		{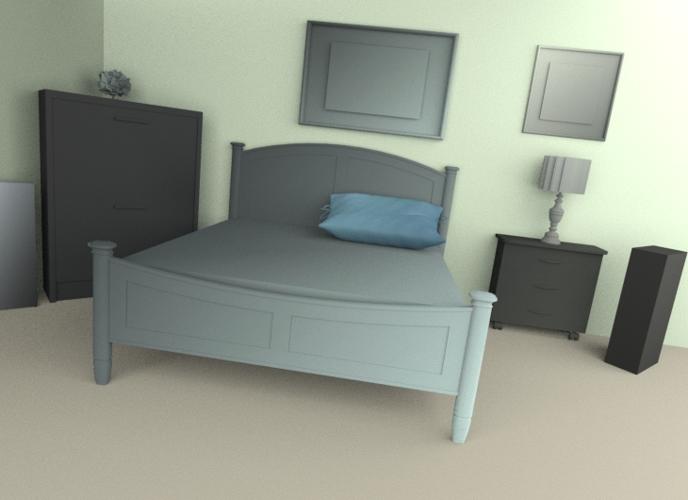}
		\includegraphics[width=\textwidth]
		{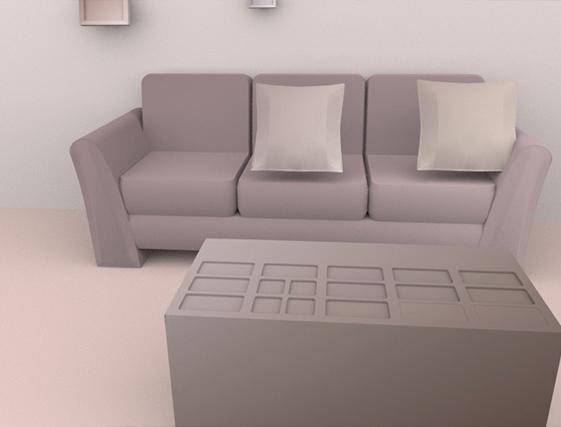}
		\includegraphics[width=\textwidth]
		{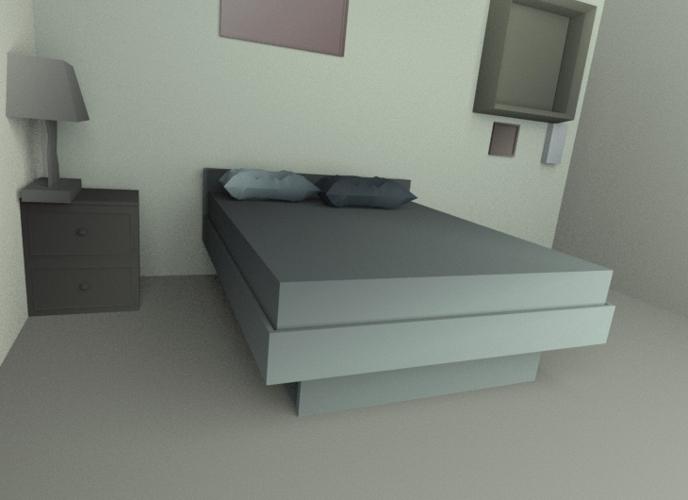}
		\includegraphics[width=\textwidth]
		{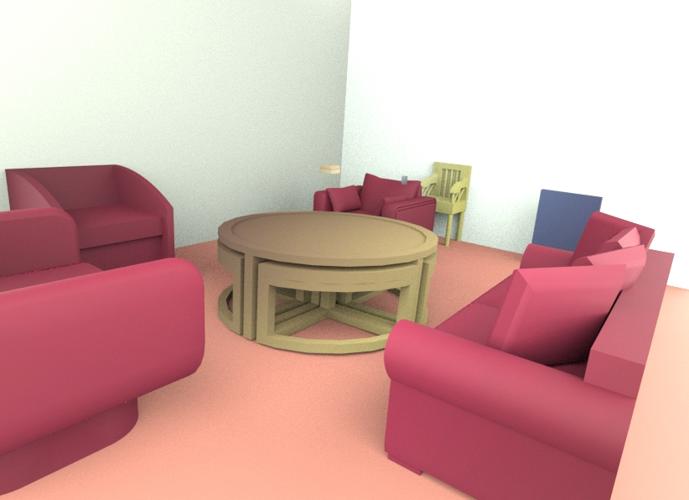}
		\includegraphics[width=\textwidth]
		{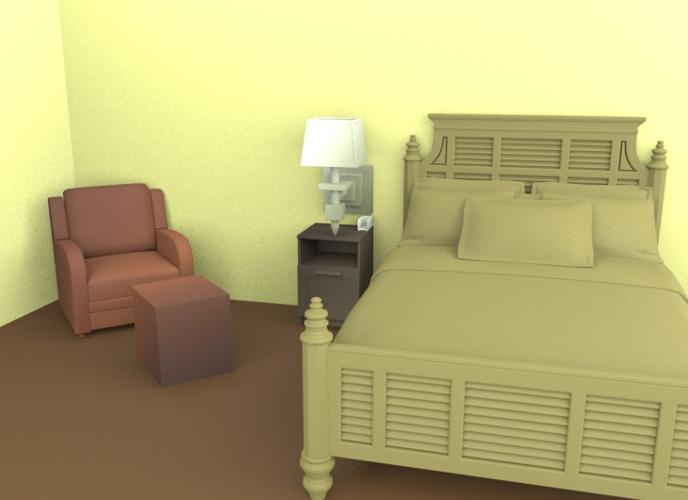}
		\caption{}
	\end{subfigure}
	\caption{Comparison with other methods. (a) and (d): The input images. (b) and (e): Reconstructed scenes from other works. The last row is provided by \cite{izadinia2017im2cad}, and the remaining results are from \cite{huang2018holistic}. (c) and (f): Our results. All the input images are from the SUN RGB-D dataset.}
	\label{fig:comparison}
\end{figure}

We compare our outputs with the state-of-the-art works from \cite{izadinia2017im2cad,huang2018holistic} (see Figure~\ref{fig:comparison}). For indoor cases with few objects and occlusions (see Figure~\ref{comp:c2}, row (1), (2), (4) and (6)), our method extracts more small-size objects (like windows, books, pictures, pillows and lamps) in addition to the main furniture than both methods. This works well with the increasing of scene complexity. Objects that are of low-resolution, hidden or partly out of view can also be captured (see Figure~\ref{comp:c1}, row (1), (3), (6) and (7)). Both of the two works \cite{izadinia2017im2cad,huang2018holistic} adopted detection-based methods to locate bounding boxes of objects in a 2D image, which would lose geometric details. Our `instance segmentation + relational reasoning' approach not only provides more object shape details, but also preserves the relative size between objects. Our context refinement also aligns the recognized models in a meaningful layout driven by the support-guided modeling.

\subsection{Quantitative Evaluation}
We here quantitatively evaluate the 3D room layout prediction, support inference and 3D object placement. 
Dense modeling of indoor scenes requires the input image to be fully segmented at the instance level. 
Therefore, we adopt the NYU v2 dataset (795 images for training and 654 images for testing) to assess the tasks of support inference, and use its manually annotated 3D scenes (a subset of the SUN RGB-D annotation dataset) to evaluate the 3D layout prediction and object placement.

\paragraph{\textbf{3D Room Layout}}
The 3D room layout presents a reference for indoor object alignment and hence influences the object placement. 
Our method is validated by measuring the average 3D IoU of room bounding boxes between the prediction and the ground-truth \cite{song2015sun}.
Table~\ref{comp:layout} illustrates the performance of our method under two configurations: 1. with camera-layout joint estimation and 2. without joint estimation (to estimate camera parameters individually from vanishing points). 
The results from this ablation experiment show that the strategy of joint estimation consistently outperforms its counterpart in all room types.
We also tested the average IoU for `living rooms' and `bedrooms' to compare with Izadinia et al. \cite{izadinia2017im2cad}. Our performance reaches 66.08\% and Izadinia et al.\cite{izadinia2017im2cad} achieves 62.6\% on a subset of SUN RGB-D dataset.

\begin{table*}[h]
	\begin{center}
		\caption{3D room layout estimation. Our method is evaluated under two configurations in different room types.}
		\label{comp:layout}
		\scalebox{1}{
			\begin{tabular}{l c c c c c c}
				\hline
				Room type  & bathroom & bedroom & classroom & computer lab & dining room &  foyer\\
				\hline
				IoU (w/o joint) & 30.71 & 39.36 & 47.60 & 20.47 & 46.28 & 54.30\\
				IoU (w/ joint) & \textbf{34.90} & \textbf{62.86} & \textbf{68.23} & \textbf{83.21} & \textbf{60.41} & \textbf{65.59}\\
				\hline
				Room type & kitchen & living room & office & playroom & study room & mean IoU\\
				\hline
				IoU (w/o joint) & 35.37 & 51.34 & 33.49 & 42.91 & 41.93 & 40.10\\
				IoU (w/ joint) & \textbf{44.01} & \textbf{67.18} & \textbf{37.55} & \textbf{55.03} & \textbf{58.22} & \textbf{57.93}\\
				\hline
		\end{tabular}}
	\end{center}
\end{table*}

\paragraph{\textbf{Support Inference}}
The testing dataset from NYU v2 contains 11,677 objects with known supporting instances and support types.
Each object is queried with four relational questions.
To make fair comparisons with existing methods, we use ground-truth segmentation to evaluate support relations (see \cite{silberman2012indoor}).
The accuracy of our method is 72.99\% at the object level,  where a prediction is marked as correct only if all the four questions are correctly answered. 
This performance reaches the same plateau as existing methods using RGB-D inputs (74.5\% by \cite{xue2015towards} and 72.6\% by \cite{silberman2012indoor}) and largely outperforms the method using RGB inputs (48.2\% by \cite{zhuo2017indoor}).
It demonstrates the feasibility of our Relation Network in parsing support relations from complicated occlusion scenarios without any depth clues.

\paragraph{\textbf{3D Object Placement}}
The accuracy of 3D object placement is tested using manually annotated 3D bounding boxes along with the evaluation benchmark provided by \cite{song2015sun}, where the mean average precision (mAP) of the 3D IoU between the predicted bounding boxes and the ground-truth is calculated. 
We align the reconstructed and ground-truth scenes to the same size by unifying the camera altitude, and compare our result with the state-of-the-art \cite{huang2018holistic}.
Different from their work, our method is designed for modeling full scenes with considering all indoor objects, while they adopted a sparsely annotated dataset SUN-RGBD for evaluation with their 30 object categories.
As the ground-truth bounding boxes of objects are not fully labeled, we remove those segmented masks that are not annotated to enable comparison under the same configuration.
Table~\ref{comp:3Dbbcmp} shows our average precision scores on the NYU-37 classes \cite{silberman2012indoor} (excluding `wall', `floor' and `ceiling'; mAP is calculated with IoU threshold at 0.15).
We obtain the mAP score at 11.49.
From Huang et al. \cite{huang2018holistic}'s work, they achieved 12.07 on 15 main furniture and 8.06 on all their 30 categories.
It shows that our approach achieves better performance in `smaller' objects, which is in line with the qualitative analysis.
The reason could be twofold: 1. a well-trained segmentation network can capture more shape details of objects (e.g. object contour) than using 2D bounding box localization; 2. most human-made objects appear with clear line segments or contours (cabinet, nightstand, dresser, etc.) which benefits our camera-layout joint estimation and model initialization.
However, for objects with a rather thin or irregular shape, or under incomplete segmentation (like chair, pillow and lamp et al.), the performance would drop by a small extent.

\begin{table*}[ht]
	\begin{center}
		\caption{3D object detection. We compare our method under three configurations: 1. without camera-layout joint estimation (w/o joint); 2. without Relation Network (w/o RN); 3. with joint estimation and Relation Network (all). The values show the average precision score on our shared object classes. The column `others' contains the remaining NYU v2 categories (mAP is averaged by 34 categories, i.e. NYU-37 classes excluding `wall', `floor', `ceiling').}
		\label{comp:3Dbbcmp}
		\scalebox{1}{
			\begin{tabular}{l c c c c c c c c c c}
				\hline
				Method & bathtub & bed & bookshelf & cabinet & chair & desk & door & dresser & fridge & lamp \\
				\hline
				Huang et al. \cite{huang2018holistic} & 2.84 & \textbf{58.29} & 7.04 & 0.48 & \textbf{13.56} & 4.79 & \textbf{1.56} & 13.71 & 15.18 & 2.41 \\
				
				Ours (w/o joint) & 30.83 & 22.62 & 5.83 & 1.82 & 1.12 & 4.31 & 0.68 & 28.53 & 25.25 & 3.12 \\
				
				Ours (w/o RN) & 40.00 & 54.21 & 6.67 & 3.59 & 2.13 & 7.61 & 0.16 & 31.74 & 45.37 & 2.78\\
				
				Ours (all) & \textbf{44.88} & 55.53 & \textbf{9.41} & \textbf{4.58} & 6.49 & \textbf{7.69} & 0.18 & \textbf{37.76} & \textbf{52.08} & \textbf{3.65} \\
				\hline
				Method & nightstand & person & pillow & shelves & sink & sofa & table & toilet & others & mAP\\
				\hline
				Huang et al. \cite{huang2018holistic} & 8.80 & 4.04 & - & - & 2.18 & 28.37 & 12.12 & 16.50 & - & -\\
				
				Ours (w/o joint) & 8.35 & 5.00 & 0.58 & 1.02 & 0.00 & 24.26 & 7.65 & 13.13 & 0.00 & 5.41\\
				
				Ours (w/o RN) & 32.51 & 8.08 & 0.20 & 3.57 & 0.25 & 31.93 & 7.56 & 10.74 & 0.00 & 8.53\\
				
				Ours (all) & \textbf{32.52} & \textbf{18.52} & \textbf{1.19} & \textbf{33.31} & \textbf{3.85} & \textbf{33.49} & \textbf{13.68} & \textbf{31.77} & 0.00 & \textbf{11.49} \\
				
				\hline 
		\end{tabular}}
	\end{center}
\end{table*}

\paragraph{\textbf{Ablation analysis}}
We implement the ablation analysis to discuss which module in our pipeline contributes most to the final 3D object placement. 
Two ablated configurations are considered (see Table~\ref{comp:3Dbbcmp}): 1. without camera-layout joint estimation \cite{izadinia2017im2cad}, 2. without Relation Network (replaced with prior-based support inference \cite{nie2018semantic}).  
The mAP scores of the first and second configurations are 5.41 and 8.53 respectively.
Our final score is 11.49. 
It implies that both the camera-layout joint estimation and relational reasoning contribute to the final performance, and room layout has a higher impact to the object placement in single-view modeling.
It is expected that, the orientation and placement of the room layout largely influence the object placement. We also observe that prior-based support inference is more sensitive to occlusions and segmentation quality \cite{nie2018semantic,xue2015towards}. When indoor scenes are cluttered, occlusions generally make the supporting surfaces invisible and the segmentation under quality. Unlike the Relation Network, the prior-based method does not take spatial relationship into account and chooses a supporting instance only considering the prior probability, making it more error-prone to complicated scenes.

\subsection{Discussions}
\label{sec:discussion}
\begin{figure}
	\centering
	\begin{subfigure}[t]{0.15\textwidth}
		\includegraphics[width=\textwidth]  
		{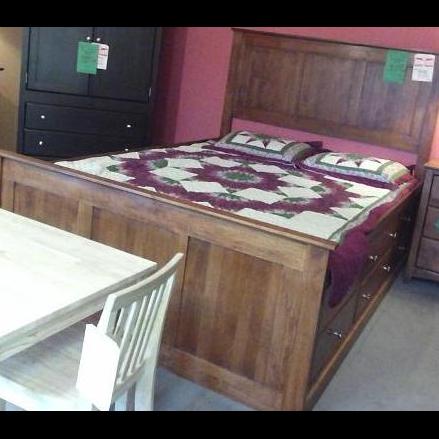}
		\includegraphics[width=\textwidth]
		{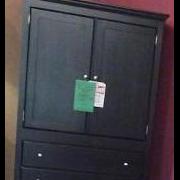}
		\caption{}
	\end{subfigure}
	\begin{subfigure}[t]{0.15\textwidth}
		\includegraphics[width=\textwidth]  
		{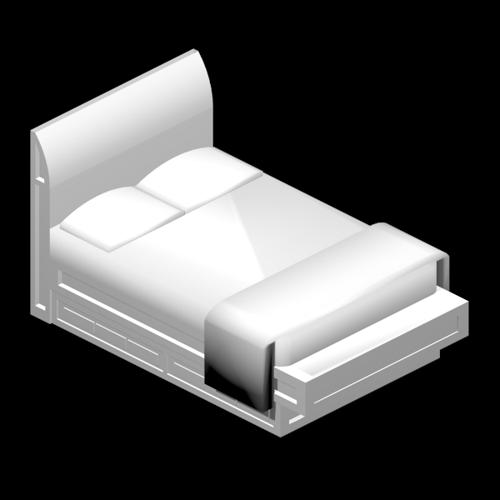}
		\includegraphics[width=\textwidth]
		{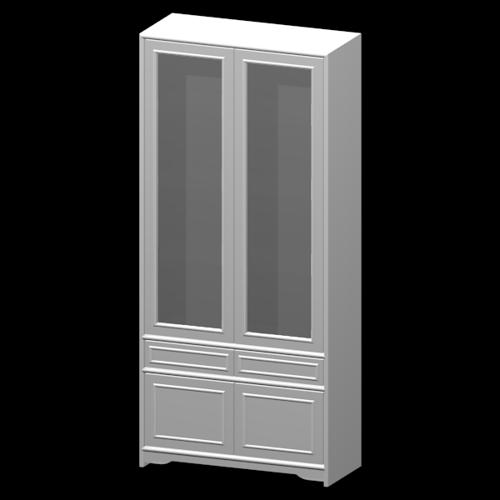}
		\caption{}
	\end{subfigure}
	\begin{subfigure}[t]{0.15\textwidth}
		\includegraphics[width=\textwidth]  
		{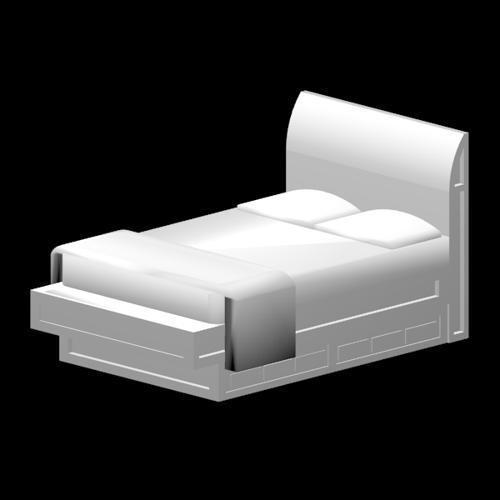}
		\includegraphics[width=\textwidth]
		{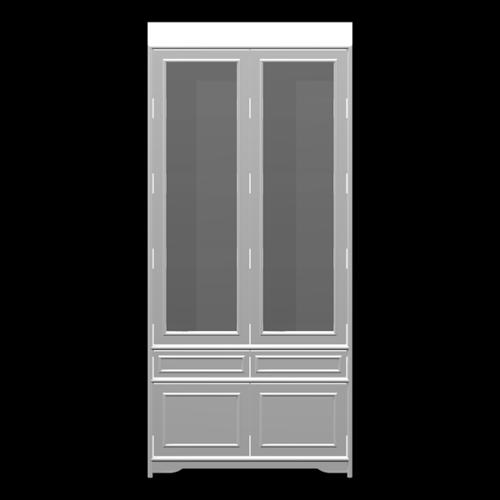}
		\caption{}
	\end{subfigure}\rulesep
	\begin{subfigure}[t]{0.15\textwidth}
		\includegraphics[width=\textwidth]
		{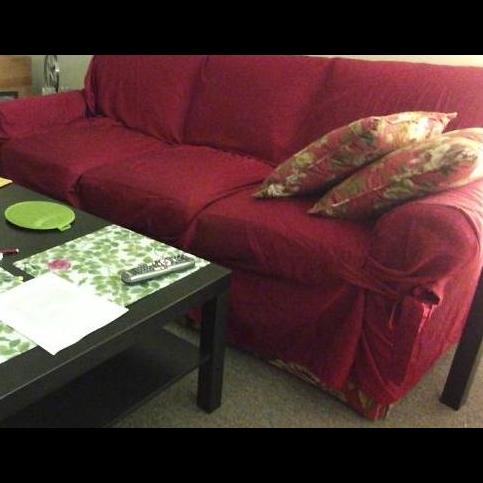}
		\includegraphics[width=\textwidth]  
		{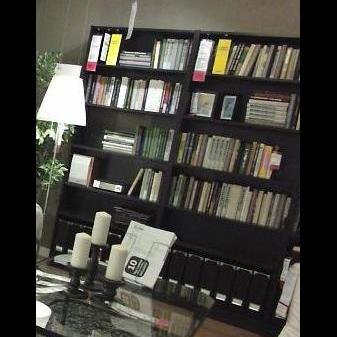}
		\caption{}
	\end{subfigure}
	\begin{subfigure}[t]{0.15\textwidth}
		\includegraphics[width=\textwidth]
		{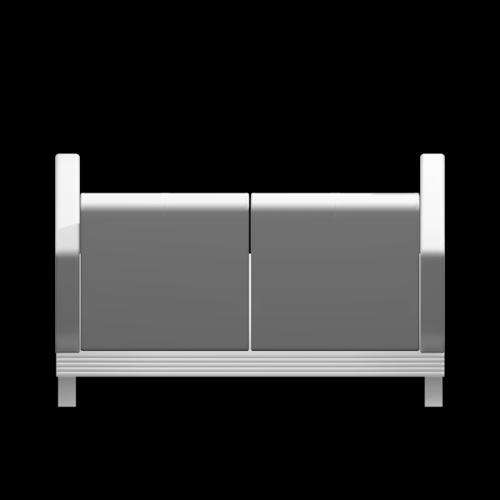}
		\includegraphics[width=\textwidth]  
		{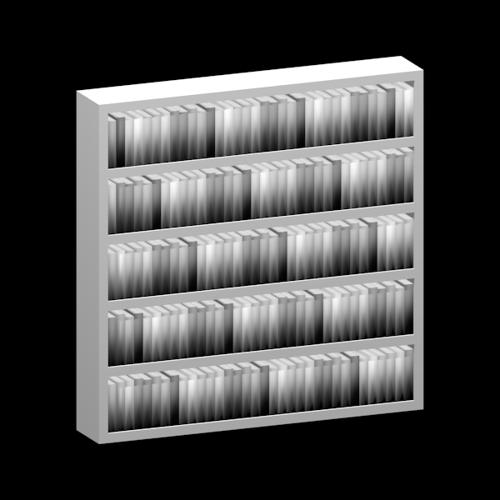}
		\caption{}
	\end{subfigure}
	\begin{subfigure}[t]{0.15\textwidth}
		\includegraphics[width=\textwidth]
		{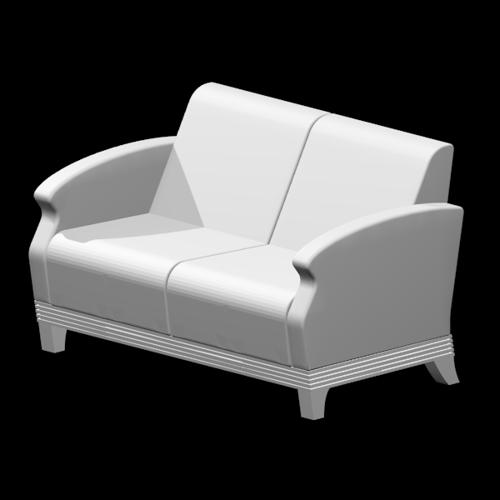}
		\includegraphics[width=\textwidth]  
		{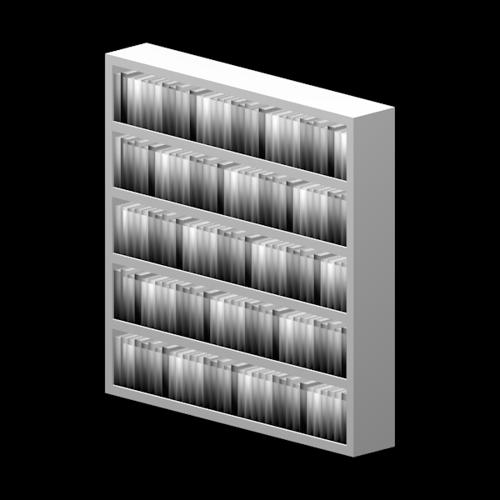}
		\caption{}
	\end{subfigure}
	\caption{Orientation correction. (a) and (d): The object images. (b) and (e): Matched models from MVRN. (c) and (f): Corrected orientations.}
	\label{fig:orient_test}
\end{figure}

\paragraph{\textbf{Improving the Estimation of Object Orientation}}
Although the view-based model matching provides an initial guess of object orientation (see Section~\ref{sec:model_ret}), those deep features are in some cases too abstract to decide sufficiently accurate orientation for trustworthy model initialization. For each object mask, we specifically append a ResNet-34 to predict the orientation angle relative to the camera. It is trained on our dataset considering eight uniformly sampled orientations (i.e. $\pi/4, \pi/2, ..., 2\pi$). However, there is a gap between the renderings (which we used for training) and the real-world images. Rather than conducting full-layer training, we fix the shallowest three layers with the weights pretrained on ImageNet to make our network sensitive to real images. 
The training data is augmented with coarse drop-out to mimic occlusion effects, and random perspective \& affine transformations to mimic different camera poses.
The top-1 precision on our testing dataset reaches 91.81\% (22342 models for training, 2482 models for testing). 
Figure~\ref{fig:orient_test} illustrates samples from the testing dataset and their predicted orientations. 
In practice, orientation of some specific models is ambiguous (e.g. symmetric shapes). Top-3 orientation candidates are selected and transformed into the room coordinate system for global scene modeling. 

\begin{figure*}
	\centering
	\includegraphics[width=\textwidth]  
	{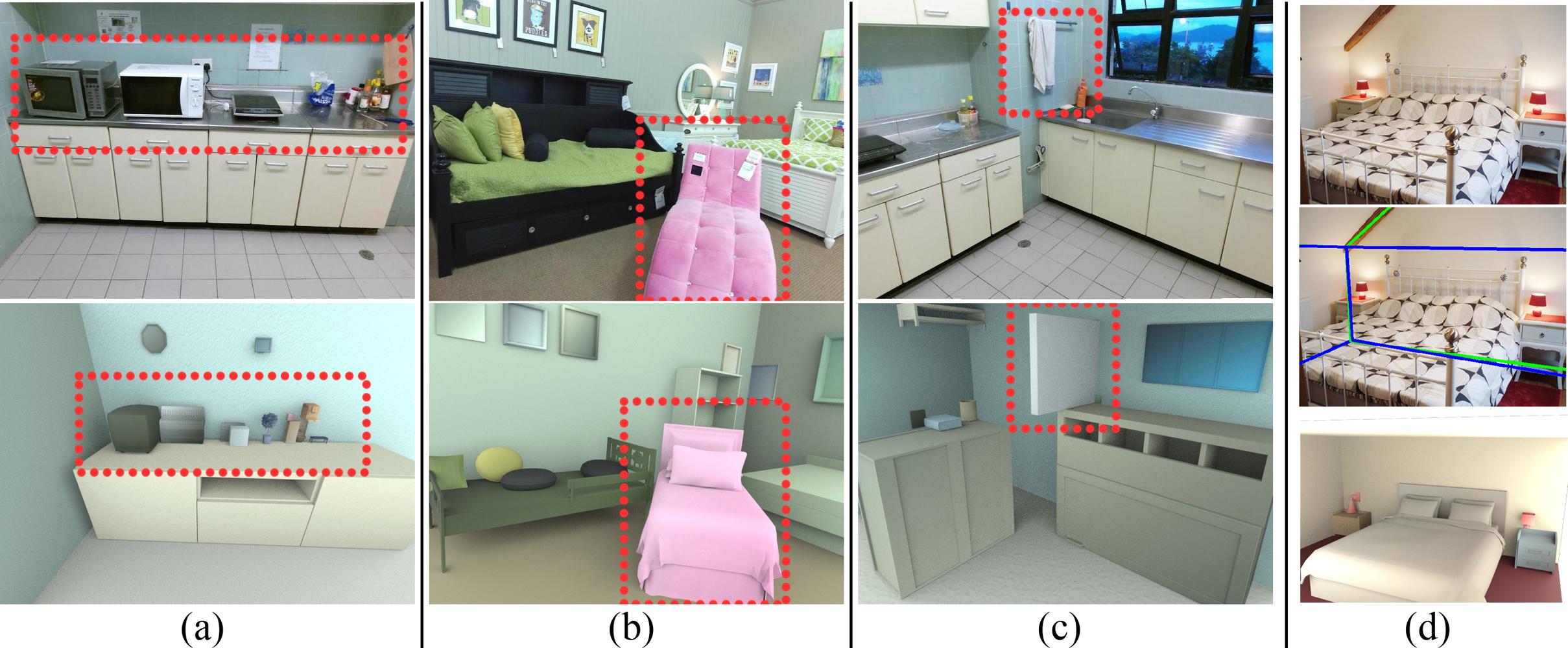}
	\caption{Limitation cases. Objects that are segmented with rather few pixels (a), out of our model repository (b) or from `other category' (right) may not get a proper geometry estimate. For `non-Manhattan' room layout (d), we fit it with a cuboid. The green and blue lines in (d) respectively represent the 2D room layout and the projection of the 3D layout.}
	\label{fig:failurecases}
\end{figure*}

\paragraph{\textbf{Limitations}}
Our method faces challenges when objects are segmented out with very few pixels (at the minimum of 24x21) which could be too raw for the MVRN to match their shape details.
Our CAD model dataset currently contains 37 common categories of indoor objects. Its capacity is limited relative to the diversity of real-world indoor environments.
While for unknown objects (labeled as `other category'), we currently use a cuboid to approximate their shape.
Besides, our current method would fit any room layout with a box, which would fail when handling extremely irregular room shapes. Therefore, those reasons above would undermine the IoU accuracy in our contextual refinement, and we illustrate those cases in Figure~\ref{fig:failurecases}.

\section{Conclusions and Future Work}
We develop a unified scene modeling approach by fully leveraging convolutional features to reconstruct semantic-enriched indoor scenes from a single RGB image. 
A shallow-to-deep process parses relational and non-relational context into structured knowledge to guide the scene modeling. 
The experiments demonstrate the capability of our approach in (1) automatically inferring the support relationship of objects, (2) dense scene modeling to recover 3D indoor geometry, with enriched semantics and trustworthy modeling results. 
Our quantitative evaluations further demonstrate the functionality and effectiveness of each substep in producing semantically-consistent 3D scenes. 

This work aims at 3D scene modeling through fully understanding scene context from images. There are high-level relational semantics among indoor objects that could be incorporated into the modeling-by-understanding approach, like other complex contact relations (e.g. a person sits on a chair and holds a mug). All these mixed semantics would help our system to better understand and represent the scene context in a meaningful way. It suggests our future work to provide an intelligent scene knowledge structure to configure and deploy them towards scene modeling.

\section*{Acknowledgment}
The research leading to these results has been partially supported by VISTA AR project (funded by the Interreg France (Channel) England, ERDF), Innovate UK Smart Grants (39012), the National Natural Science Foundation of China (61702433, 61661146002), the China Scholarship Council and Bournemouth University.

\bibliographystyle{unsrt}
\bibliography{template} 

\appendix
\section{Technical illustrations}
\label{appendix_a}
The network configurations and parameter decisions involved in our scene modeling are detailed in this part.

\subsection{Indoor scene segmentation}
\label{appendix_seg}
As Mask R-CNN \cite{he2017mask} is designed for general instance segmentation, to make it robust in learning from a small indoor dataset (795 images in our case), we augment the training data with a horizontal flip, and train the network by stages. Specifically, the whole training is divided into three phases, we firstly train the Region Proposal Network, Feature Pyramid Network and mask prediction layers with other parts frozen (60 epochs with learning rate at 1e-3), and fine-tune the ResNet by freezing the shallowest four layers (120 epochs with learning rate at 1e-3) followed by an all-layer training (160 epochs with learning rate at 1e-4). In the inference phase, the searched region proposals go through Non-Maximal Suppression to remove overlaps and keep objects with higher classification scores.

\subsection{Model Retrieval}
\label{appendix_model_ret}
To build the CAD model dataset, we collect 26,695 indoor models covering 37 categories from ShapeNet \cite{chang2015shapenet} and SUNCG \cite{song2016ssc}, along with a `cuboid' category for objects that are labeled as `other' in NYU v2.  
We align and render each model from 32 viewpoints for appearance-based matching, with two elevation angles (15 and 30 degrees) and 16 uniform azimuth angles. The
Multi-View Convolutional Network \cite{su2015multi} is customized with 32 parallel branches of ResNet-50 \cite{he2016deep} as feature extractors (with the last layer removed).
All those ResNets share the same weights. 
The deep features outputted from those branches are max-pooled and fully connected for recognition. 
The full network is pretrained on ShapeNet for shape recognition task.
In our scene modeling, the major color texture from object masks is mapped to CAD models for rendering 3D scenes.

\subsection{Relation Network}
\label{appendix_rel}
The whole architecture consists of three parts (see Figure~5): the Vision part, the Question part, and the Relation reasoning part. The Vision part is designed to encode the image and its segmentation by a set of abstract CNN features. The Question part is to rephrase each question into an encoded vector to ensure our system able to understand human language. 
The Relational reasoning part is responsible to analyze the image features and answer the corresponding questions. In the Vision part, we adopt five layers of convolutional kernels (3x3x64 for each layer with the stride and padding size at 2 and 1 respectively). Each convolution is followed by a ReLU and a Batch Normalization layer. The input end is a 300x300x4 matrix (the resized image appended with its mask), and it outputs a 10x10x64 feature map which can be seen as 10x10 of 64-dimensional feature vectors. In the Relational reasoning part, we get exhaustive pair combinations of those 10x10 feature vectors. Each pair of combination is concatenated with their 2D image coordinates correspondingly and the question vector. Thus the image features and the question vector are concatenated into 100x100 visual question vectors. All those vectors separately go through four fully-connected layers, and it generates 100x100 512-dimensional vectors. We take element-wise summation of them and output a (104 dimensional) answer vector after walking-through three fully-connected layers. All the three fully-connected layers above consist of 512 hidden neurons, and each layer is followed by a ReLU unit except the final prediction layer. The initial learning rate is at 0.001 with the wight decay rate at 0.5 in every 10 steps. 60 epoches in total are used for training.

\subsection{Global scene optimization}
\label{appendix_global_opt}
In Section~6, we set the room height at three meters, and the height of every objects are calculated relatively. To ensure that each height estimate is in a reasonable interval, we parse the ScanNet dataset \cite{dai2017scannet} to conclude a prior height distribution for each object category (see Figure~\ref{fig:height_dist1} - \ref{fig:height_dist4}). Each sample in this normal distribution represents a height ratio of a real scanned object relative to the room. A height estimate is regarded as outliers if it is outside the 3$\sigma$ interval, and should be replaced by the mean value. 

The object sizes and positions are fine-tuned with our contextual refinement. In the optimization problem (see Equation (4)), there are six continuous variables (in $\bm{S}_{i}$ and $\bm{p}_{i}$) we can control in the optimization process with BOBYAQ method. The constraints (5) and (6) have guaranteed that all objects are attached on their supporting surface. Practically, we further constrain the boundary of $\bm{S}_{i}$ to make its size only adjustable in a given interval. we use $s_{i,3}$ in $\bm{S}_{i}$ to control the aspect ratio of a CAD model, and $s_{i,1}$ and $s_{i,2}$ to decide its horizontal ratio relative to its height. For common objects (labeled as known NYU v2 categories), we opt to set $s_{i,3} \in [0.9, 1.1]$, and $s_{i,1}, s_{i,2} \in [0.8, 1.2]$. For other objects (labeled as 'other furniture' or 'other structure'), 3D cuboid is used for model retrieval. In this case, we set the boundary of the horizontal ratio more flexible as $s_{i,1}, s_{i,2} \in [0.1, 10]$.

\section{Priors for support inference and height estimation}
\label{appendix_prior_height}
We parse the ScanNet \cite{dai2017scannet} dataset to get the priors about support relationships and object heights.
It contains 1,513 real scene scans with 37,831 indoor objects, and those objects are categorized by the same label set with our experiments.
We estimate the bounding box of each object and get the height distribution as the Figure~\ref{fig:height_dist1} - \ref{fig:height_dist4} shows.
Each sample in these distributions is a ratio number of the object height to the room height.
If a height estimate is beyond $[\mu-3\sigma, \mu+3\sigma]$ ($\mu$ is the mean value and $\sigma$ is the standard deviation of the corresponding distribution), we replace the estimate with $\mu$ to initialize the object height.

Moreover, we extract the point cloud of objects to obtain support relationships within all of the scans and get one-to-one support relationship priors (with the method in \cite{wong2015smartannotator}) as the Figure~\ref{fig:support_priors} shows. Each block in the two matrices denotes the number of cases that an object (in row) is supported by another object (in column) from below (Figure~\ref{fig:support_priors}(a)) or behind (Figure~\ref{fig:support_priors}(b)). Floating objects are removed, and each object must be supported by another object. When multiple support relationships exist, only the primary one is kept (see \cite{wong2015smartannotator}).

\section{2D object segmentation comparisons with existing works}
\label{appendix_seg_comp}
2D segmentation is designed to provide the object locations in the image. 
Detection loss in 2D images directly results in their 3D counterparts missing in the final CAD scenes. 
Besides that, whether an object is segmented with a fine-grained mask would also affect the geometry estimation.
With this concern, we measure the Pixel Accuracy (PA), Mean Accuracy (MA) and Intersection over Union (IoU) \cite{garcia2017review} between the predicted and ground-truth masks to assess our performance on 40 categories in NYU v2 dataset. 
In testing, we select object masks with detection score greater than 0.5 from Mask R-CNN and layout masks from FCN to fully segment images. Table~\ref{comp:seg} illustrates the comparison with state-of-the-art methods.
The results demonstrate that we achieve higher performance in terms of PA and IoU scores. 
It is worth noting that we are mostly concerned about the IoU score which is the optimization target of our contextual refinement.

\begin{table}[ht]
	\begin{center}
		\caption{Semantic segmentation on NYU v2 (40 classes). IoU* score is the metric we are concerned in the step of contextual refinement.}
		\label{comp:seg}
		\begin{tabular}{l c c c c}
			\hline
			Method      & Data type & PA & MA & IoU* \\
			\hline
			Gupta et al.\cite{gupta2014learning}  & RGB-D & 60.3 & -    & 28.6\\
			FCN-32s \cite{long2015fully}          & RGB   & 60.0 & 42.2 & 29.2\\
			FCN-HHA \cite{long2015fully}          & RGB-D & 65.4 & 46.1 & 34.0\\
			Lin et al.\cite{lin2018exploring}    & RGB   & 70.0 & \textbf{53.6} & 40.6\\
			\hline
			Our work        & RGB & \textbf{70.3} & 49.0 & \textbf{41.6}\\
			\hline
		\end{tabular}
	\end{center}
\end{table}
The 2D IoU from Mask R-CNN \cite{he2017mask} only reaches 41.6\% though it have reached the state-of-the-art. The accuracy of 3D object placement (i.e. 3D IoU) generally should be much lower for the depth ambiguity. Its indeed a bottleneck for all kinds of single image based scene reconstruction methods \cite{huang2018holistic,izadinia2017im2cad}. However, different from 2D segmentation, the physical plausibility in 3D scene modeling (i.e. relative orientations, sizes, and support relations between objects) could affect the visual performance greater, comparing with the impact from object placement accuracy (i.e. 3D IoU).

\begin{figure}[!h]
	\centering
	\begin{subfigure}[t]{0.24\textwidth}
		\includegraphics[width=\textwidth]  
		{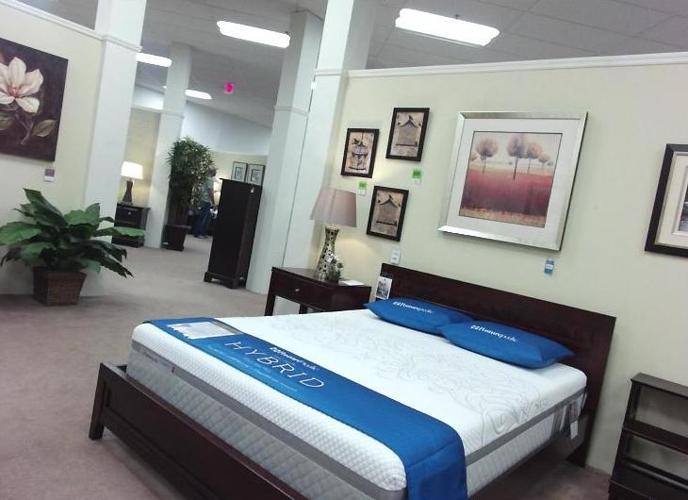}
	\end{subfigure}
	\begin{subfigure}[t]{0.24\textwidth}
		\includegraphics[width=\textwidth]  
		{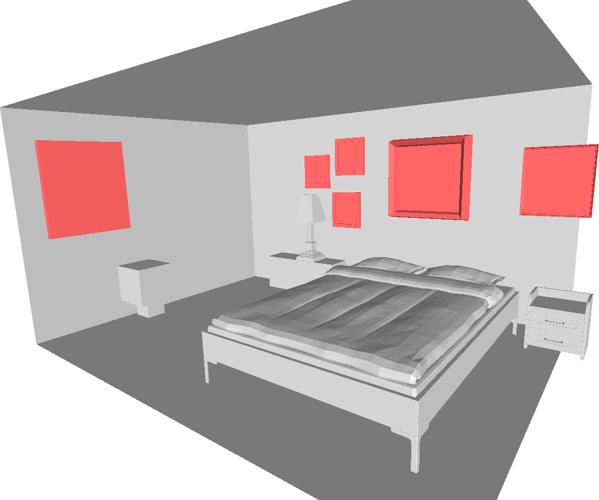}
	\end{subfigure}
	\begin{subfigure}[t]{0.24\textwidth}
		\includegraphics[width=\textwidth]  
		{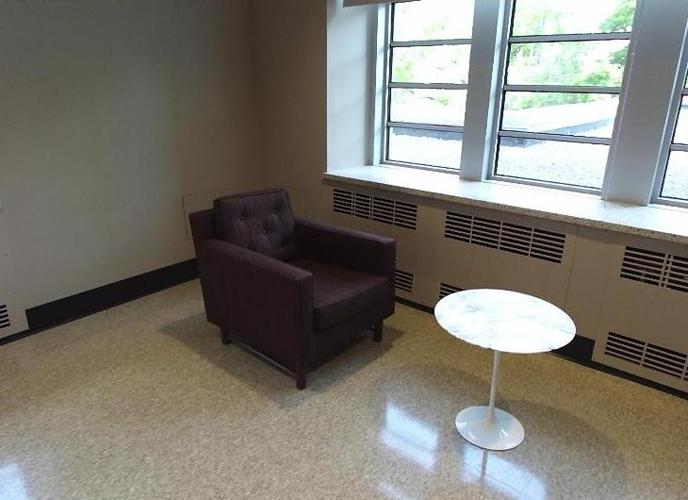}
	\end{subfigure}
	\begin{subfigure}[t]{0.24\textwidth}
		\includegraphics[width=\textwidth]  
		{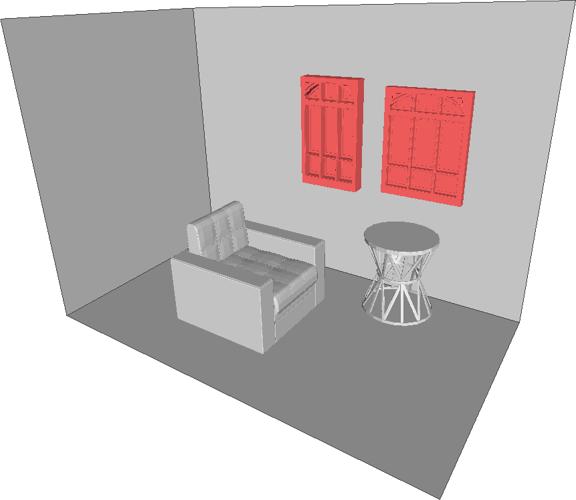}
	\end{subfigure}
	\caption{Reconstruction of `thin' structures.}
	\label{fig:thin_cases}
\end{figure}

\begin{figure}
	\centering
	\includegraphics[width=0.8\textwidth]{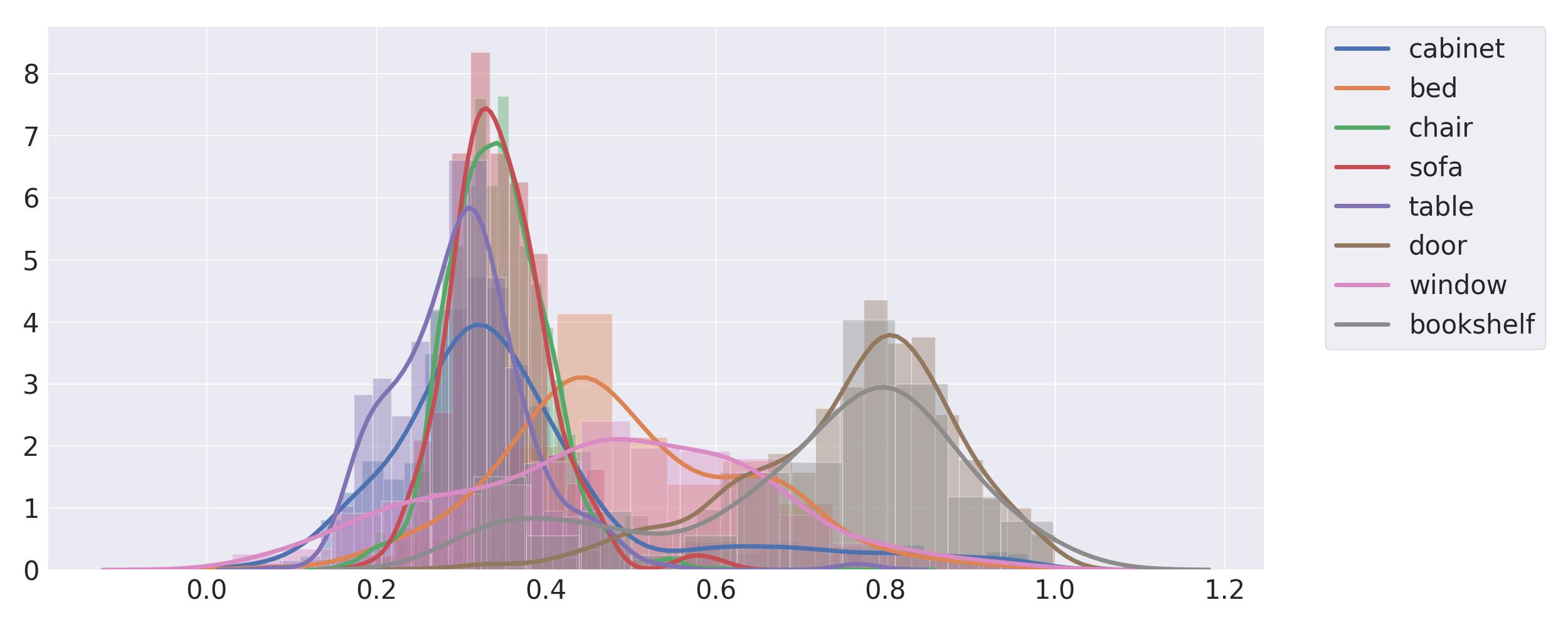}
	\caption{Height distribution for each object category. (1-8 categories)}
	\label{fig:height_dist1}
\end{figure}

\begin{figure}
	\centering
	\includegraphics[width=0.8\textwidth]{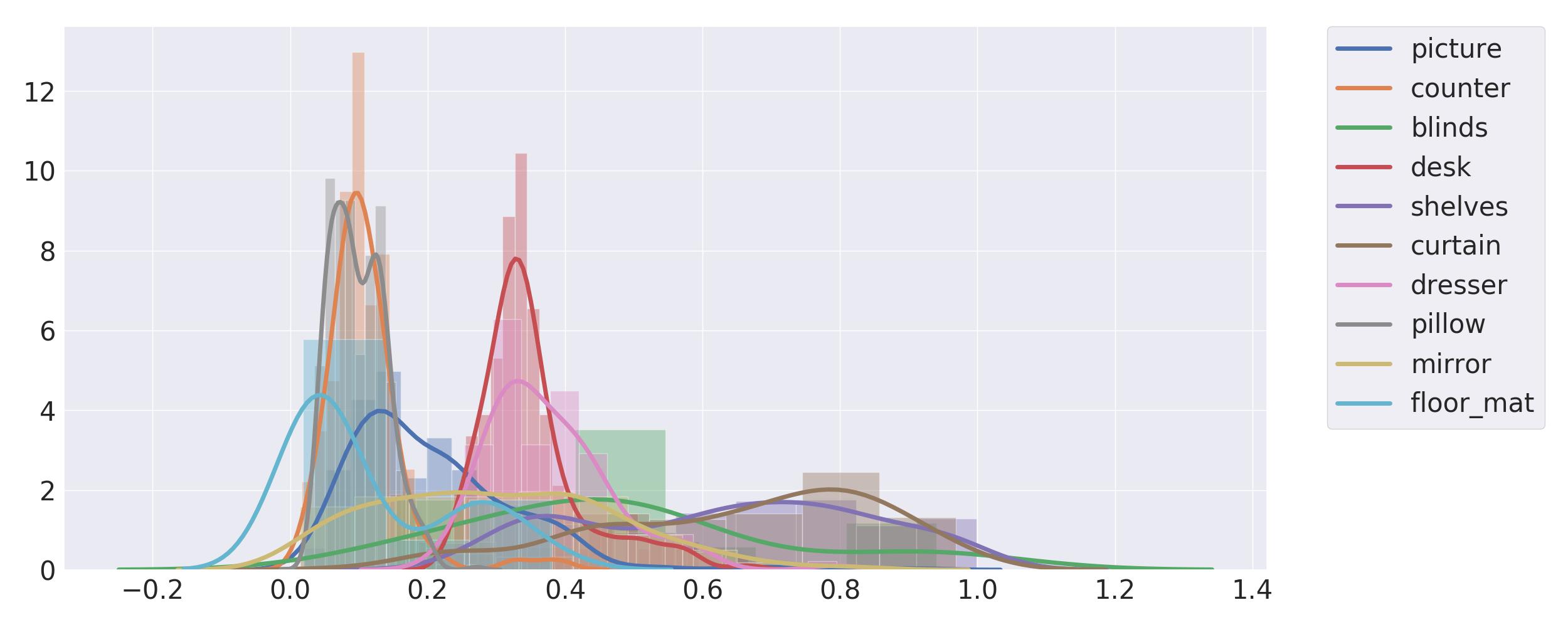}
	\caption{Height distribution for each object category. (9-18 categories)}
	\label{fig:height_dist2}
\end{figure}

\begin{figure}
	\centering
	\includegraphics[width=0.8\textwidth]{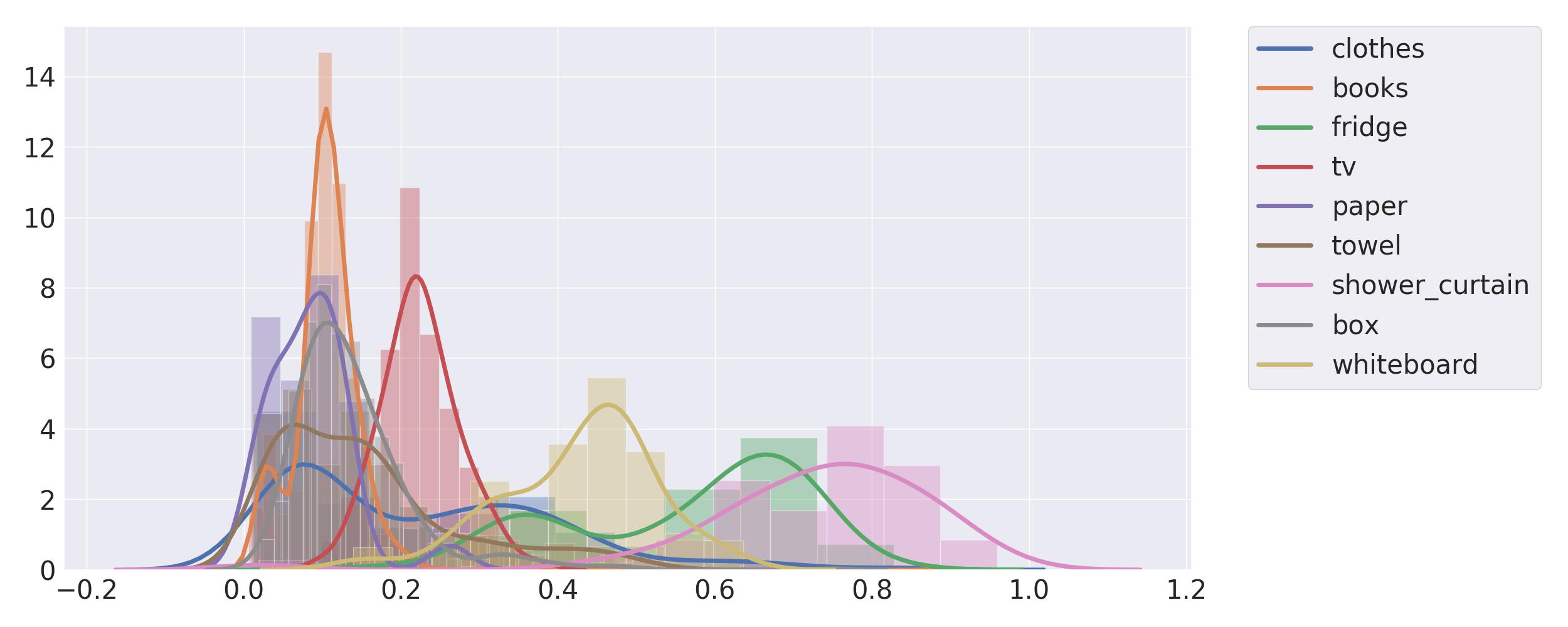}
	\caption{Height distribution for each object category. (19-27 categories)}
	\label{fig:height_dist3}
\end{figure}

\begin{figure}
	\centering
	\includegraphics[width=0.8\textwidth]{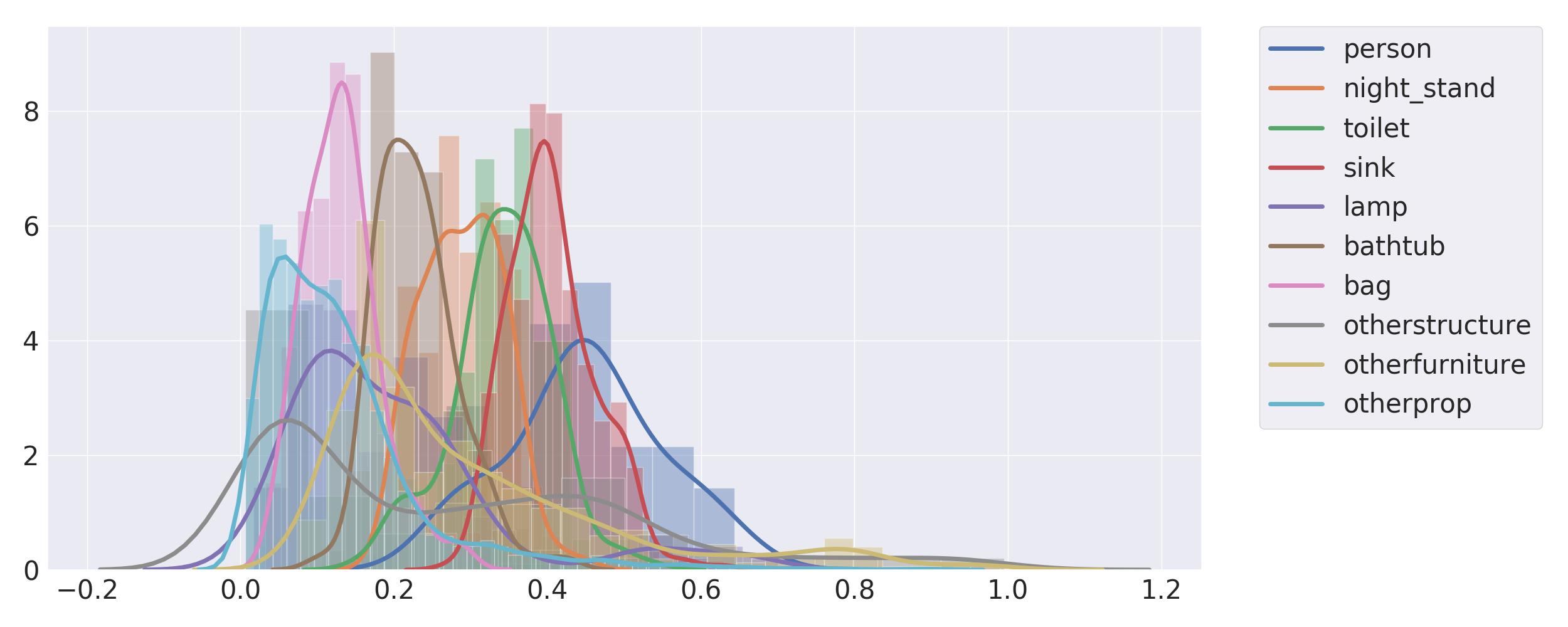}
	\caption{Height distribution for each object category. (28-37 categories)}
	\label{fig:height_dist4}
\end{figure}

In our work, there basically are two factors we most concern: plausibility and placement accuracy. On this basis, we found that obtaining trustworthy physical constraints shows better plausibility and takes more semantic meanings (e.g. support relations) than only chasing placement accuracy. We present an example in Figure~\ref{fig:thin_cases}. In indoor scenes, there are 40 object categories (NYU-40 \cite{silberman2012indoor}). Except big-size categories like beds, sofas, tables, etc., most objects are thin or small and occupy little spatial volume (see the pictures and windows in Figure~\ref{fig:thin_cases}). In our experiment, we observed that the 3D IoUs between them and the ground-truth are close to zero, because of their `skinny' size making the IoU metric vulnerable to placement disparities. However, they are still reconstructed with plausible visual performance because their orientations, sizes and support relations are reasonable. That means, a small 3D offset from the ground-truth will largely lower the accuracy of 3D IoU, but would not affect the visual plausibility given reasonable physical constraints (support relations, orientations and object sizes).

\section{Intermediate results in scene modeling}
\label{appendix_results}
We randomly pick around 50 indoor images with different complexity from SUN-RGBD dataset \cite{song2015sun}. The modeling results with intermediate outputs are illustrated in Figure~\ref{set5}. The first column shows the input image. The layout edge map and label map are placed in the second and the third column respectively. The fourth column presents the jointly estimated room layout. We illustrate the scene segmentation and the support inference results in the fifth column. Note that the support relationship is represented with an arrow. For example, the red arrow from A to B denotes A supports B from below, and the blue arrow denotes A supports B from behind. We put the modeled scenes in the sixth column (raw scene meshes without texture-mapping and rendering)

\begin{figure}[!ht]
	\centering
	\begin{subfigure}[t]{\textwidth}
		\includegraphics[width=\textwidth]  
		{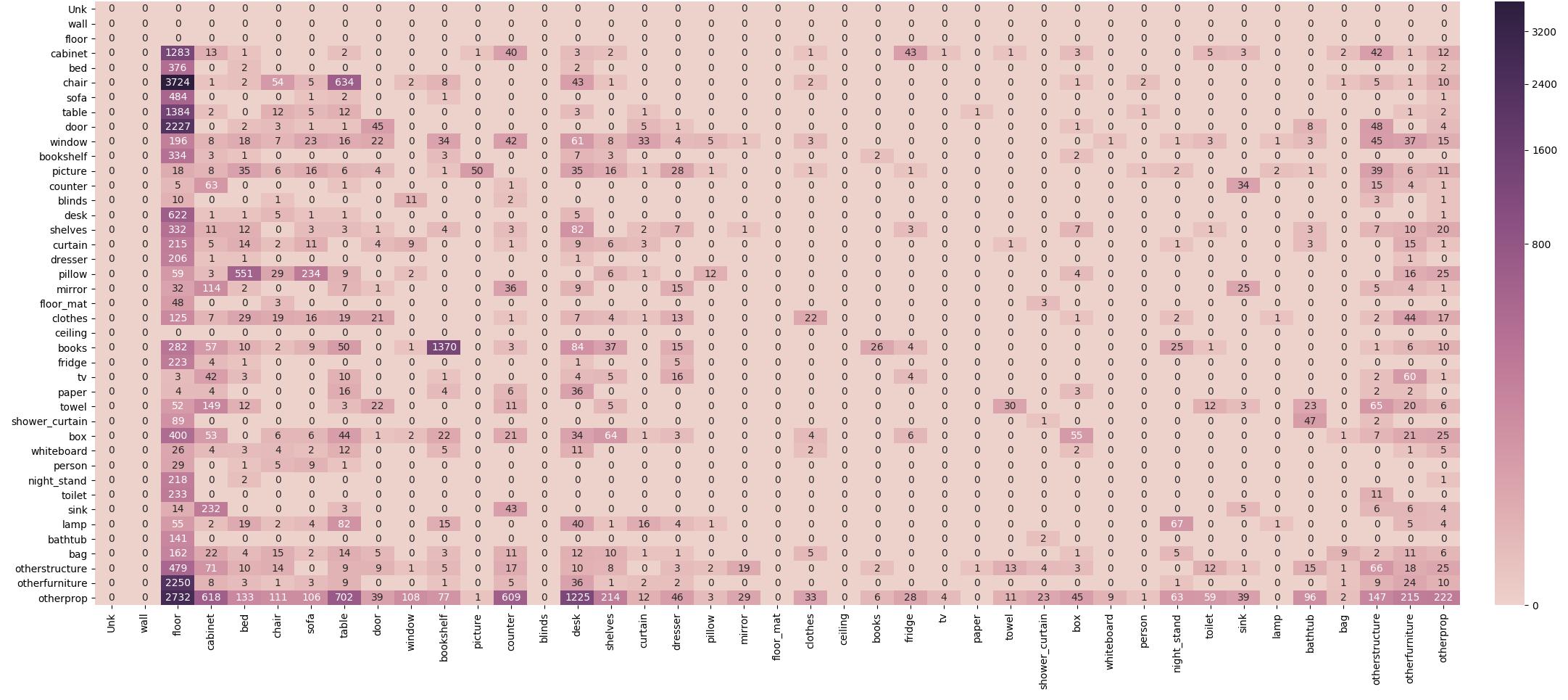}
		\caption{Support from below}
	\end{subfigure}\hfill
	\begin{subfigure}[t]{\textwidth}
		\includegraphics[width=\textwidth]
		{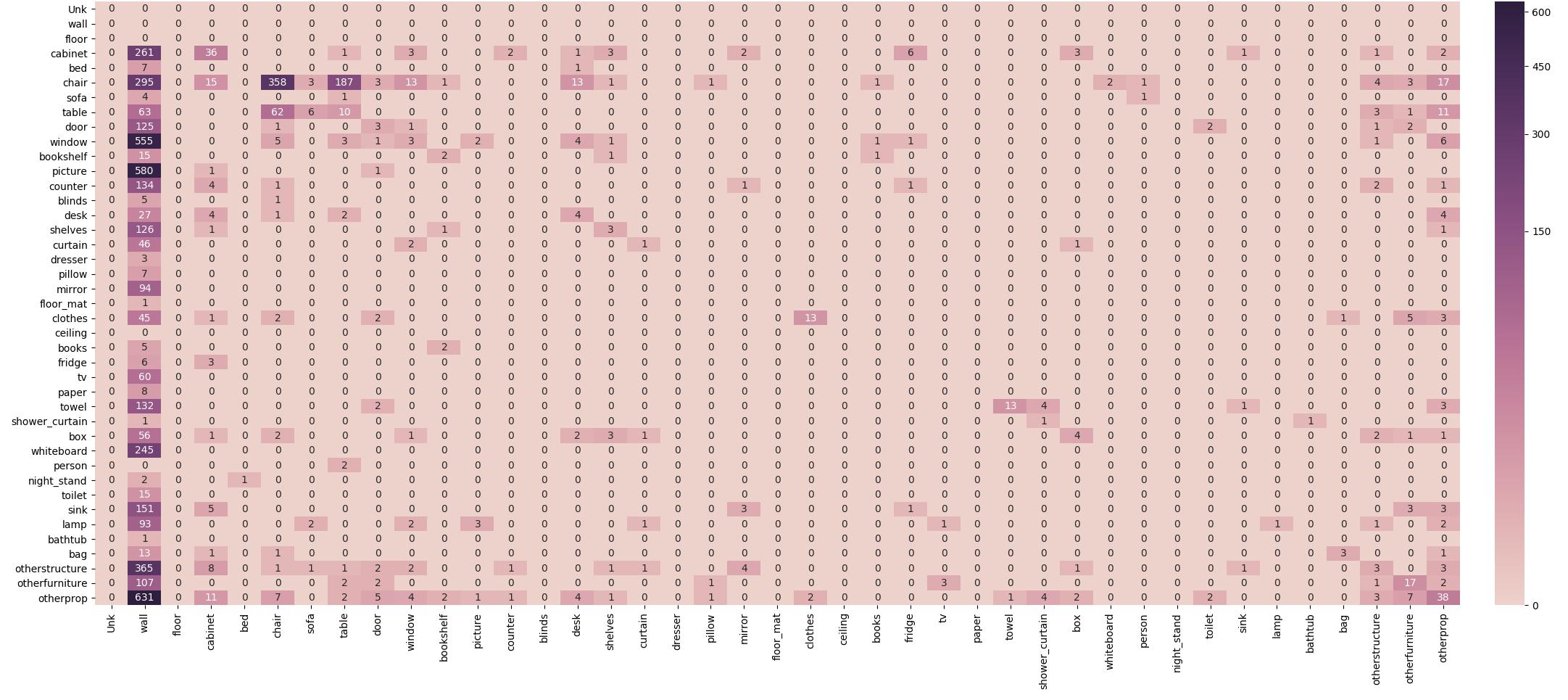}
		\caption{Support from behind}
	\end{subfigure}\hfill
	\caption{Support relationship priors}
	\label{fig:support_priors}
\end{figure}

\begin{figure}
	\centering
	\begin{subfigure}[t]{0.161\textwidth}
		\includegraphics[width=\textwidth]  
		{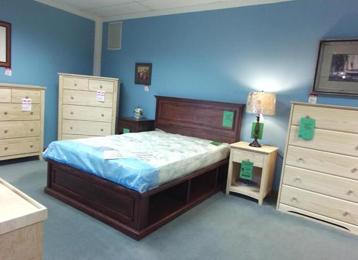}
		\includegraphics[width=\textwidth]
		{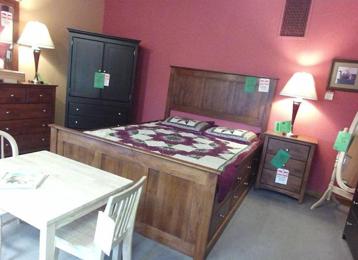}
		\includegraphics[width=\textwidth]
		{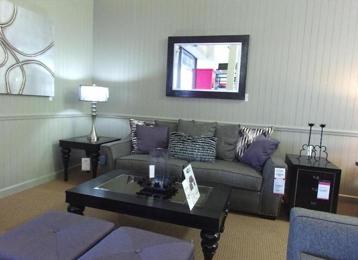}
		\includegraphics[width=\textwidth]
		{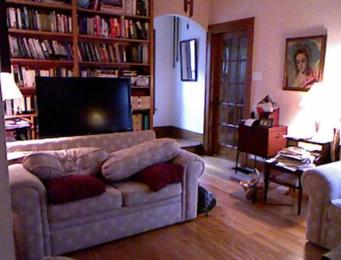}
		\includegraphics[width=\textwidth]  
		{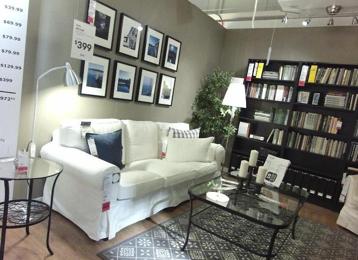}
		\includegraphics[width=\textwidth]
		{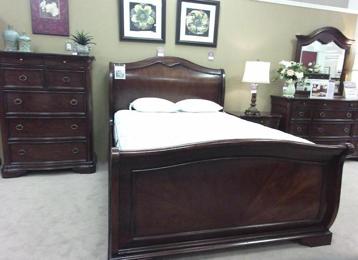}
		\includegraphics[width=\textwidth]
		{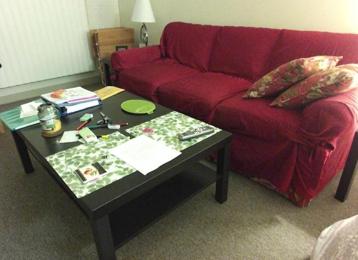}
		\includegraphics[width=\textwidth]
		{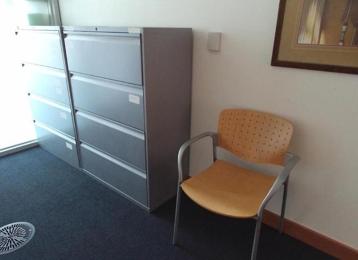}
		\includegraphics[width=\textwidth]
		{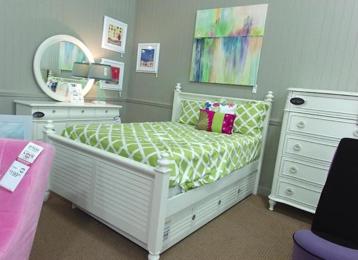}
		\includegraphics[width=\textwidth]
		{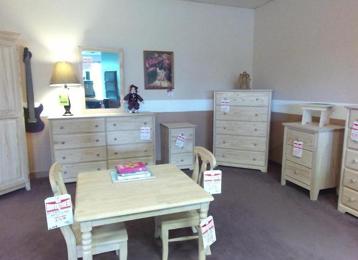}
		\includegraphics[width=\textwidth]
		{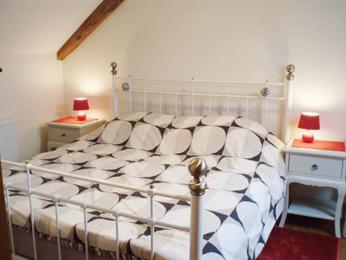}
		\caption{}
	\end{subfigure}
	\begin{subfigure}[t]{0.161\textwidth}
		\includegraphics[width=\textwidth]  
		{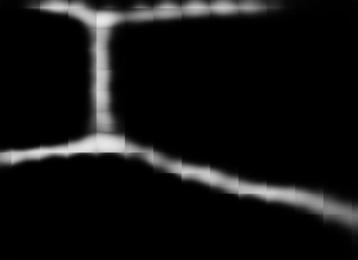}
		\includegraphics[width=\textwidth]
		{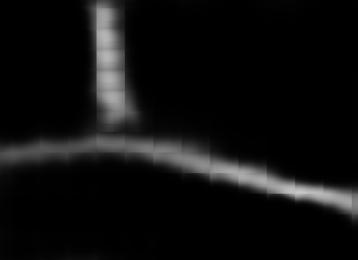}
		\includegraphics[width=\textwidth]
		{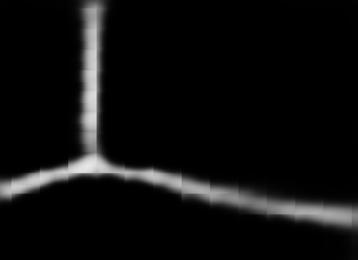}
		\includegraphics[width=\textwidth]
		{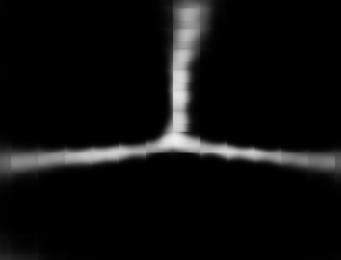}
		\includegraphics[width=\textwidth]  
		{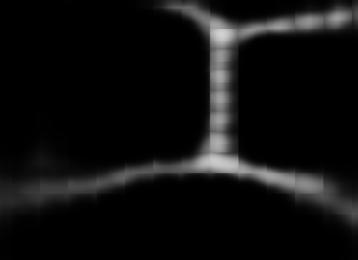}
		\includegraphics[width=\textwidth]
		{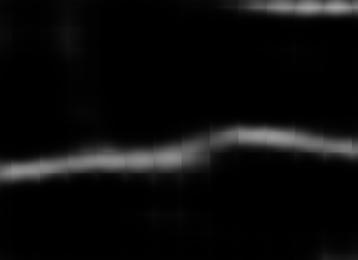}
		\includegraphics[width=\textwidth]
		{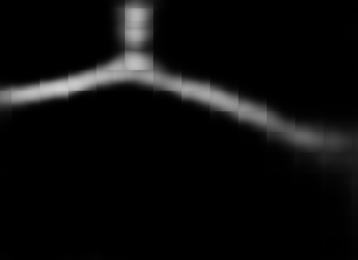}
		\includegraphics[width=\textwidth]
		{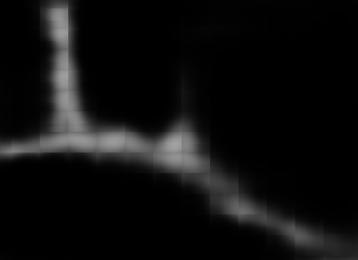}
		\includegraphics[width=\textwidth]
		{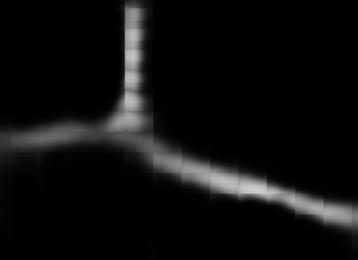}
		\includegraphics[width=\textwidth]
		{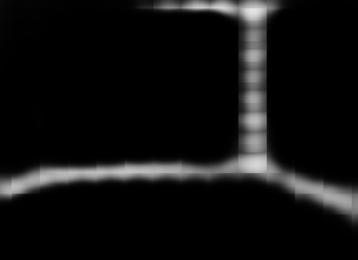}
		\includegraphics[width=\textwidth]
		{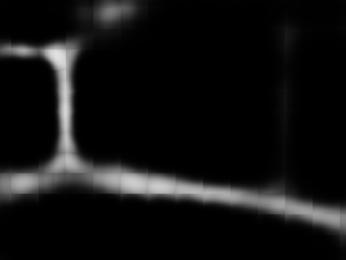}
		\caption{}
	\end{subfigure}
	\begin{subfigure}[t]{0.161\textwidth}
		\includegraphics[width=\textwidth]  
		{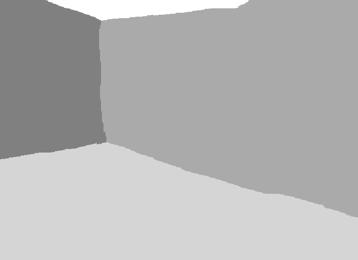}
		\includegraphics[width=\textwidth]
		{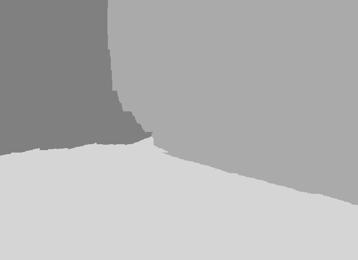}
		\includegraphics[width=\textwidth]
		{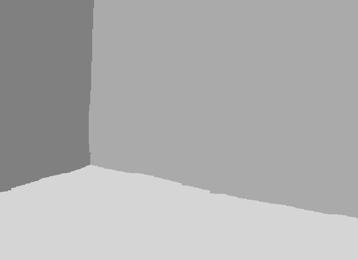}
		\includegraphics[width=\textwidth]
		{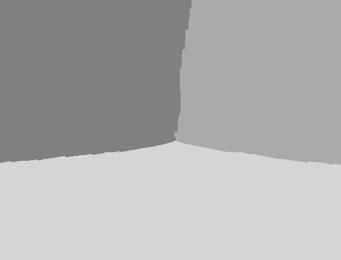}
		\includegraphics[width=\textwidth]  
		{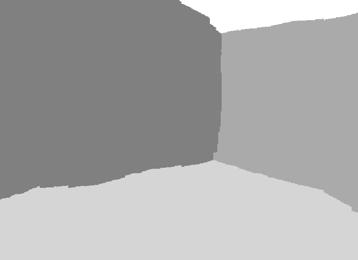}
		\includegraphics[width=\textwidth]
		{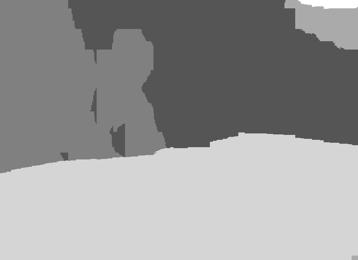}
		\includegraphics[width=\textwidth]
		{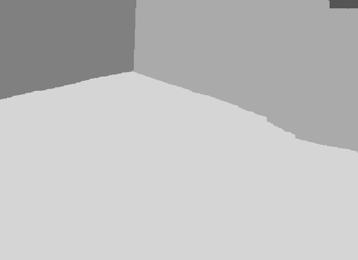}
		\includegraphics[width=\textwidth]
		{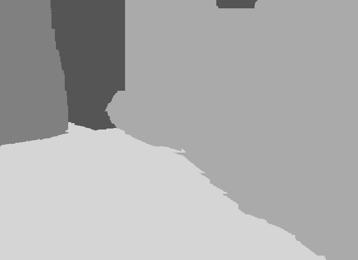}
		\includegraphics[width=\textwidth]
		{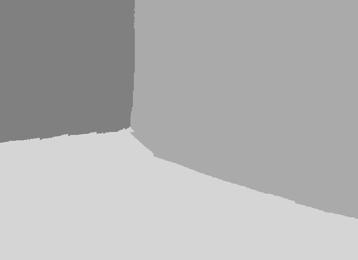}
		\includegraphics[width=\textwidth]
		{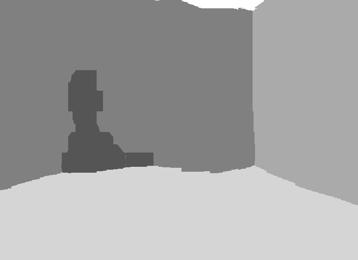}
		\includegraphics[width=\textwidth]
		{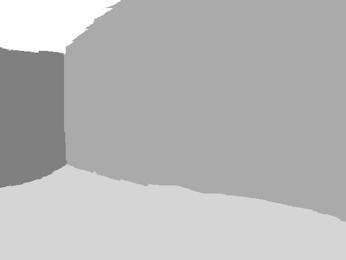}
		\caption{}
	\end{subfigure}
	\begin{subfigure}[t]{0.161\textwidth}
		\includegraphics[width=\textwidth]
		{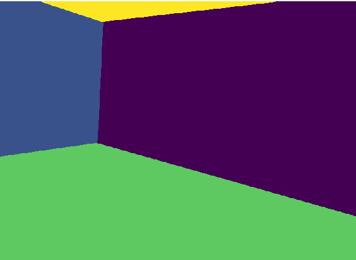}
		\includegraphics[width=\textwidth]  
		{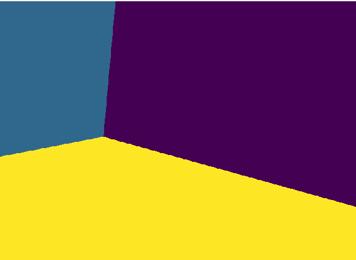}
		\includegraphics[width=\textwidth]
		{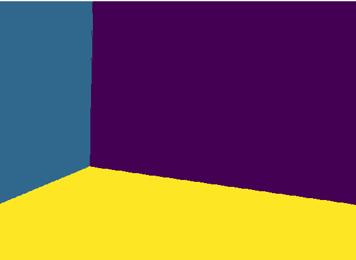}
		\includegraphics[width=\textwidth]
		{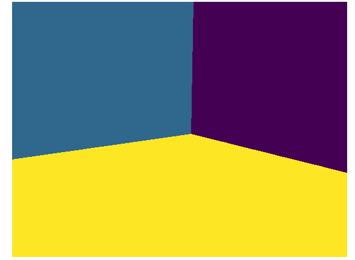}
		\includegraphics[width=\textwidth]
		{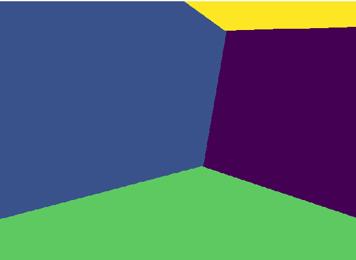}
		\includegraphics[width=\textwidth]  
		{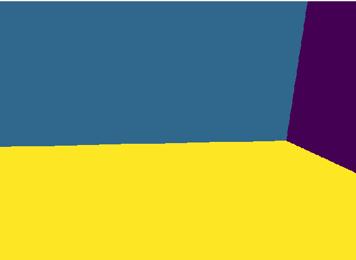}
		\includegraphics[width=\textwidth]
		{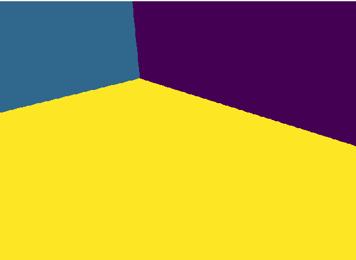}
		\includegraphics[width=\textwidth]
		{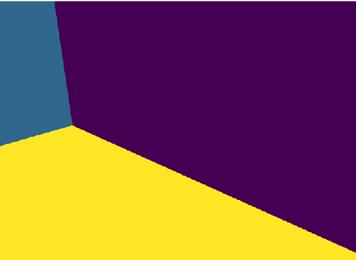}
		\includegraphics[width=\textwidth]
		{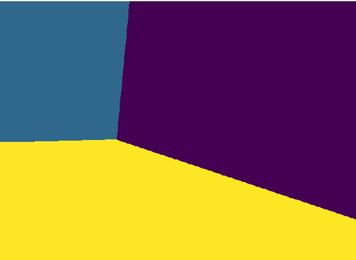}
		\includegraphics[width=\textwidth]
		{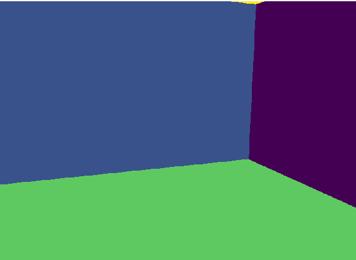}
		\includegraphics[width=\textwidth]
		{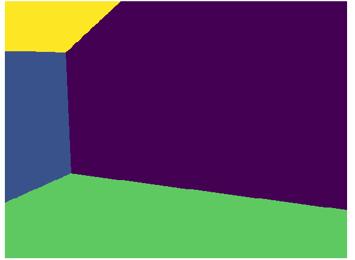}
		\caption{}
	\end{subfigure}
	\begin{subfigure}[t]{0.161\textwidth}
		\includegraphics[width=\textwidth]
		{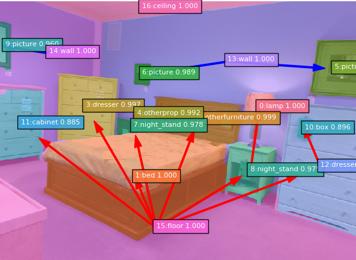}
		\includegraphics[width=\textwidth]  
		{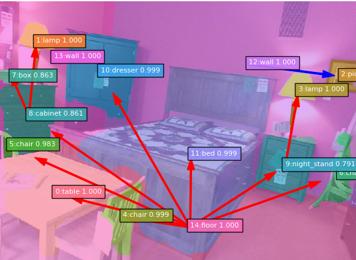}
		\includegraphics[width=\textwidth]
		{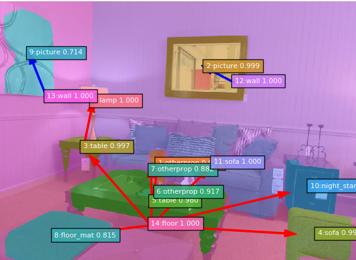}
		\includegraphics[width=\textwidth]
		{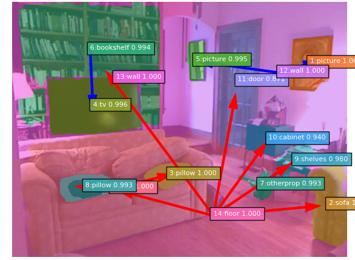}
		\includegraphics[width=\textwidth]
		{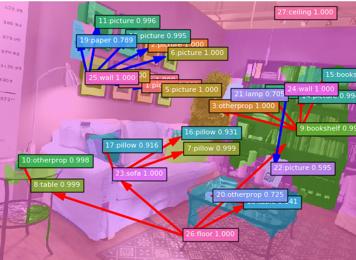}
		\includegraphics[width=\textwidth]  
		{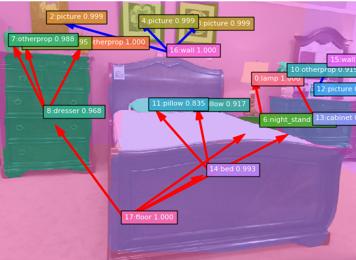}
		\includegraphics[width=\textwidth]
		{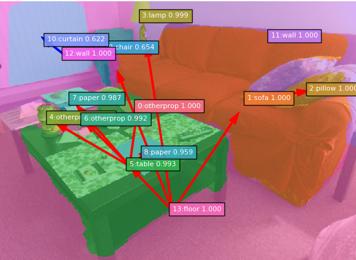}
		\includegraphics[width=\textwidth]
		{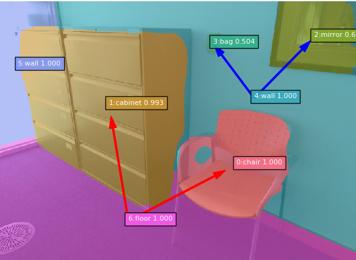}
		\includegraphics[width=\textwidth]
		{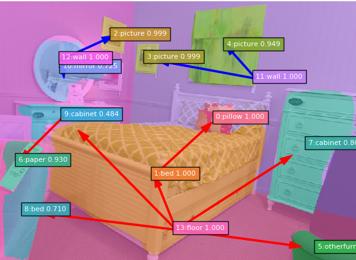}
		\includegraphics[width=\textwidth]
		{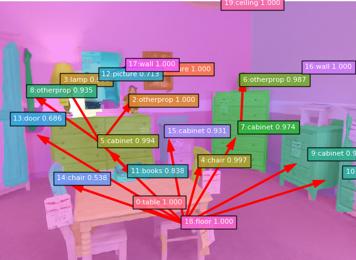}
		\includegraphics[width=\textwidth]
		{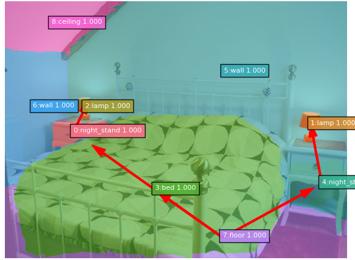}
		\caption{}
	\end{subfigure}
	\begin{subfigure}[t]{0.161\textwidth}
		\includegraphics[width=\textwidth]
		{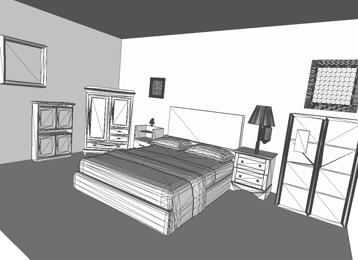}
		\includegraphics[width=\textwidth]  
		{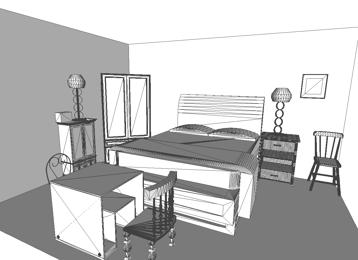}
		\includegraphics[width=\textwidth]
		{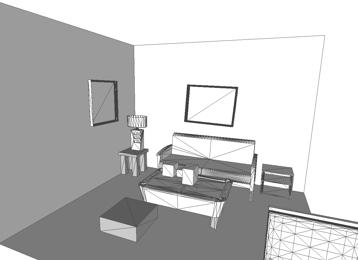}
		\includegraphics[width=\textwidth]
		{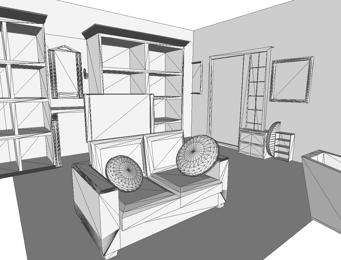}
		\includegraphics[width=\textwidth]
		{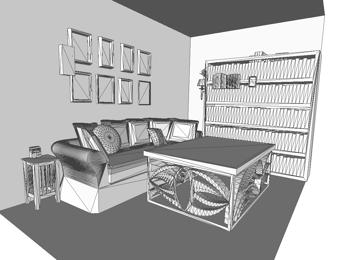}
		\includegraphics[width=\textwidth]  
		{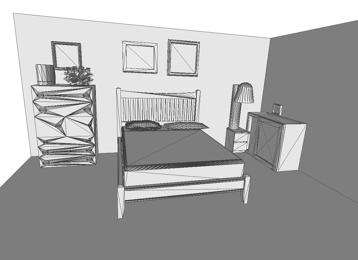}
		\includegraphics[width=\textwidth]
		{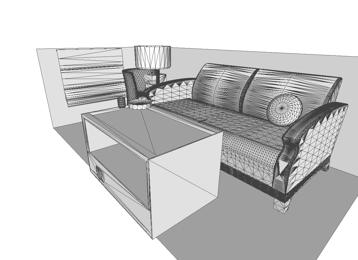}
		\includegraphics[width=\textwidth]
		{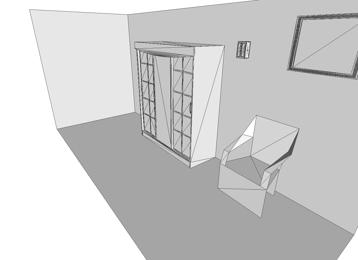}
		\includegraphics[width=\textwidth]
		{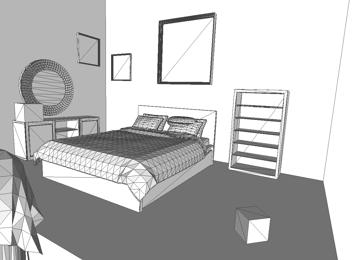}
		\includegraphics[width=\textwidth]
		{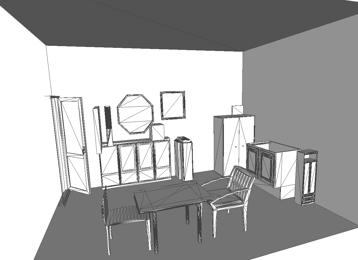}
		\includegraphics[width=\textwidth]
		{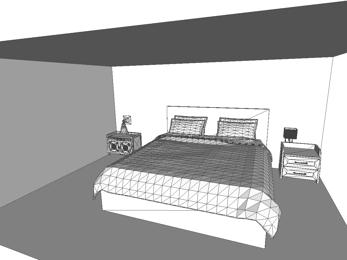}
		\caption{}
	\end{subfigure}
	Continue to the next page.
	\addtocounter{figure}{-1}
	\label{set1}
\end{figure}

\begin{figure}
	\centering
	\begin{subfigure}[t]{0.161\textwidth}
		\includegraphics[width=\textwidth]  
		{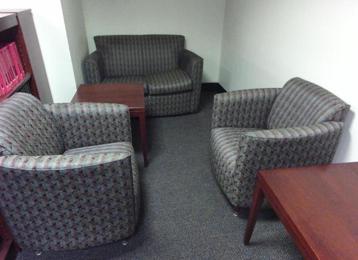}
		\includegraphics[width=\textwidth]
		{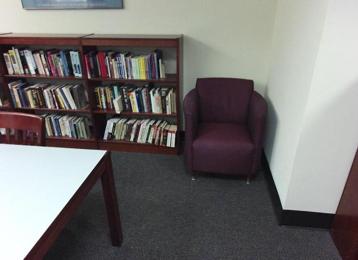}
		\includegraphics[width=\textwidth]
		{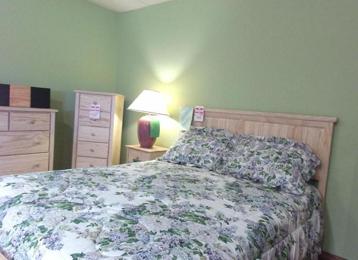}
		\includegraphics[width=\textwidth]
		{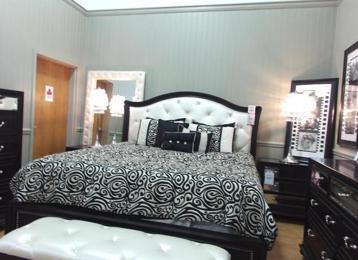}
		\includegraphics[width=\textwidth]
		{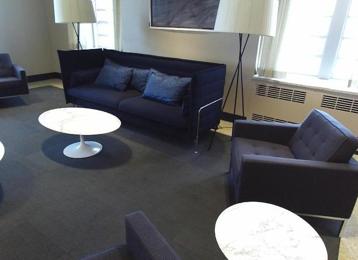}
		\includegraphics[width=\textwidth]
		{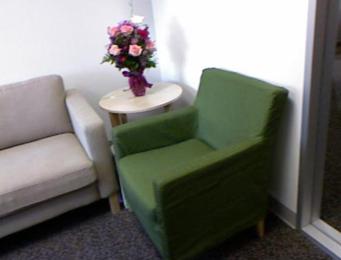}
		\includegraphics[width=\textwidth]
		{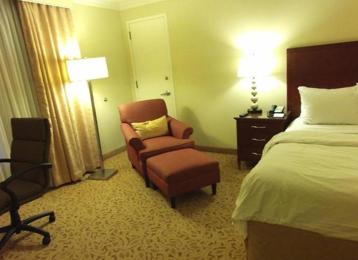}
		\includegraphics[width=\textwidth]
		{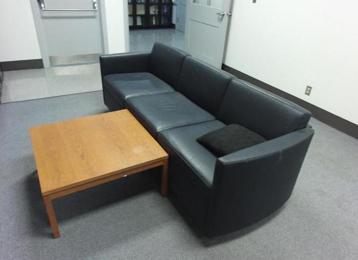}
		\includegraphics[width=\textwidth]
		{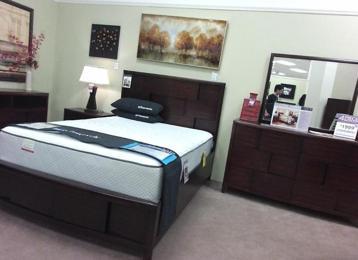}
		\includegraphics[width=\textwidth]
		{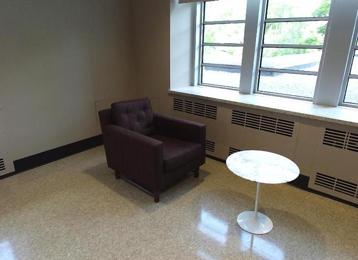}
		\includegraphics[width=\textwidth]
		{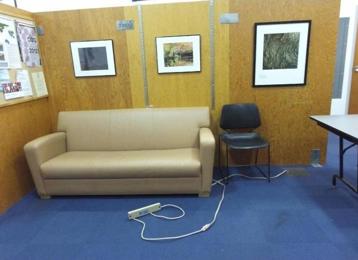}
		\caption{}
	\end{subfigure}
	\begin{subfigure}[t]{0.161\textwidth}
		\includegraphics[width=\textwidth]  
		{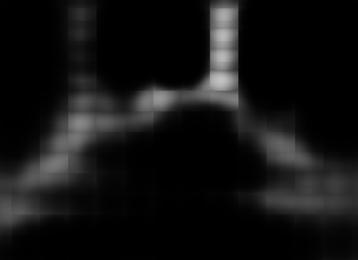}
		\includegraphics[width=\textwidth]
		{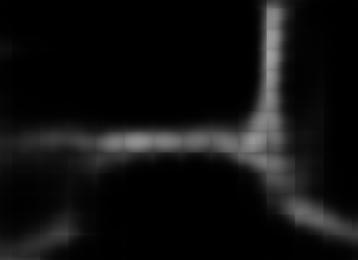}
		\includegraphics[width=\textwidth]
		{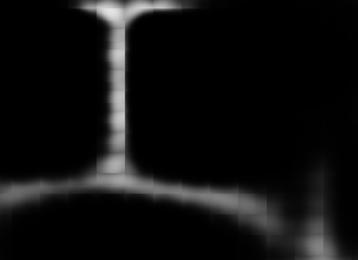}
		\includegraphics[width=\textwidth]
		{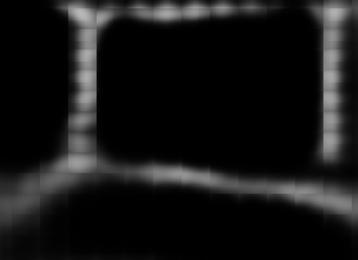}
		\includegraphics[width=\textwidth]
		{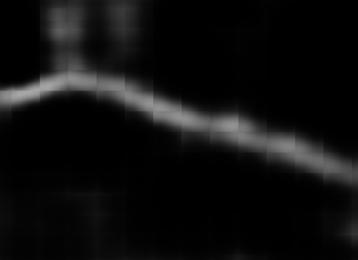}
		\includegraphics[width=\textwidth]
		{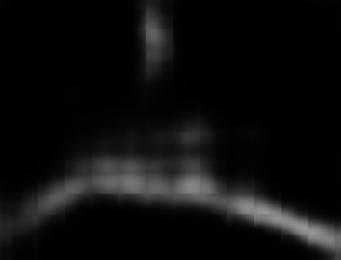}
		\includegraphics[width=\textwidth]
		{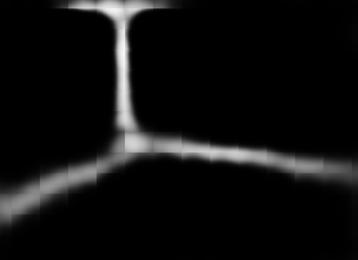}
		\includegraphics[width=\textwidth]
		{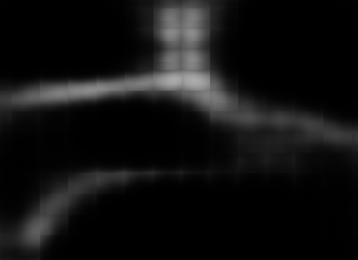}
		\includegraphics[width=\textwidth]
		{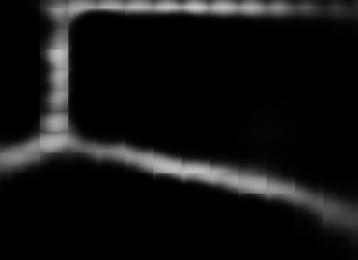}
		\includegraphics[width=\textwidth]
		{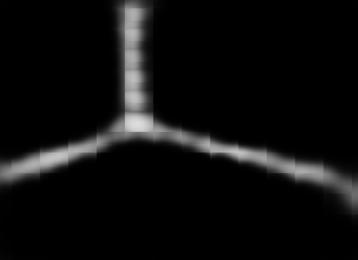}
		\includegraphics[width=\textwidth]
		{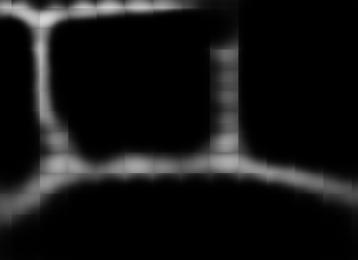}
		\caption{}
	\end{subfigure}
	\begin{subfigure}[t]{0.161\textwidth}
		\includegraphics[width=\textwidth]  
		{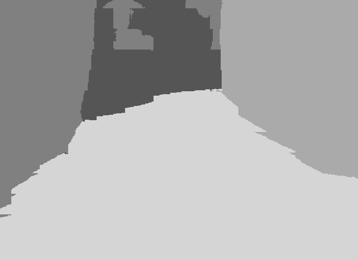}
		\includegraphics[width=\textwidth]
		{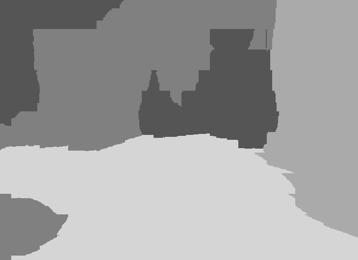}
		\includegraphics[width=\textwidth]
		{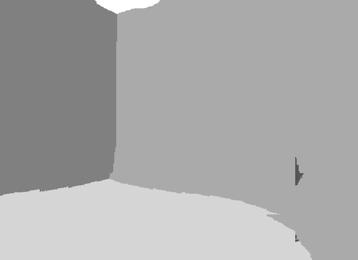}
		\includegraphics[width=\textwidth]
		{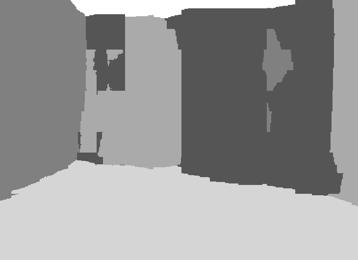}
		\includegraphics[width=\textwidth]
		{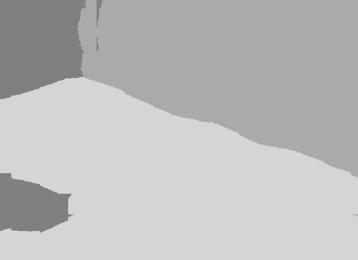}
		\includegraphics[width=\textwidth]
		{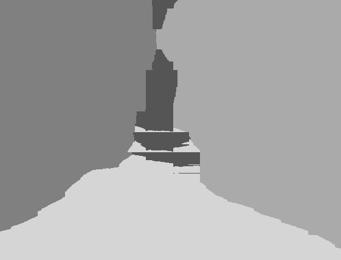}
		\includegraphics[width=\textwidth]
		{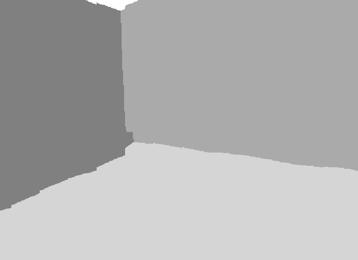}
		\includegraphics[width=\textwidth]
		{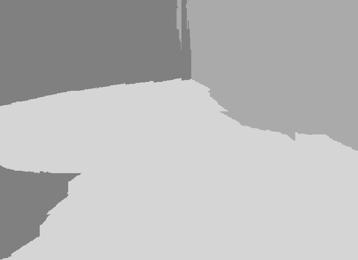}
		\includegraphics[width=\textwidth]
		{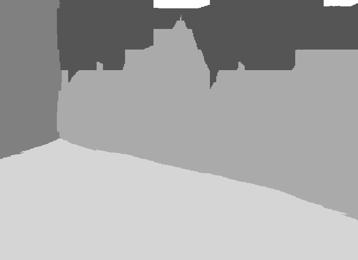}
		\includegraphics[width=\textwidth]
		{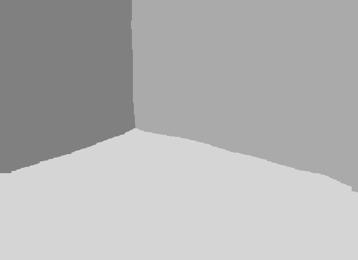}
		\includegraphics[width=\textwidth]
		{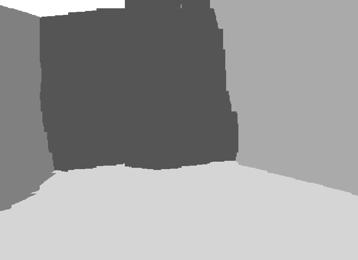}
		\caption{}
	\end{subfigure}
	\begin{subfigure}[t]{0.161\textwidth}
		\includegraphics[width=\textwidth]
		{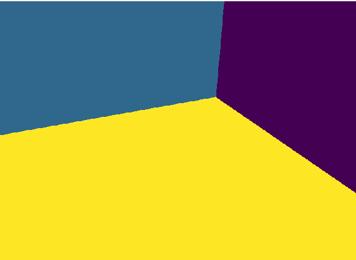}
		\includegraphics[width=\textwidth]  
		{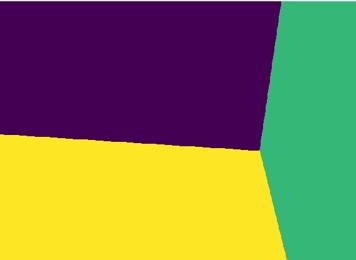}
		\includegraphics[width=\textwidth]
		{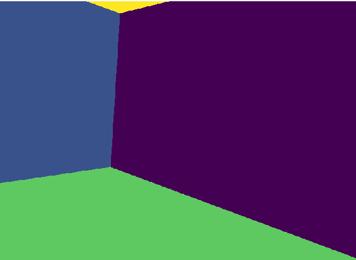}
		\includegraphics[width=\textwidth]
		{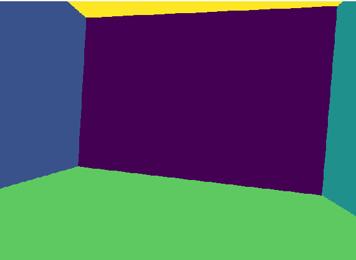}
		\includegraphics[width=\textwidth]
		{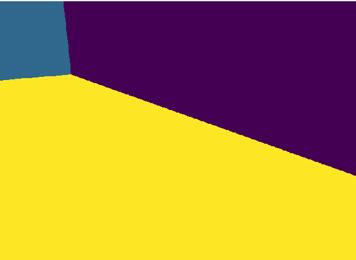}
		\includegraphics[width=\textwidth]
		{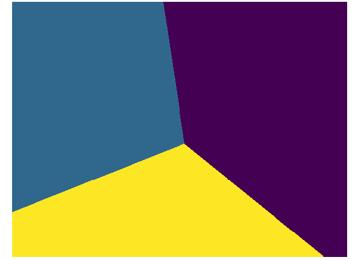}
		\includegraphics[width=\textwidth]
		{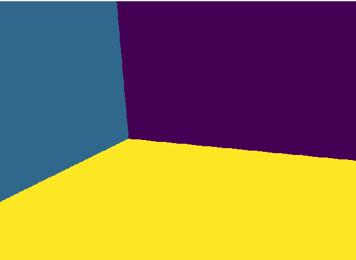}
		\includegraphics[width=\textwidth]
		{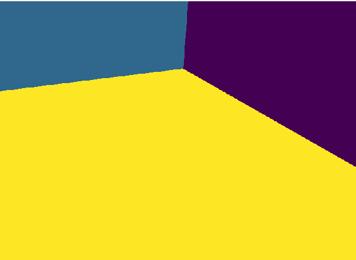}
		\includegraphics[width=\textwidth]
		{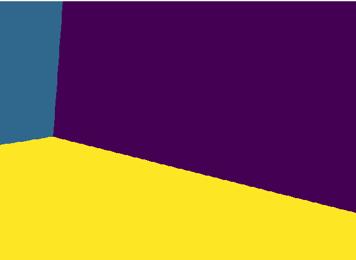}
		\includegraphics[width=\textwidth]
		{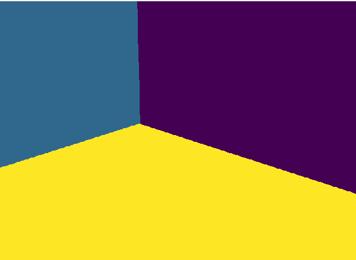}
		\includegraphics[width=\textwidth]
		{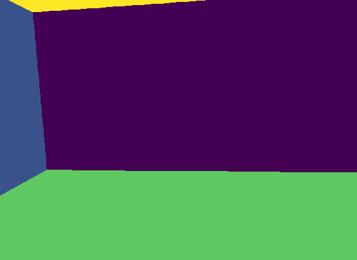}
		\caption{}
	\end{subfigure}
	\begin{subfigure}[t]{0.161\textwidth}
		\includegraphics[width=\textwidth]
		{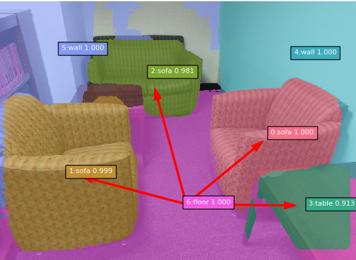}
		\includegraphics[width=\textwidth]  
		{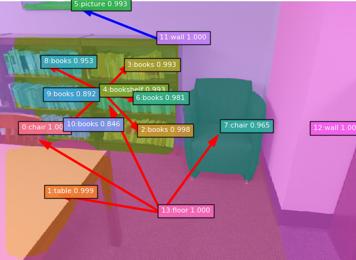}
		\includegraphics[width=\textwidth]
		{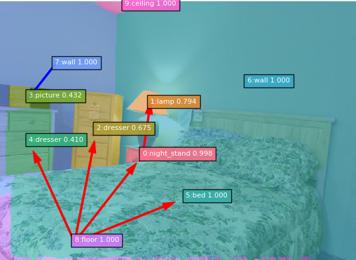}
		\includegraphics[width=\textwidth]
		{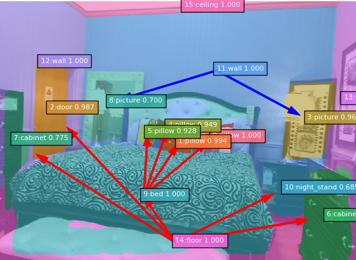}
		\includegraphics[width=\textwidth]
		{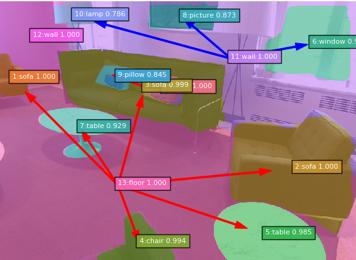}
		\includegraphics[width=\textwidth]
		{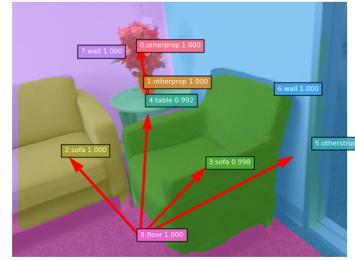}
		\includegraphics[width=\textwidth]
		{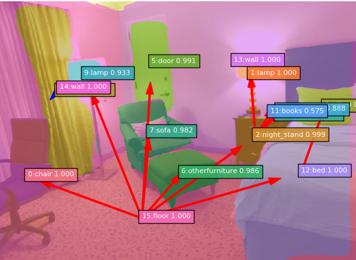}
		\includegraphics[width=\textwidth]
		{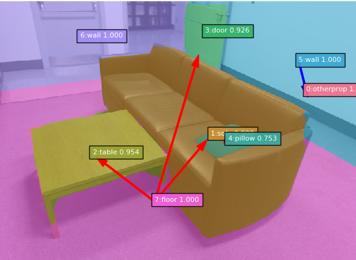}
		\includegraphics[width=\textwidth]
		{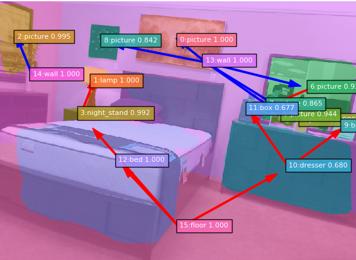}
		\includegraphics[width=\textwidth]
		{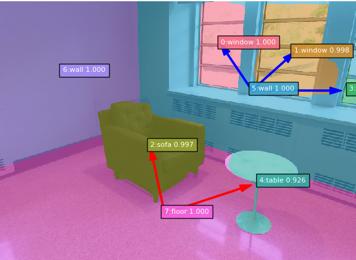}
		\includegraphics[width=\textwidth]
		{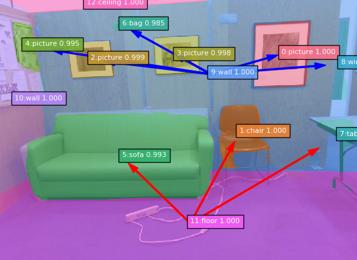}
		\caption{}
	\end{subfigure}
	\begin{subfigure}[t]{0.161\textwidth}
		\includegraphics[width=\textwidth]
		{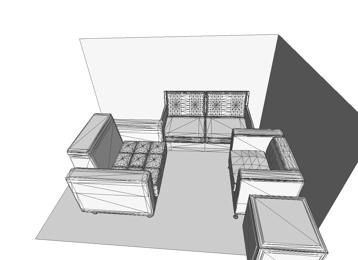}
		\includegraphics[width=\textwidth]  
		{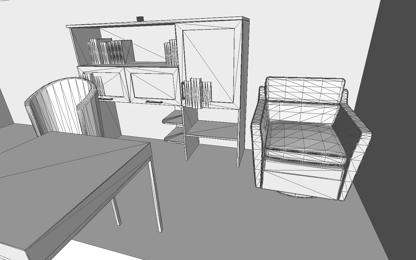}
		\includegraphics[width=\textwidth]
		{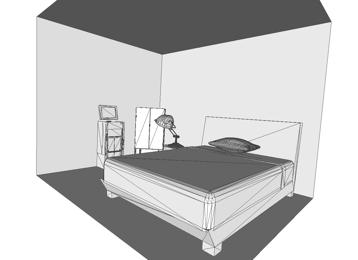}
		\includegraphics[width=\textwidth]
		{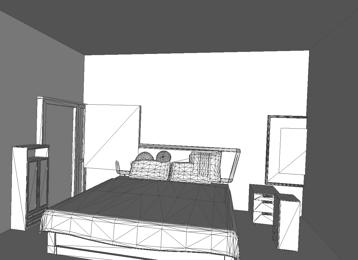}
		\includegraphics[width=\textwidth]
		{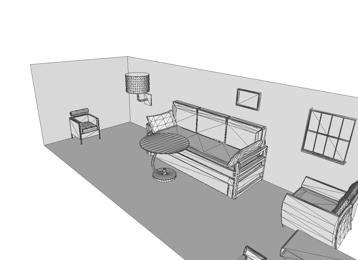}
		\includegraphics[width=\textwidth]
		{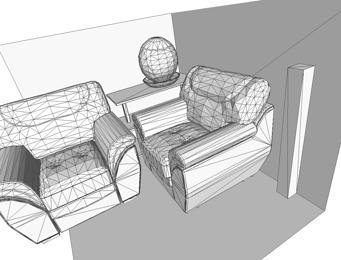}
		\includegraphics[width=\textwidth]
		{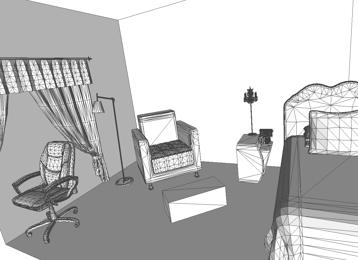}
		\includegraphics[width=\textwidth]
		{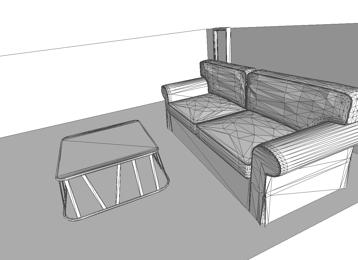}
		\includegraphics[width=\textwidth]
		{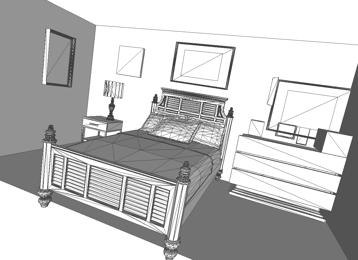}
		\includegraphics[width=\textwidth]
		{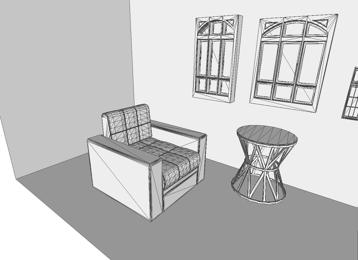}
		\includegraphics[width=\textwidth]
		{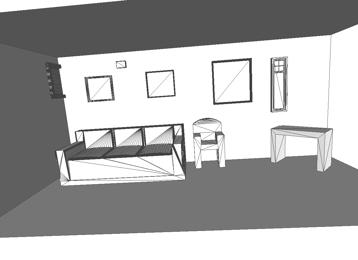}
		\caption{}
	\end{subfigure}
	Continue to the next page.
	\addtocounter{figure}{-1}
	\label{set2}
\end{figure}

\begin{figure}
	\centering
	\begin{subfigure}[t]{0.161\textwidth}
		\includegraphics[width=\textwidth]  
		{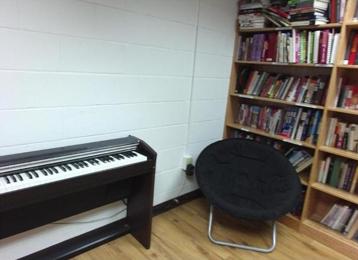}
		\includegraphics[width=\textwidth]
		{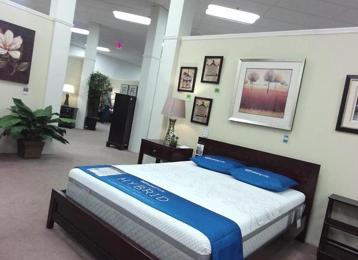}
		\includegraphics[width=\textwidth]
		{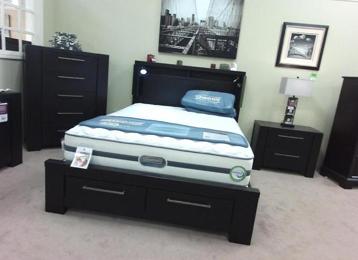}
		\includegraphics[width=\textwidth]
		{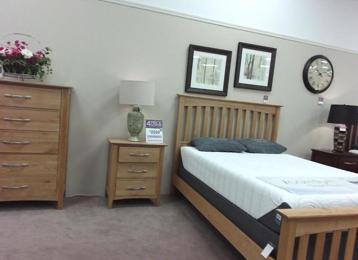}
		\includegraphics[width=\textwidth]
		{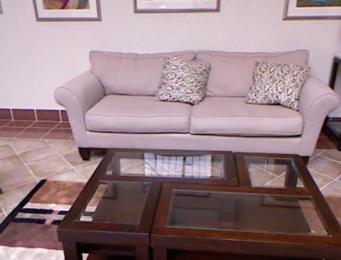}
		\includegraphics[width=\textwidth]
		{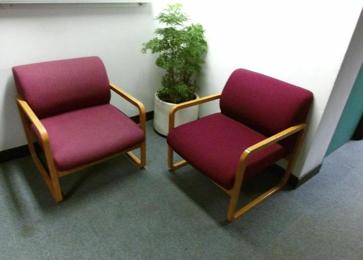}
		\includegraphics[width=\textwidth]
		{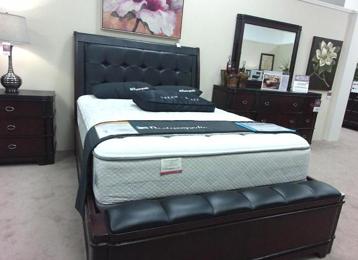}
		\includegraphics[width=\textwidth]
		{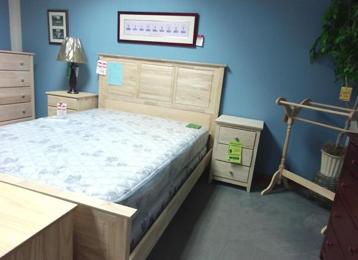}
		\includegraphics[width=\textwidth]
		{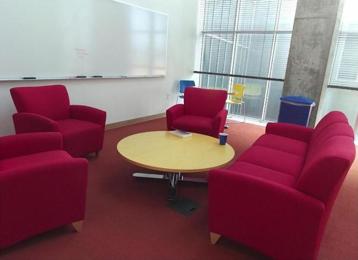}
		\includegraphics[width=\textwidth]
		{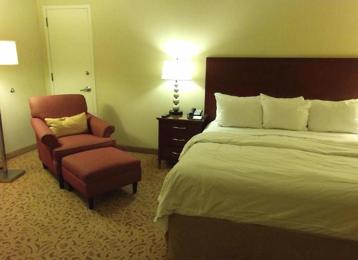}
		\includegraphics[width=\textwidth]
		{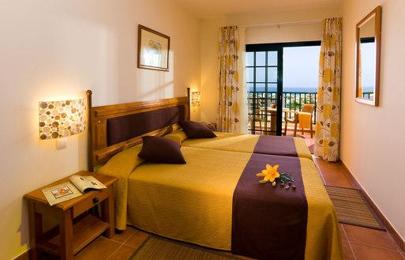}
		\caption{}
	\end{subfigure}
	\begin{subfigure}[t]{0.161\textwidth}
		\includegraphics[width=\textwidth]  
		{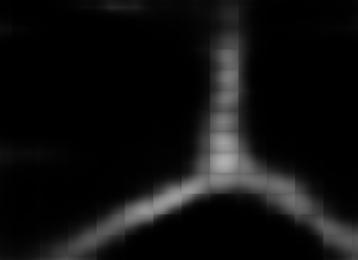}
		\includegraphics[width=\textwidth]
		{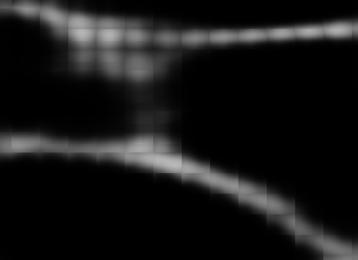}
		\includegraphics[width=\textwidth]
		{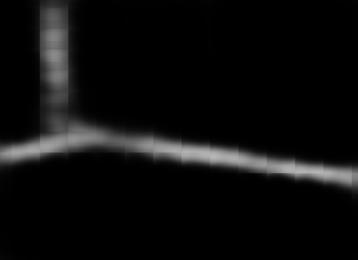}
		\includegraphics[width=\textwidth]
		{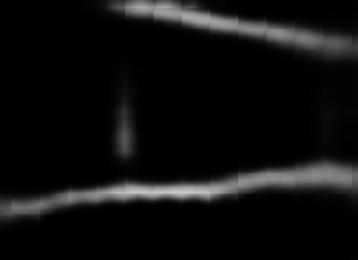}
		\includegraphics[width=\textwidth]
		{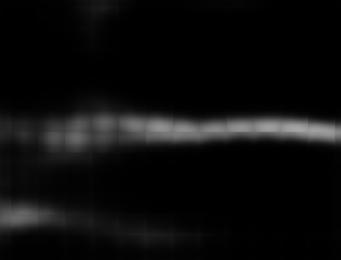}
		\includegraphics[width=\textwidth]
		{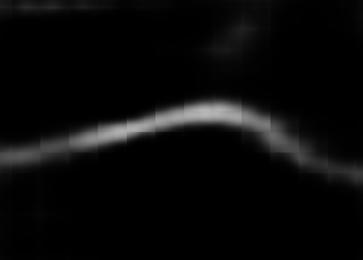}
		\includegraphics[width=\textwidth]
		{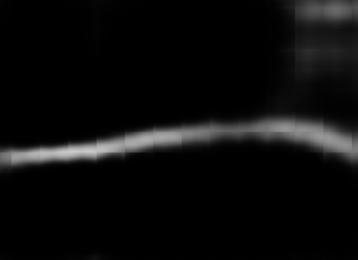}
		\includegraphics[width=\textwidth]
		{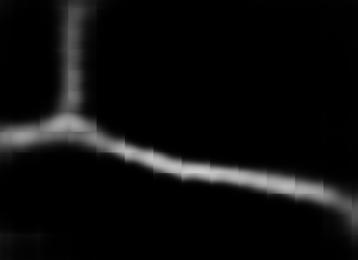}
		\includegraphics[width=\textwidth]
		{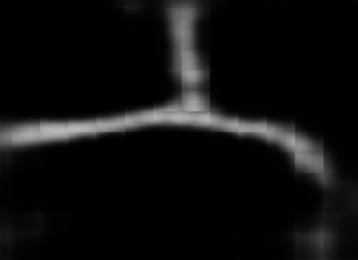}
		\includegraphics[width=\textwidth]
		{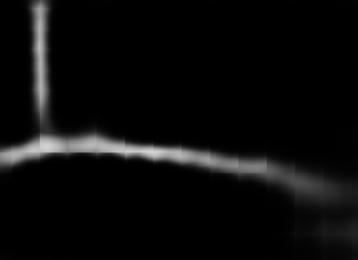}
		\includegraphics[width=\textwidth]
		{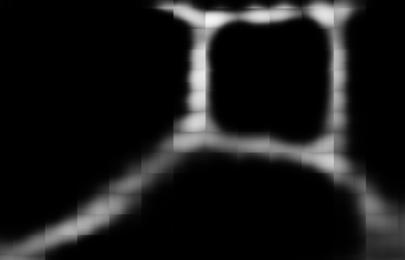}
		\caption{}
	\end{subfigure}
	\begin{subfigure}[t]{0.161\textwidth}
		\includegraphics[width=\textwidth]  
		{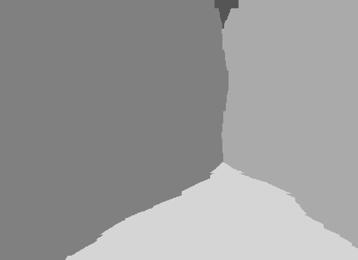}
		\includegraphics[width=\textwidth]
		{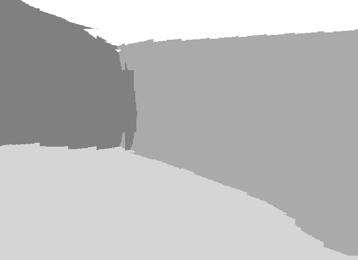}
		\includegraphics[width=\textwidth]
		{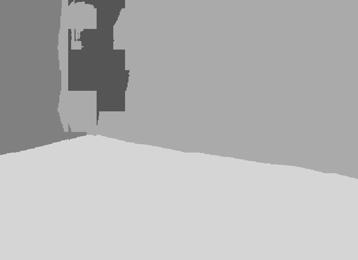}
		\includegraphics[width=\textwidth]
		{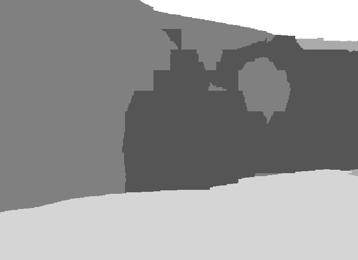}
		\includegraphics[width=\textwidth]
		{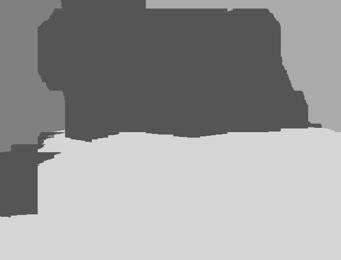}
		\includegraphics[width=\textwidth]
		{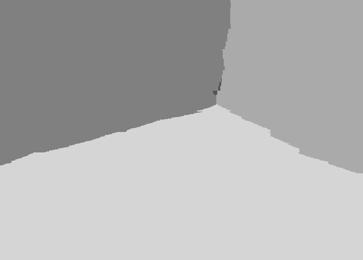}
		\includegraphics[width=\textwidth]
		{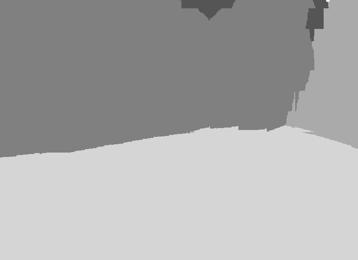}
		\includegraphics[width=\textwidth]
		{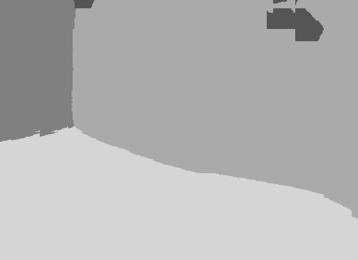}
		\includegraphics[width=\textwidth]
		{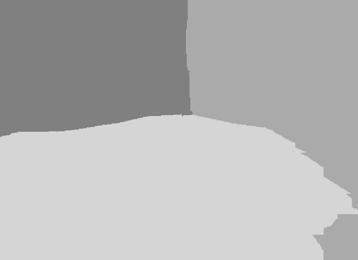}
		\includegraphics[width=\textwidth]
		{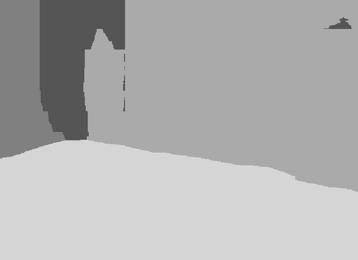}
		\includegraphics[width=\textwidth]
		{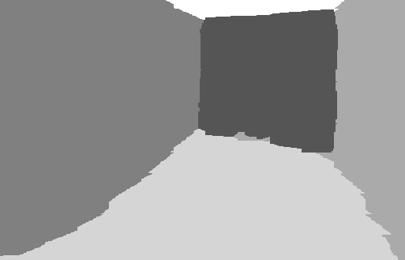}
		\caption{}
	\end{subfigure}
	\begin{subfigure}[t]{0.161\textwidth}
		\includegraphics[width=\textwidth]
		{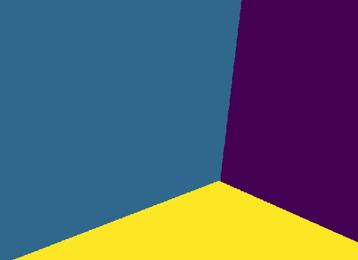}
		\includegraphics[width=\textwidth]  
		{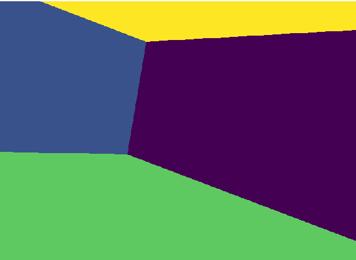}
		\includegraphics[width=\textwidth]
		{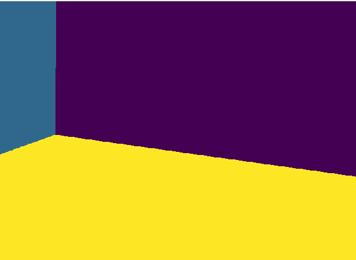}
		\includegraphics[width=\textwidth]
		{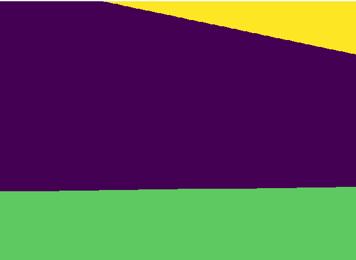}
		\includegraphics[width=\textwidth]
		{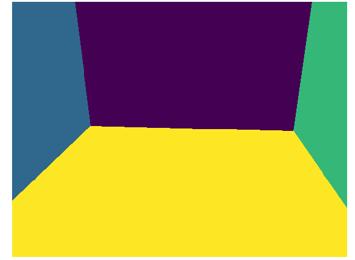}
		\includegraphics[width=\textwidth]
		{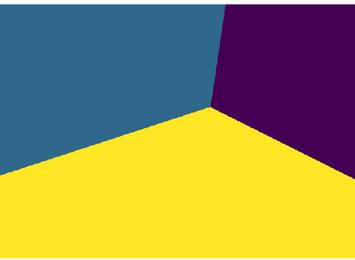}
		\includegraphics[width=\textwidth]
		{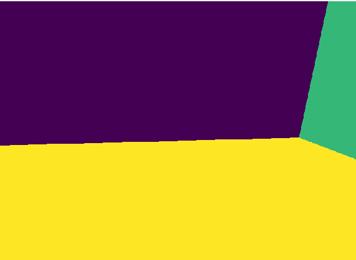}
		\includegraphics[width=\textwidth]
		{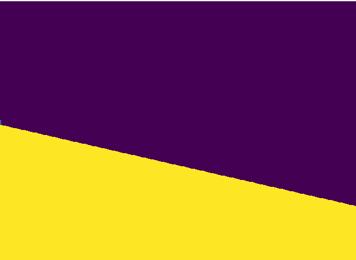}
		\includegraphics[width=\textwidth]
		{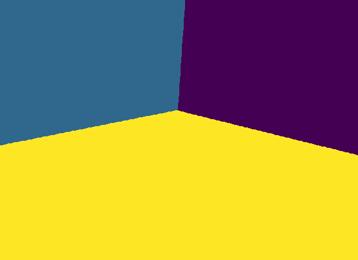}
		\includegraphics[width=\textwidth]
		{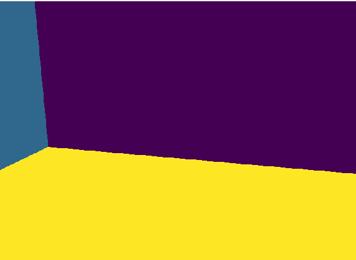}
		\includegraphics[width=\textwidth]
		{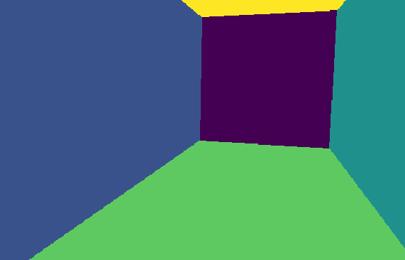}
		\caption{}
	\end{subfigure}
	\begin{subfigure}[t]{0.161\textwidth}
		\includegraphics[width=\textwidth]
		{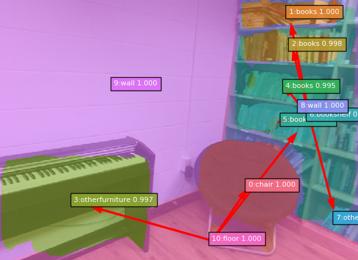}
		\includegraphics[width=\textwidth]  
		{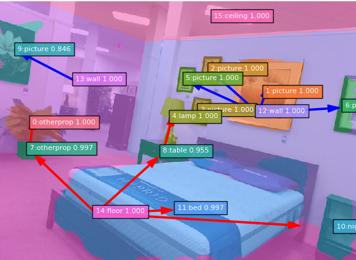}
		\includegraphics[width=\textwidth]
		{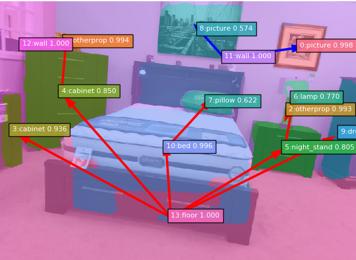}
		\includegraphics[width=\textwidth]
		{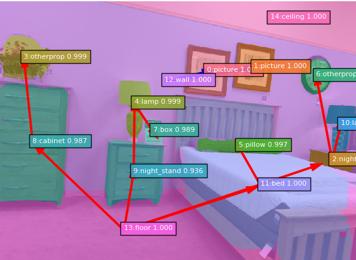}
		\includegraphics[width=\textwidth]
		{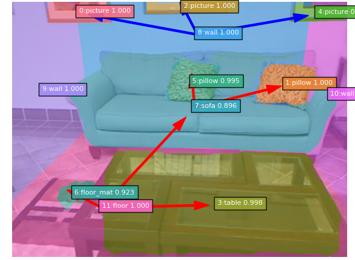}
		\includegraphics[width=\textwidth]
		{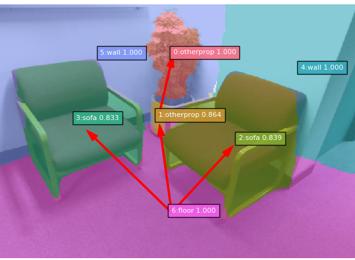}
		\includegraphics[width=\textwidth]
		{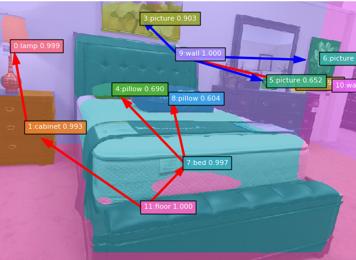}
		\includegraphics[width=\textwidth]
		{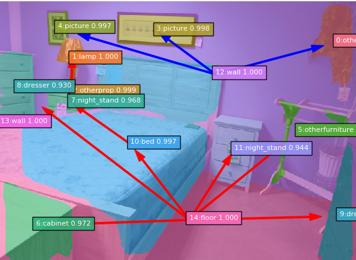}
		\includegraphics[width=\textwidth]
		{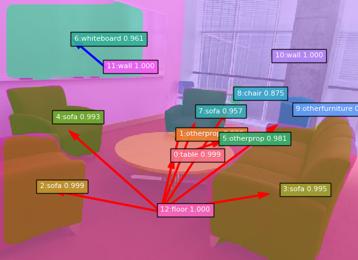}
		\includegraphics[width=\textwidth]
		{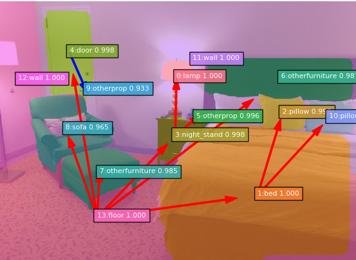}
		\includegraphics[width=\textwidth]
		{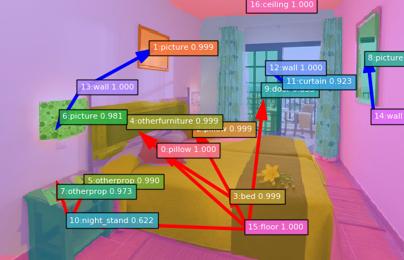}
		\caption{}
	\end{subfigure}
	\begin{subfigure}[t]{0.161\textwidth}
		\includegraphics[width=\textwidth]
		{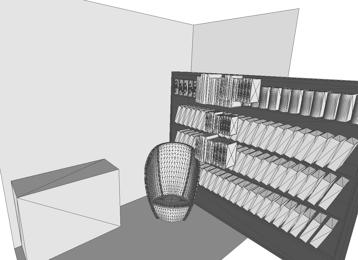}
		\includegraphics[width=\textwidth]  
		{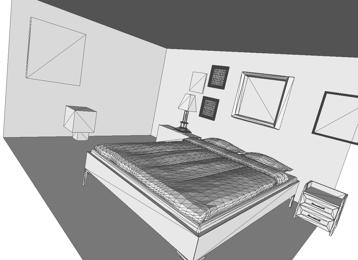}
		\includegraphics[width=\textwidth]
		{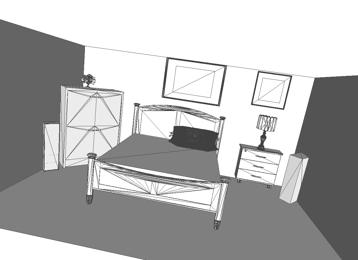}
		\includegraphics[width=\textwidth]
		{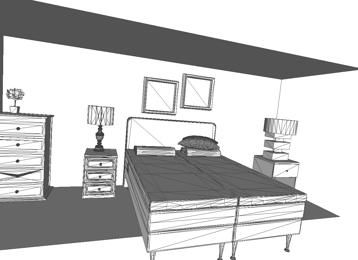}
		\includegraphics[width=\textwidth]
		{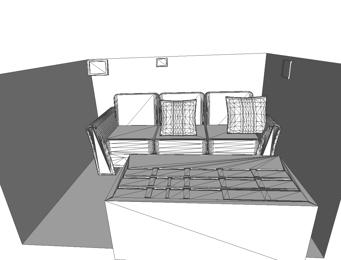}
		\includegraphics[width=\textwidth]
		{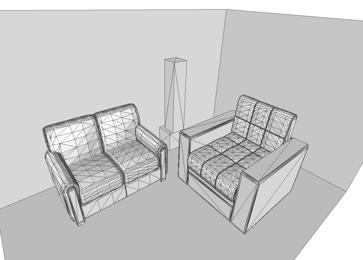}
		\includegraphics[width=\textwidth]
		{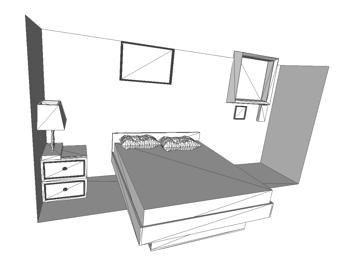}
		\includegraphics[width=\textwidth]
		{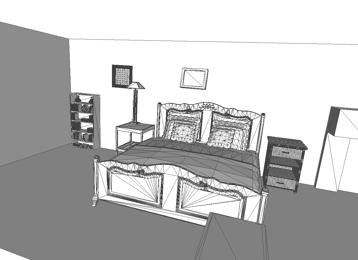}
		\includegraphics[width=\textwidth]
		{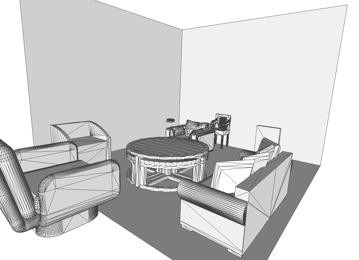}
		\includegraphics[width=\textwidth]
		{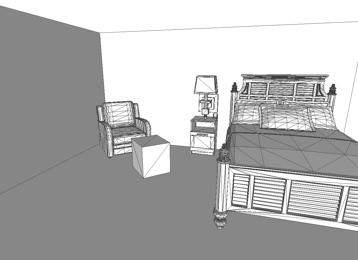}
		\includegraphics[width=\textwidth]
		{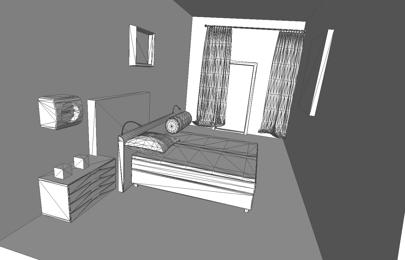}
		\caption{}
	\end{subfigure}
	Continue to the next page.
	\addtocounter{figure}{-1}
	\label{set3}
\end{figure}

\begin{figure}
	\centering
	\begin{subfigure}[t]{0.161\textwidth}
		\includegraphics[width=\textwidth]  
		{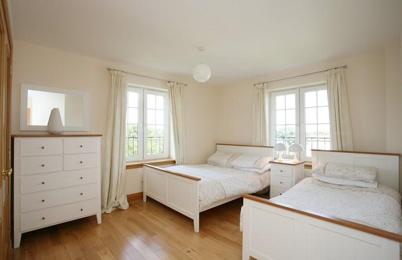}
		\includegraphics[width=\textwidth]
		{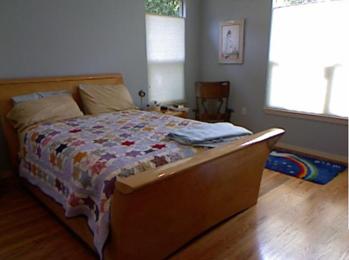}
		\includegraphics[width=\textwidth]
		{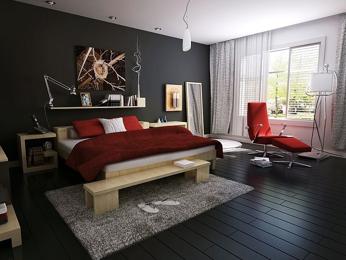}
		\includegraphics[width=\textwidth]
		{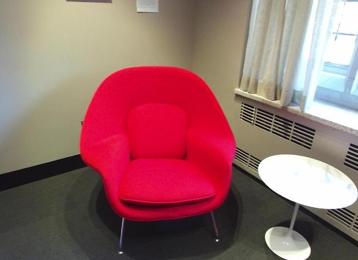}
		\includegraphics[width=\textwidth]
		{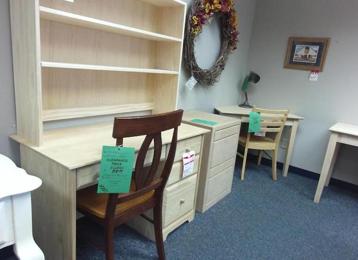}
		\includegraphics[width=\textwidth]
		{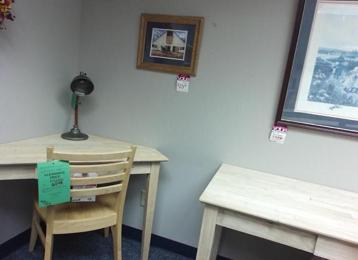}
		\includegraphics[width=\textwidth]
		{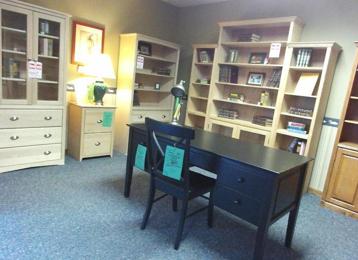}
		\includegraphics[width=\textwidth]
		{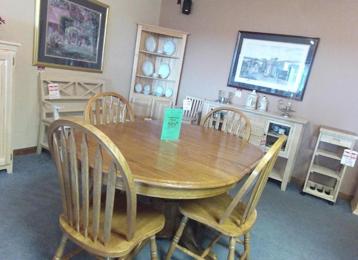}
		\includegraphics[width=\textwidth]
		{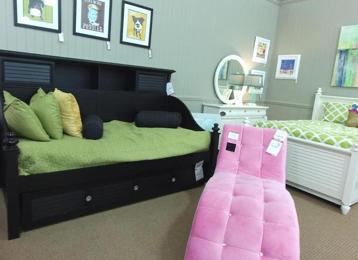}
		\includegraphics[width=\textwidth]
		{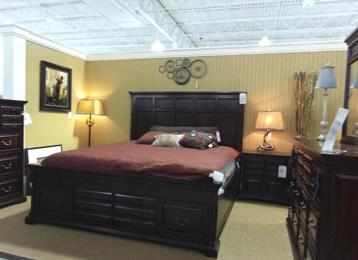}
		\includegraphics[width=\textwidth]
		{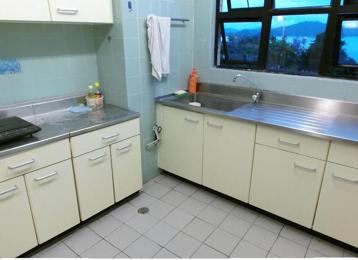}
		\caption{}
	\end{subfigure}
	\begin{subfigure}[t]{0.161\textwidth}
		\includegraphics[width=\textwidth]  
		{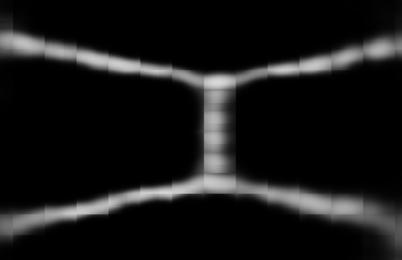}
		\includegraphics[width=\textwidth]
		{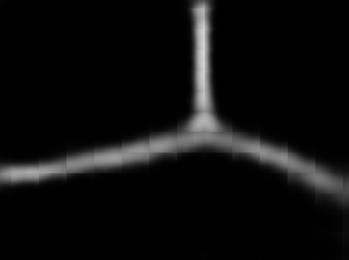}
		\includegraphics[width=\textwidth]
		{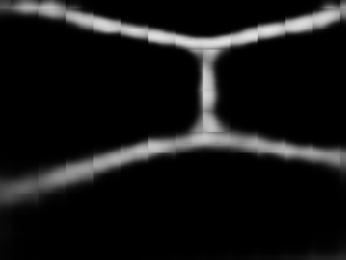}
		\includegraphics[width=\textwidth]
		{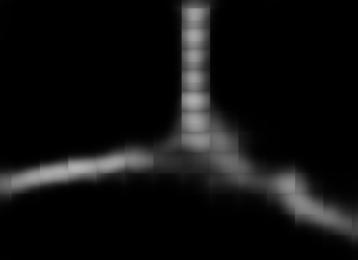}
		\includegraphics[width=\textwidth]
		{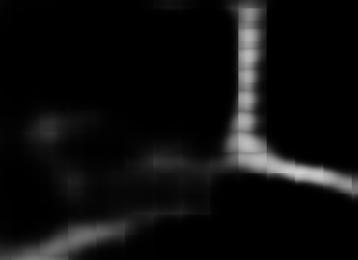}
		\includegraphics[width=\textwidth]
		{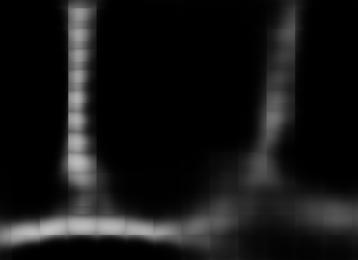}
		\includegraphics[width=\textwidth]
		{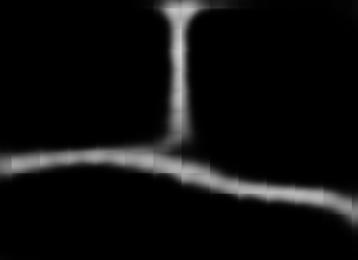}
		\includegraphics[width=\textwidth]
		{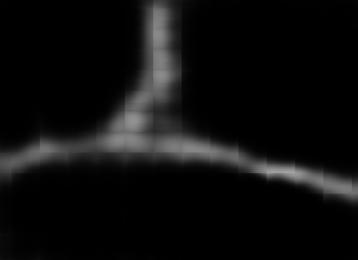}
		\includegraphics[width=\textwidth]
		{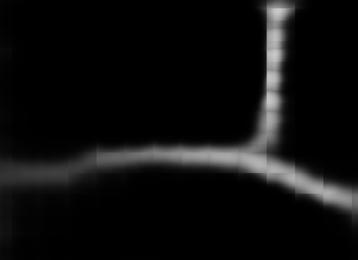}
		\includegraphics[width=\textwidth]
		{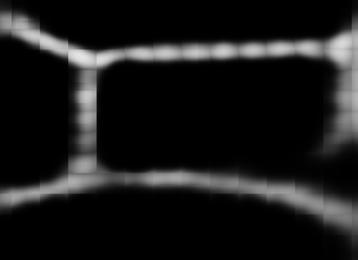}
		\includegraphics[width=\textwidth]
		{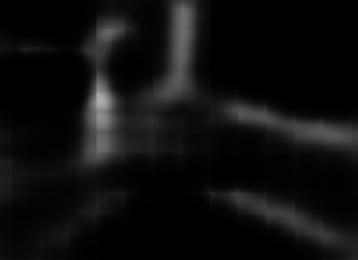}
		\caption{}
	\end{subfigure}
	\begin{subfigure}[t]{0.161\textwidth}
		\includegraphics[width=\textwidth]  
		{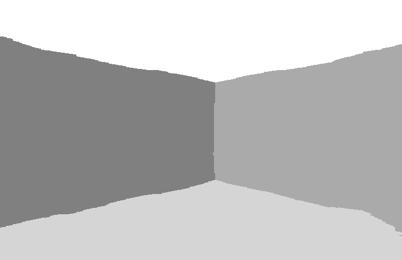}
		\includegraphics[width=\textwidth]
		{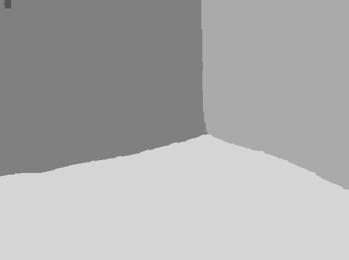}
		\includegraphics[width=\textwidth]
		{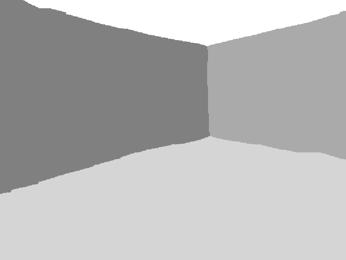}
		\includegraphics[width=\textwidth]
		{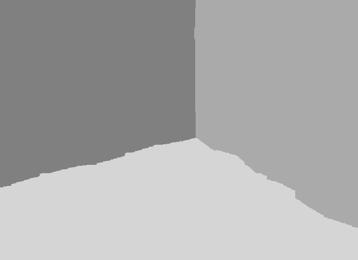}
		\includegraphics[width=\textwidth]
		{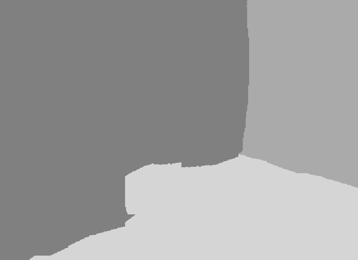}
		\includegraphics[width=\textwidth]
		{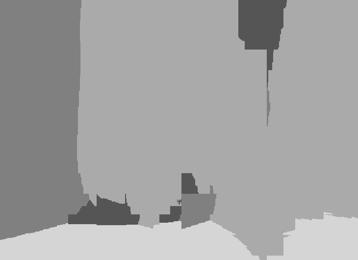}
		\includegraphics[width=\textwidth]
		{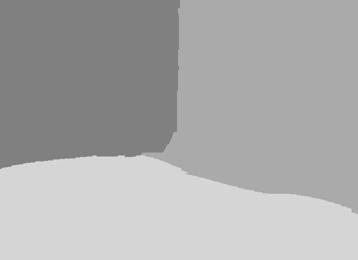}
		\includegraphics[width=\textwidth]
		{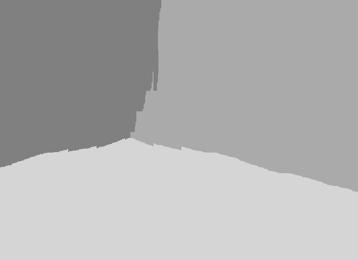}
		\includegraphics[width=\textwidth]
		{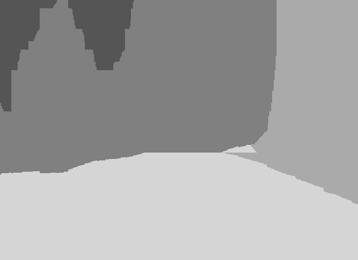}
		\includegraphics[width=\textwidth]
		{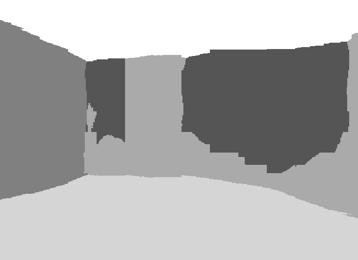}
		\includegraphics[width=\textwidth]
		{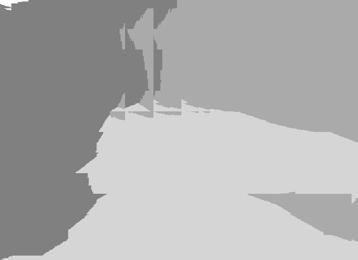}
		\caption{}
	\end{subfigure}
	\begin{subfigure}[t]{0.161\textwidth}
		\includegraphics[width=\textwidth]  
		{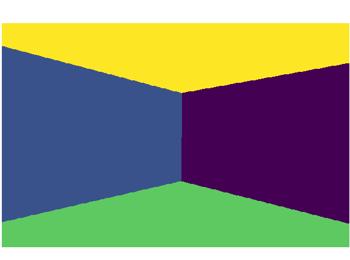}
		\includegraphics[width=\textwidth]
		{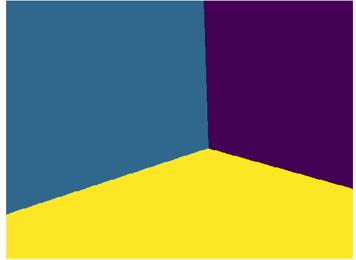}
		\includegraphics[width=\textwidth]
		{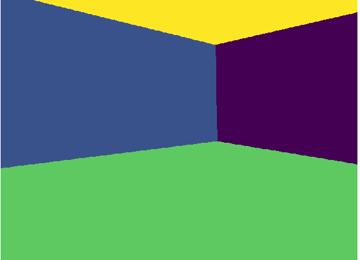}
		\includegraphics[width=\textwidth]
		{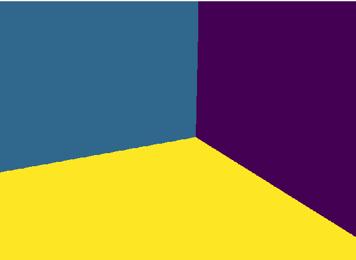}
		\includegraphics[width=\textwidth]
		{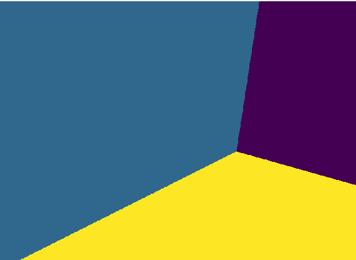}
		\includegraphics[width=\textwidth]
		{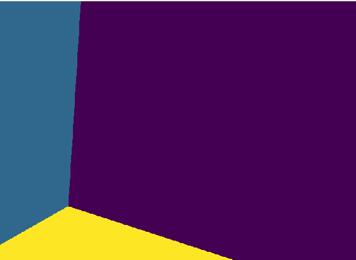}
		\includegraphics[width=\textwidth]
		{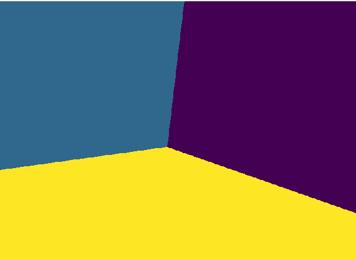}
		\includegraphics[width=\textwidth]
		{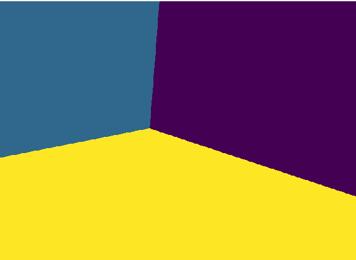}
		\includegraphics[width=\textwidth]
		{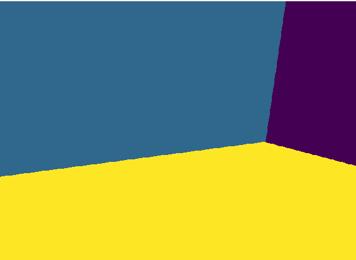}
		\includegraphics[width=\textwidth]
		{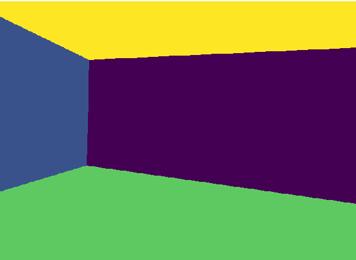}
		\includegraphics[width=\textwidth]
		{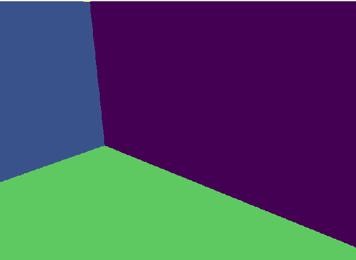}
		\caption{}
	\end{subfigure}
	\begin{subfigure}[t]{0.161\textwidth}
		\includegraphics[width=\textwidth]  
		{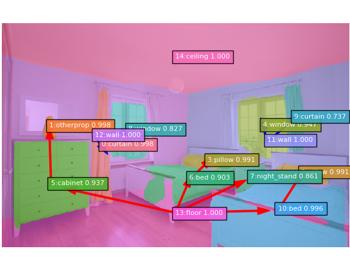}
		\includegraphics[width=\textwidth]
		{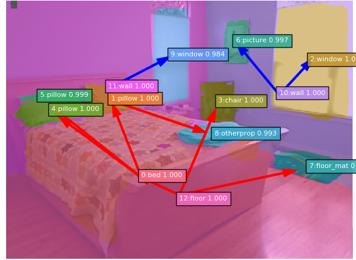}
		\includegraphics[width=\textwidth]
		{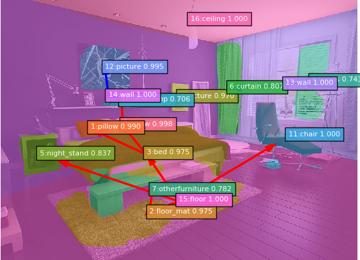}
		\includegraphics[width=\textwidth]
		{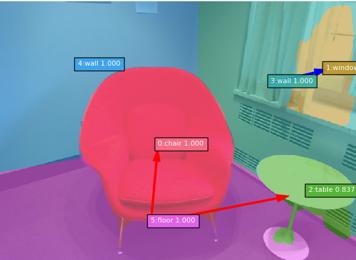}
		\includegraphics[width=\textwidth]
		{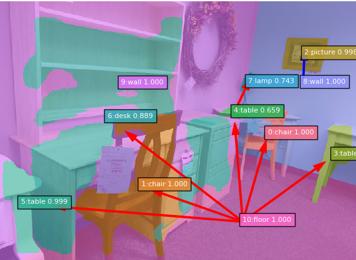}
		\includegraphics[width=\textwidth]
		{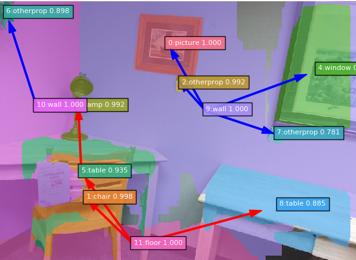}
		\includegraphics[width=\textwidth]
		{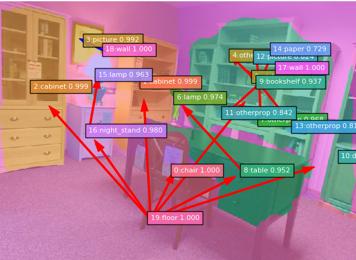}
		\includegraphics[width=\textwidth]
		{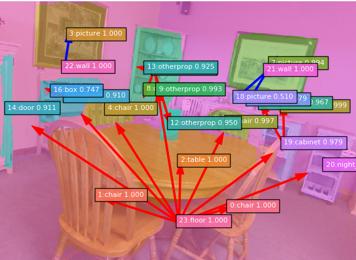}
		\includegraphics[width=\textwidth]
		{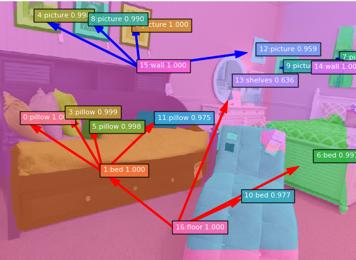}
		\includegraphics[width=\textwidth]
		{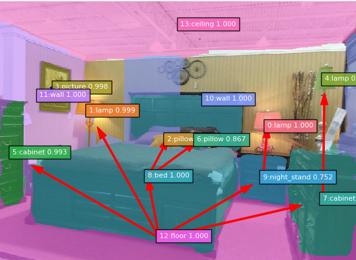}
		\includegraphics[width=\textwidth]
		{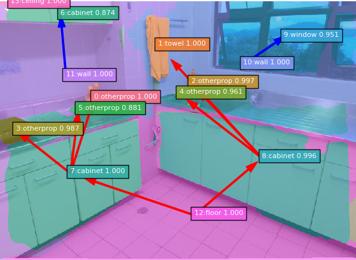}
		\caption{}
	\end{subfigure}
	\begin{subfigure}[t]{0.161\textwidth}
		\includegraphics[width=\textwidth]  
		{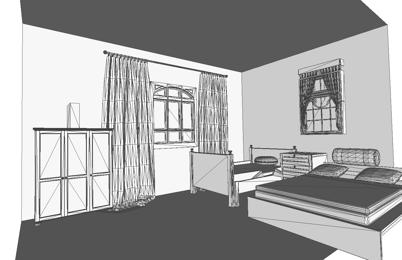}
		\includegraphics[width=\textwidth]
		{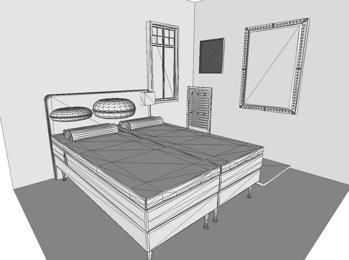}
		\includegraphics[width=\textwidth]
		{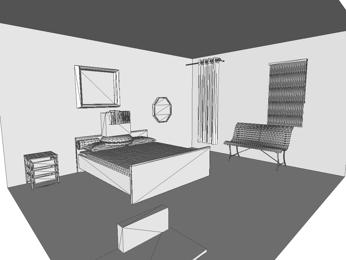}
		\includegraphics[width=\textwidth]
		{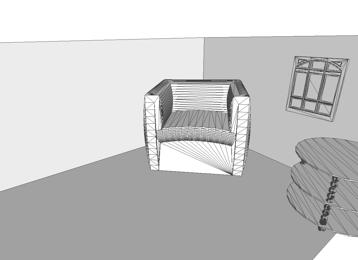}
		\includegraphics[width=\textwidth]
		{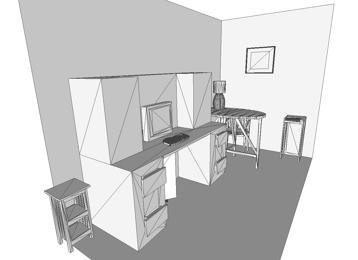}
		\includegraphics[width=\textwidth]
		{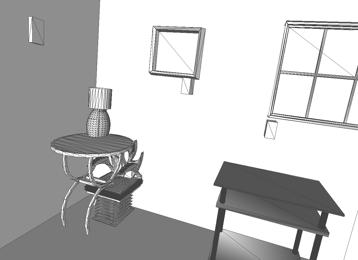}
		\includegraphics[width=\textwidth]
		{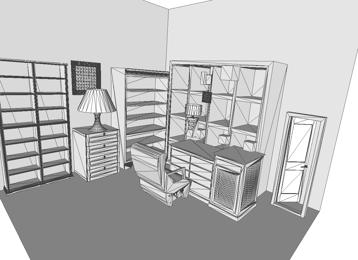}
		\includegraphics[width=\textwidth]
		{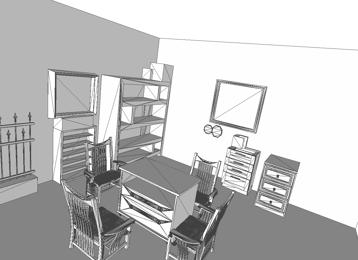}
		\includegraphics[width=\textwidth]
		{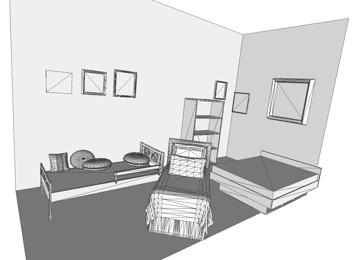}
		\includegraphics[width=\textwidth]
		{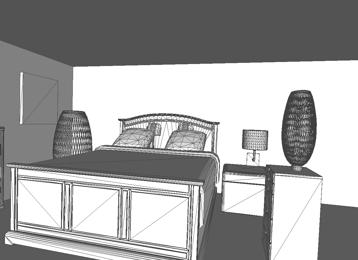}
		\includegraphics[width=\textwidth]
		{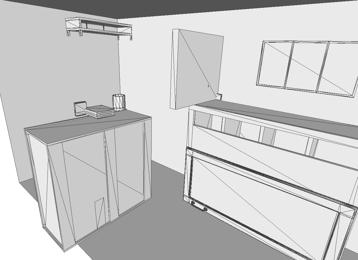}
		\caption{}
	\end{subfigure}
	Continue to the next page.
	\addtocounter{figure}{-1}
	\label{set4}
\end{figure}

\begin{figure}
	\centering
	\begin{subfigure}[t]{0.161\textwidth}
		\includegraphics[width=\textwidth]  
		{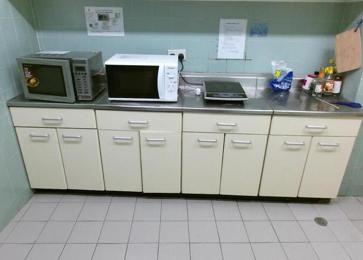}
		\includegraphics[width=\textwidth]
		{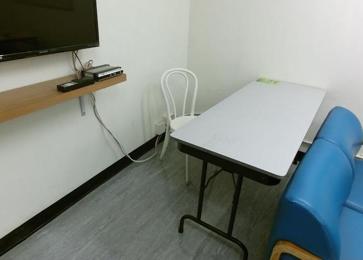}
		\includegraphics[width=\textwidth]
		{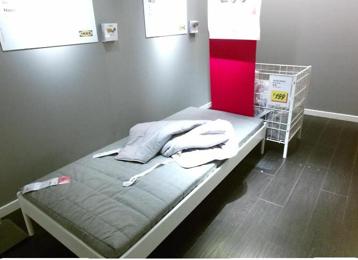}
		\includegraphics[width=\textwidth]
		{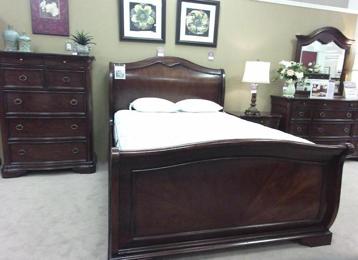}
		\includegraphics[width=\textwidth]
		{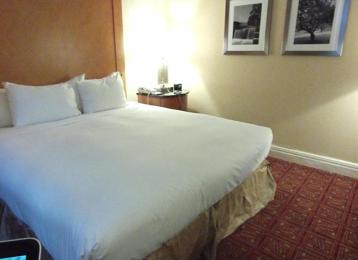}
		\includegraphics[width=\textwidth]
		{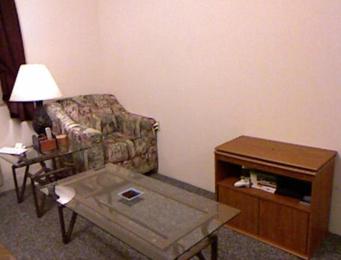}
		\includegraphics[width=\textwidth]
		{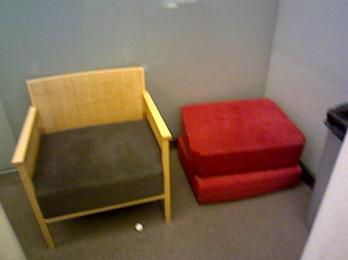}
		\includegraphics[width=\textwidth]
		{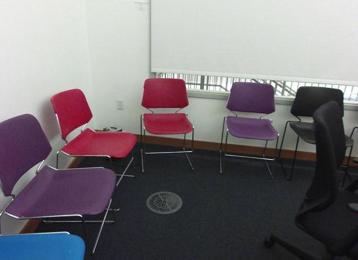}
		\caption{}
	\end{subfigure}
	\begin{subfigure}[t]{0.161\textwidth}
		\includegraphics[width=\textwidth]  
		{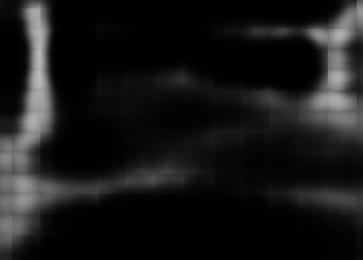}
		\includegraphics[width=\textwidth]
		{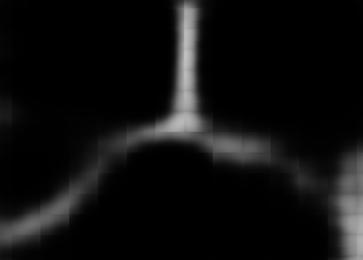}
		\includegraphics[width=\textwidth]
		{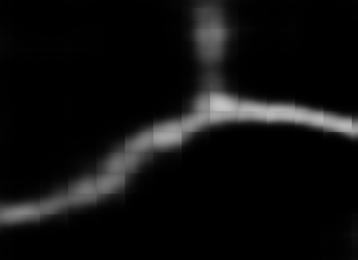}
		\includegraphics[width=\textwidth]
		{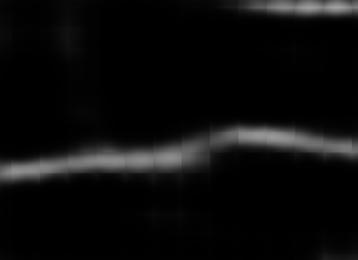}
		\includegraphics[width=\textwidth]
		{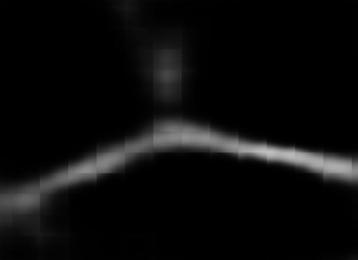}
		\includegraphics[width=\textwidth]
		{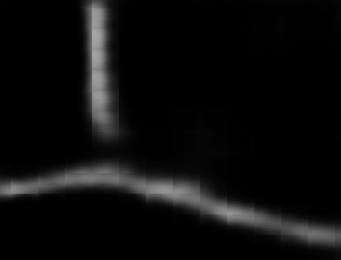}
		\includegraphics[width=\textwidth]
		{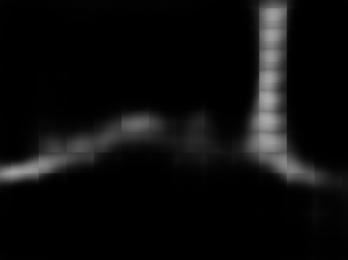}
		\includegraphics[width=\textwidth]
		{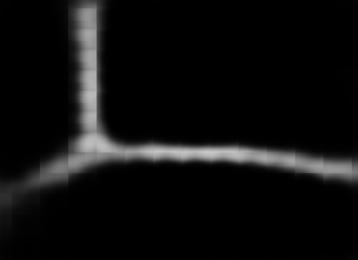}
		\caption{}
	\end{subfigure}
	\begin{subfigure}[t]{0.161\textwidth}
		\includegraphics[width=\textwidth]  
		{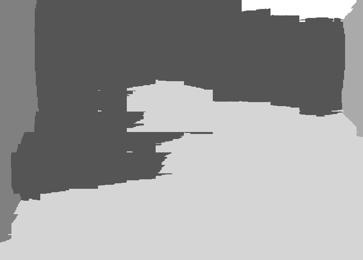}
		\includegraphics[width=\textwidth]
		{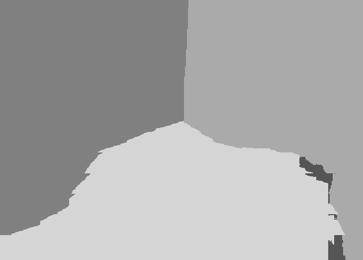}
		\includegraphics[width=\textwidth]
		{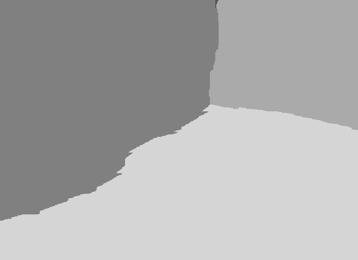}
		\includegraphics[width=\textwidth]
		{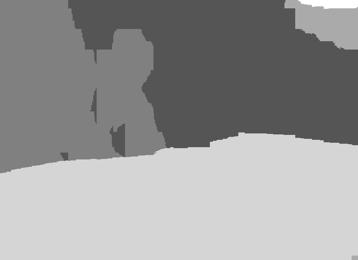}
		\includegraphics[width=\textwidth]
		{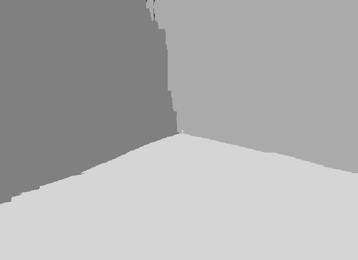}
		\includegraphics[width=\textwidth]
		{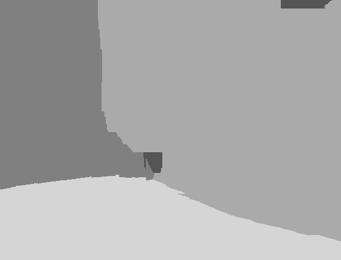}
		\includegraphics[width=\textwidth]
		{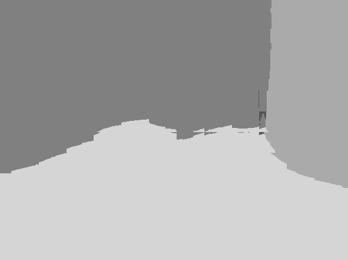}
		\includegraphics[width=\textwidth]
		{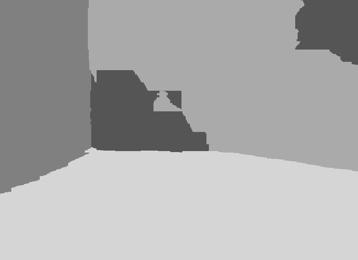}
		\caption{}
	\end{subfigure}
	\begin{subfigure}[t]{0.161\textwidth}
		\includegraphics[width=\textwidth]  
		{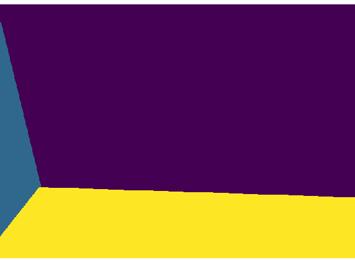}
		\includegraphics[width=\textwidth]
		{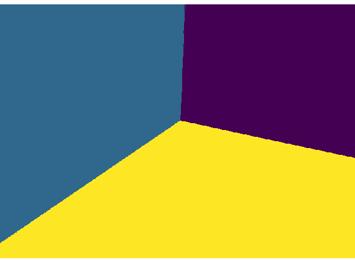}
		\includegraphics[width=\textwidth]
		{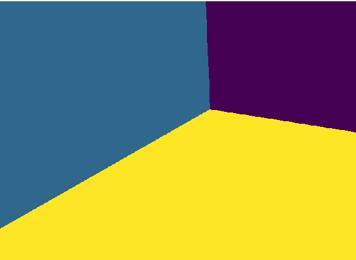}
		\includegraphics[width=\textwidth]
		{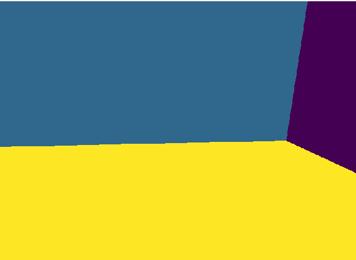}
		\includegraphics[width=\textwidth]
		{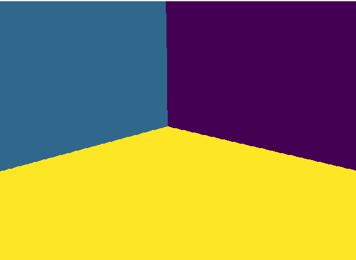}
		\includegraphics[width=\textwidth]
		{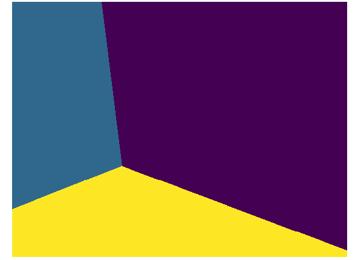}
		\includegraphics[width=\textwidth]
		{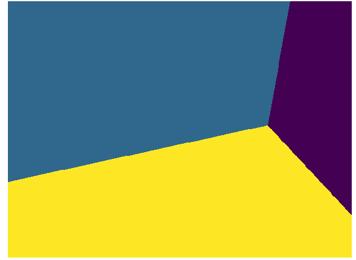}
		\includegraphics[width=\textwidth]
		{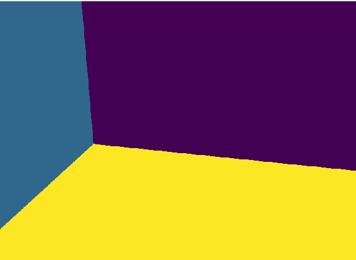}
		\caption{}
	\end{subfigure}
	\begin{subfigure}[t]{0.161\textwidth}
		\includegraphics[width=\textwidth]  
		{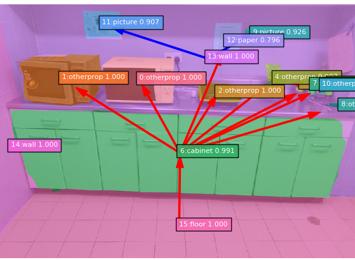}
		\includegraphics[width=\textwidth]
		{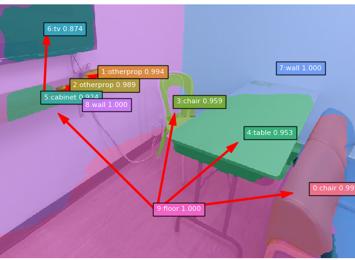}
		\includegraphics[width=\textwidth]
		{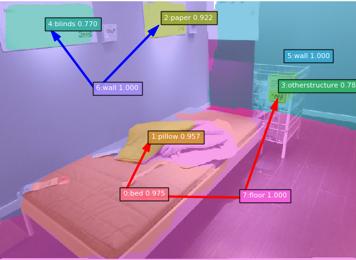}
		\includegraphics[width=\textwidth]
		{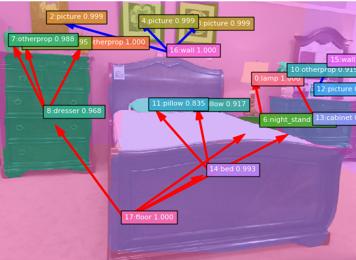}
		\includegraphics[width=\textwidth]
		{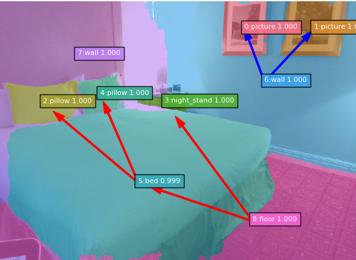}
		\includegraphics[width=\textwidth]
		{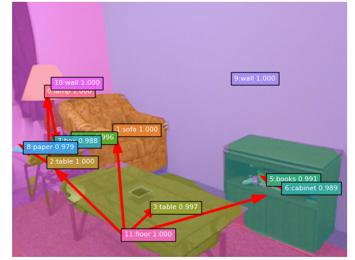}
		\includegraphics[width=\textwidth]
		{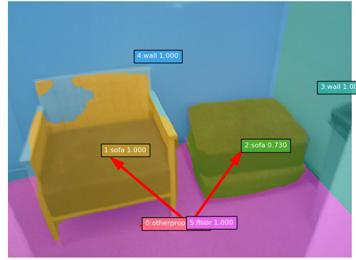}
		\includegraphics[width=\textwidth]
		{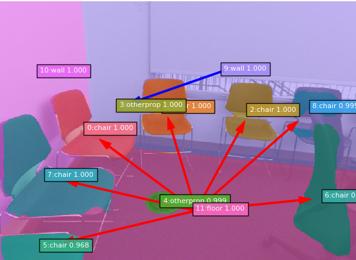}
		\caption{}
	\end{subfigure}
	\begin{subfigure}[t]{0.161\textwidth}
		\includegraphics[width=\textwidth]  
		{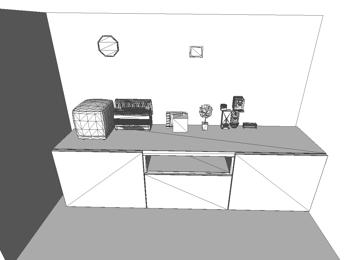}
		\includegraphics[width=\textwidth]
		{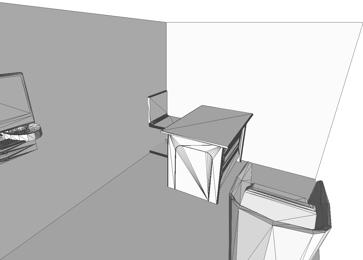}
		\includegraphics[width=\textwidth]
		{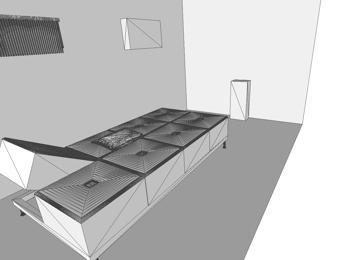}
		\includegraphics[width=\textwidth]
		{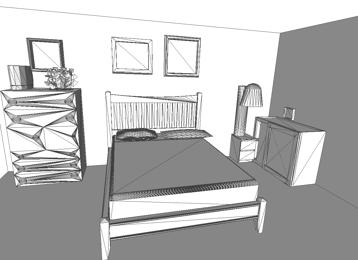}
		\includegraphics[width=\textwidth]
		{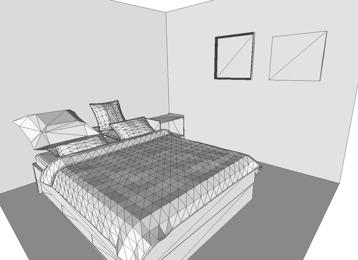}
		\includegraphics[width=\textwidth]
		{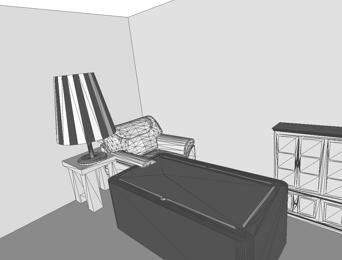}
		\includegraphics[width=\textwidth]
		{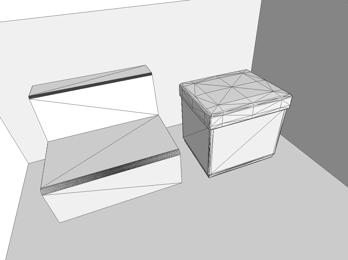}
		\includegraphics[width=\textwidth]
		{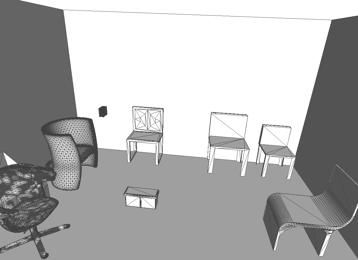}
		\caption{}
	\end{subfigure}
	\caption{Intermediate results in scene modeling.}
	\addtocounter{figure}{-1}
	\label{set5}
\end{figure}

\end{document}